\documentclass[10pt,twocolumn,letterpaper]{article}

\usepackage[pagenumbers]{iccv} 

\usepackage[hang, flushmargin]{footmisc}
\usepackage{tabularx}
\usepackage{booktabs}
\usepackage{multirow}
\usepackage{hhline}
\usepackage{colortbl}
\usepackage{xcolor}
\usepackage{makecell}
\usepackage[accsupp]{axessibility}

\definecolor{iccvblue}{rgb}{0.21,0.49,0.74}
\usepackage[pagebackref,breaklinks,colorlinks,allcolors=iccvblue]{hyperref}

\def\paperID{6801} 
\def\confName{ICCV}
\def\confYear{2025}

\begin{document}

\newcommand{\Etal}   {{\textit{et al.}}}

\newcommand{\jy}[1]{{\textbf{\textcolor{MidnightBlue}{[JY] }}\textcolor{MidnightBlue}{#1}}}

\newcommand{\ma}[1]{{\textbf{\textcolor{red}{[MA] }}\textcolor{red}{#1}}}

\newcommand{\by}[1]{{\textbf{\textcolor{violet}{[BY] }}\textcolor{violet}{#1}}}

\newcommand{\sai}[1]{{\textbf{\textcolor{magenta}{[SAI] }}\textcolor{magenta}{#1}}}

\newcommand{\change}[1]{{\color{red}#1}}

\newcommand{\cm}{\checkmark}

\definecolor{lightlightgray}{gray}{0.96}

\renewcommand{\thefootnote}{\textcolor{red}{\arabic{footnote}}}

\newcommand\blfootnote[1]{%
  \begingroup
  \renewcommand\thefootnote{}\footnote{#1}%
  \addtocounter{footnote}{-1}%
  \endgroup
}
\title{Multispectral Demosaicing via Dual Cameras}

\author{SaiKiran Tedla$^\texttt{*1,2}$\hspace{0.4cm} Junyong Lee$^\texttt{*1}$\hspace{0.4cm} Beixuan Yang$^\texttt{2}$\hspace{0.4cm} Mahmoud Afifi$^\texttt{1}$\hspace{0.4cm} Michael S. Brown$^\texttt{1,2}$\\
$^\texttt{1}$AI Center-Toronto, Samsung Electronics\hspace{0.4cm} $^\texttt{2}$York University \\
{\tt\small {\{s.tedla,j.lee8,m.afifi1,michael.b1\}@samsung.com}\hspace{0.4cm}\tt\small {\{tedlasai,byang,mbrown\}@yorku.ca}}
}

\maketitle
\begin{abstract}

Multispectral (MS) images capture detailed scene information across a wide range of spectral bands, making them invaluable for applications requiring rich spectral data. Integrating MS imaging into multi-camera devices, such as smartphones, has the potential to enhance both spectral applications and RGB image quality. A critical step in processing MS data is demosaicing, which reconstructs color information from the mosaic MS images captured by the camera. This paper proposes a method for MS image demosaicing specifically designed for dual-camera setups where both RGB and MS cameras capture the same scene. Our approach leverages co-captured RGB images, which typically have higher spatial fidelity, to guide the demosaicing of lower-fidelity MS images. We introduce the Dual-camera RGB-MS Dataset -- a large collection of paired RGB and MS mosaiced images with ground-truth demosaiced outputs -- that enables training and evaluation of our method. Experimental results demonstrate that our method achieves state-of-the-art accuracy compared to existing techniques.


\end{abstract}

\section{Introduction}
\label{sec:intro}
Multispectral (MS) imaging extends beyond standard RGB imaging by capturing spectral information across multiple wavelengths, often including visible and near-infrared spectra. This enables precise analysis for applications such as agriculture, medical imaging, and environmental monitoring \cite{arad2022ntire, warren2005agricultural, levenson2006multispectral, de2010uav}. MS data has also shown great potential for image enhancement \cite{zhang2008enhancing, shen2015multispectral, niu2023nir, JDM-HDRNet_ECCV2024}, making it a valuable addition to imaging pipelines.\blfootnote{Code and data are available at \href{https://ms-demosaic.github.io/}{https://ms-demosaic.github.io/}}
\blfootnote{$^*$These authors contributed equally to this work.}

As multi-camera systems become more common in modern smartphones, interest in integrating MS and RGB imaging has increased to leverage additional spectral data that can complement RGB images. 
While hyperspectral (HS) images provide denser and more contiguous spectral information than MS images, MS sensors are more practical for mobile devices, as HS imaging typically requires expensive and time-consuming capture systems~\cite{Basedow1995hydice, Schechner2002General, Gao2010Sanpsot}.
In turn, there has been growing research focused on incorporating MS sensors into mobile devices, demonstrating their ability to enhance performance in mobile RGB imaging tasks such as illuminant spectral estimation~\cite{glatt2024beyond}, image restoration~\cite{shen2015multispectral}, low-light enhancement~\cite{niu2023nir}, and tone adjustment~\cite{JDM-HDRNet_ECCV2024}.
However, most methods leverage MS imaging as a complementary prior to improving RGB-targeted tasks rather than focusing on enhancing MS image quality directly.
This is primarily due to the lower fidelity typically associated with the MS imaging pipeline.

\begin{figure}
\centering
\includegraphics[width=0.9\linewidth]{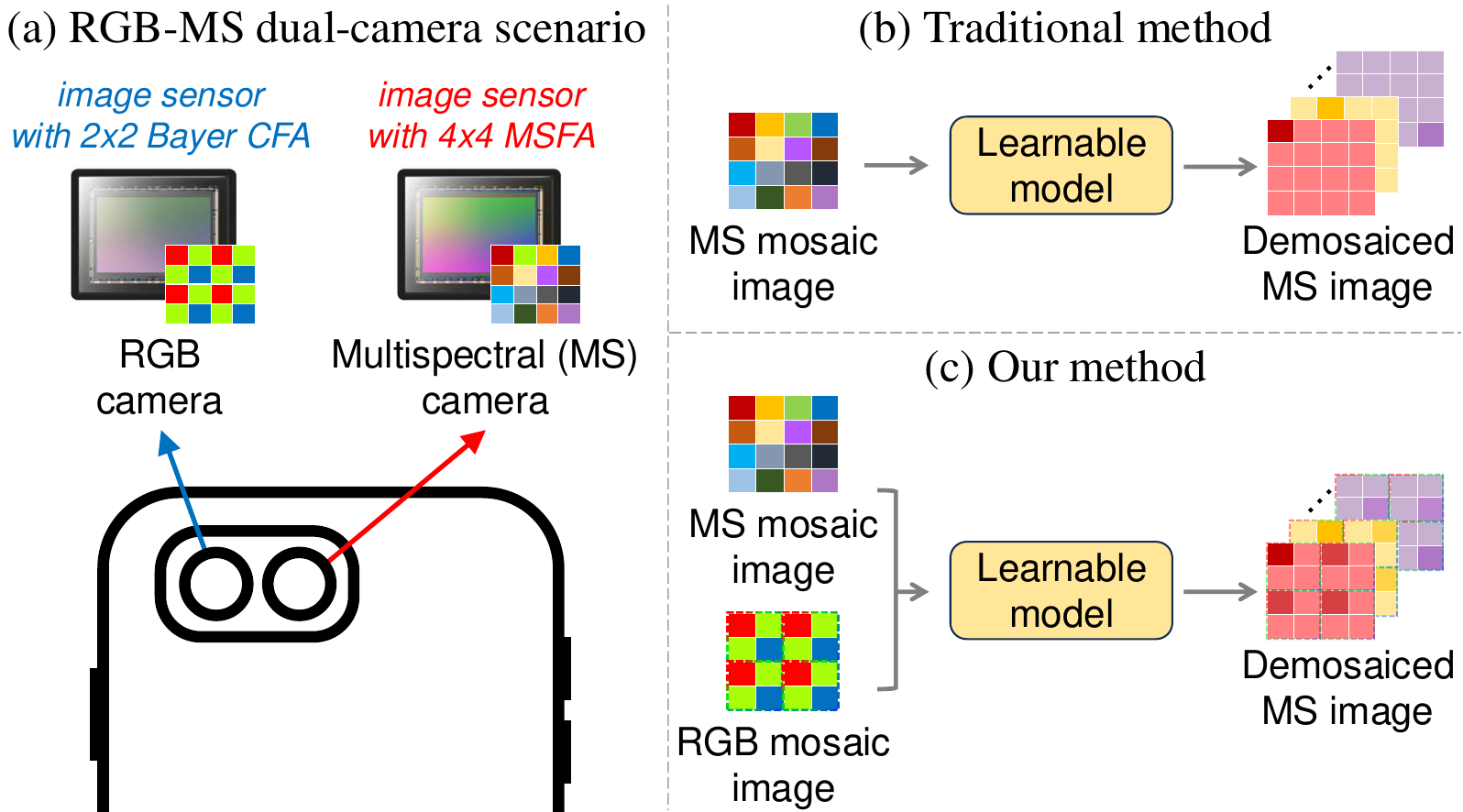}
\vspace{-2mm}
\caption{We propose a learning-based multispectral (MS) demosaicing method for a practical scenario (a), assuming a mobile device with a dual-camera setup, equipped with both RGB and MS cameras. Unlike traditional MS demosaicing approaches (b), our method (c) leverages the higher spatial resolution of the RGB mosaic from the RGB camera to guide and enhance the quality of the demosaiced MS image, achieving state-of-the-art results.}
\vspace{-10pt}
\label{fig:teaser}
\end{figure}

A standard RGB imaging pipeline adopts color filter arrays (CFAs) over the image sensor to capture red, green, and blue spectral bands. Each pixel records a single band, resulting in a single-channel mosaic raw image.
A commonly used CFA is the Bayer array \cite{Bayer1976Bayer}, which arranges the bands in a 2$\times$2 mosaic pattern.
Demosaicing algorithms then reconstruct the 3-channel RGB image by estimating the missing color values using the partial mosaic data \cite{li2008image}.


In contrast, an MS imaging pipeline employ more complex multispectral filter arrays (MSFAs), often arranged in 4$\times$4 mosaic patterns, to capture multiple spectral bands \cite{arad2022ntire_demosaicing, hahn2020detailed, murakami2012hybrid, lapray2014multispectral}.
While MSFA provides richer spectral information, the increased number of bands results in sparser mosaic data, making the demosaicing process considerably more challenging than in RGB demosaicing \cite{arad2022ntire_demosaicing}.

To illustrate, consider a practical example of smartphones integrating RGB and MS cameras in a multi-camera setup (Fig.~\ref{fig:teaser}a). Both cameras share identical lenses and sensors in this configuration, with the RGB and MS cameras using the 2$\times$2 Bayer CFA and 4$\times$4 MSFA, respectively. Although both cameras capture mosaic raw images at the same resolution, MS demosaicing is inherently more complex. Meanwhile, RGB demosaicing benefits from fewer missing pixels per color channel and denser spatial data.


This observation motivates us to develop a method specifically for MS demosaicing. Our method leverages the increasing potential of integrating RGB and MS cameras within the same device \cite{lee2022compact, roh2023spectral, glatt2024beyond}.
In particular, we utilize the RGB camera image as guidance in MS demosaicing (Fig.~\ref{fig:teaser}c), using its higher spatial fidelity to compensate for the MS image captured with MSFAs, which trades spatial resolution for more spectral channels compared to RGB CFAs. This ensures that the reconstructed MS images preserve rich spectral information and achieve the same resolution and spatial fidelity as the RGB counterparts.

\vspace{-12pt}
\paragraph{Contribution}
We propose an MS demosaicing method that leverages high-fidelity RGB images to address the low-fidelity nature of MS mosaic raw images (\cref{fig:teaser}).
We focus on a mobile setup with RGB and MS cameras mounted on a smartphone. As to the best of our knowledge, no handheld device with dual RGB and MS cameras provides access to both RGB and MS images, leading us to introduce a dual-camera RGB-MS dataset to train and validate our model.
%
Unlike existing datasets \cite{Tosi22RGBMSMatching, glatt2024beyond, JDM-HDRNet_ECCV2024}, which lack ground-truth MS images,
our dataset provides high-fidelity ground-truth MS images with detail comparable to RGB counterparts, enabling accurate evaluation of RGB-guided MS demosaicing.
Training on this dataset, our model learns to leverage high-resolution RGB data, achieving state-of-the-art performance and demonstrating the potential of dual-camera systems to enhance MS image quality significantly.
\section{Related work}
\label{sec:rel}

We first define the relation of RGB, MS, hyperspectral (HS) images and how they are captured. We note the boundaries between MS and HS are not well-defined, but we utilize the definitions from \cite{glatt2024beyond}.
First, RGB images capture spectral content that is integrated across three filters. RGB data is commonly captured on Bayer~\cite{Bayer1976Bayer} sensors (2$\times$2 CFA) by trading some spatial resolution for color information.  Next, MS image capture spectral content integrated across more filters. In our case, we examine cameras with 16 filters arranged in a 4$\times$4 CFA. Finally, HS images capture spectral content (without demosaicing) in a large number of contiguous spectral bands, offering high spectral resolution, but are impractical for smartphones due to expensive and slow scanning setups~\cite{Basedow1995hydice, Schechner2002General, Gao2010Sanpsot}. Our framework is designed for real-time MS imaging, making it feasible for mobile devices and practical applications. We now discuss the most relevant works to ours which are those of HS reconstruction and RGB/MS demosaicing.

\vspace{-12pt}
\paragraph{HS Image Reconstruction} 
HS image reconstruction methods generally fall into three categories:
(1) spectral super-resolution~\cite{mstpp, qu2023}, enhancing spectral resolution from high-resolution RGB/MS images (e.g., 3-ch RGB or 16-ch MS) to HS images; 
(2) spatial super-resolution~\cite{zhang2023essaformer, hssr1}, improving the spatial resolution of HS images while preserving spectral detail;
and (3) a hybrid approach~\cite{wu2023hsr, dct, fualigned, laiunaligned}, using RGB/MS images to guide HS reconstruction by combining the spatial advantages of RGB/MS images with the spectral richness of HS images. 
Our method is quite similar to hybrid HS image reconstruction~\cite{wu2023hsr, dct, fualigned}, since we leverage spatial advantages of RGB images for high-fidelity MS reconstruction.

\vspace{-12pt}
\paragraph{Demosaicing}
RGB demosaicing, extensively studied through classical signal processing and recent learning-based methods~\cite{malvar2004high, zhang2005color, hirakawa2005adaptive, menon2006demosaicing, wang2014multilayer, kim2019deep, gharbi2016deep}, use CFA mosaics (typically 2$\times$2 Bayer~\cite{Bayer1976Bayer}) to leverage dense spatial information for high-quality reconstruction. In contrast, MS demosaicing, though less explored, commonly works with 4$\times$4 MSFA mosaics and addresses challenges related to sparse spatial data~\cite{ssmt, feng2021MCan}. 
The limitations of both RGB and MS demosaicing motivate our framework, which combines the strengths of both approaches. Dense spatial information of RGB mosaics helps guide high-quality MS image reconstruction, despite their limited spectral content.

 \begin{figure*}
    \centering
    \includegraphics[width=0.85\textwidth]{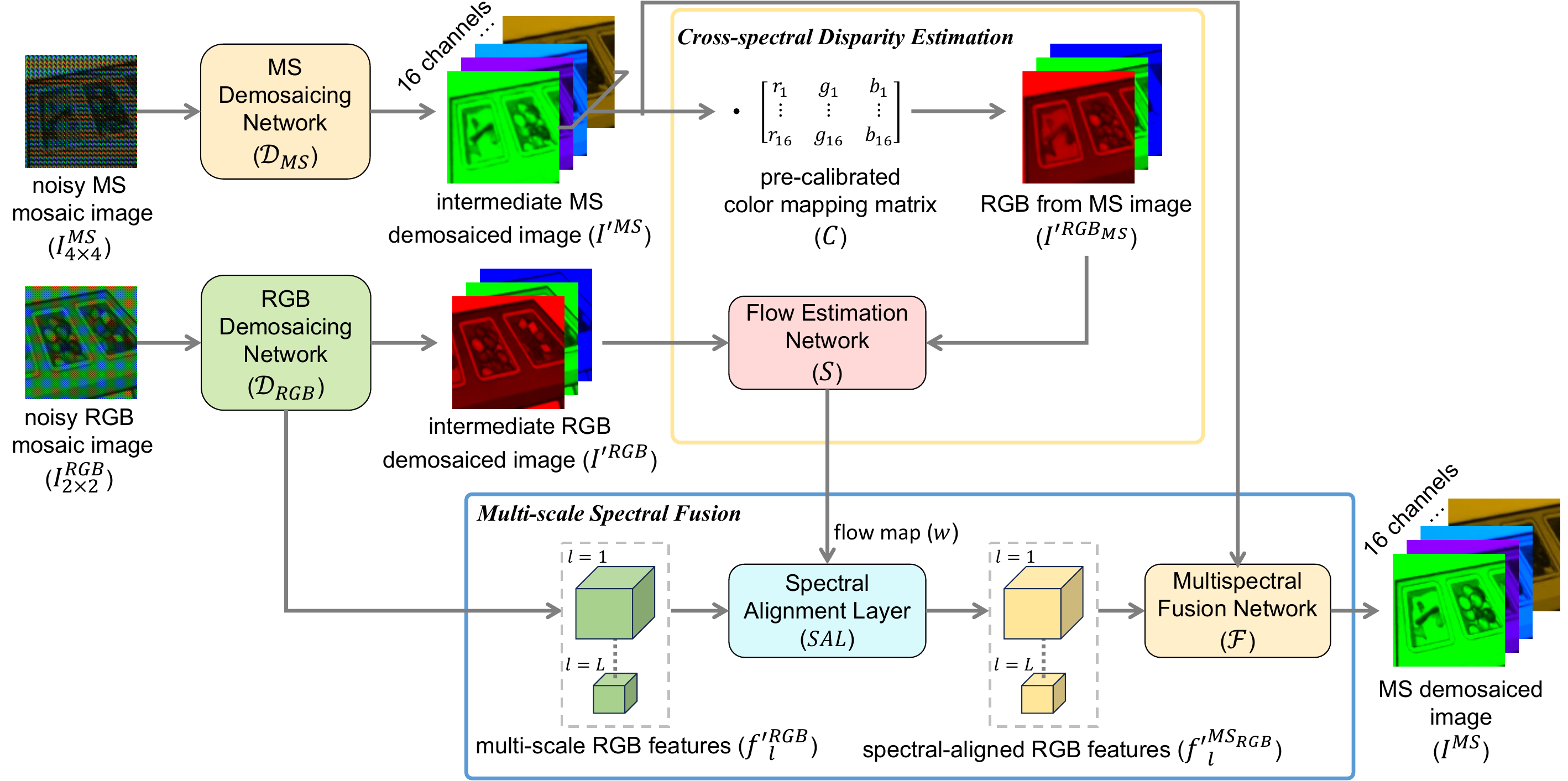}
    \vspace{-2mm}
    \caption{Our framework enhances MS demosaicing by integrating high-fidelity details from the co-captured RGB image. The framework consists of two stages: demosaicing (\cref{ssec:demosaicing}) and fusion (\cref{ssec:fusion}). In the demosaicing stage, the MS and RGB networks, $\mathcal{D}_{MS}$ and $\mathcal{D}_{RGB}$, reconstruct the MS and RGB images, ${I'}^{MS}$ and ${I'}^{RGB}$, respectively. In the fusion stage, high-fidelity details from the RGB image are fused into the MS image, while addressing both geometric and spectral disparities. The \emph{Cross-spectral Disparity Estimation} module computes flow map $w$ between MS and RGB images by first transforming the MS image into RGB color space to ensure spectral compatibility during flow estimation. Then, the \emph{Spectral Alignment Layer} ($SAL$) refines multi-scale RGB demosaicing features ${f'}_{l}^{RGB}$ into ${f'}_{l}^{{MS}_{RGB}}$, simultaneously compensating for geometric and spectral differences to align them with the MS image. Finally, the \emph{Multispectral Fusion Network} ($\mathcal{F}$) integrates refined RGB features ${f'}_{l}^{{MS}_{RGB}}$ into ${I'}^{MS}$, producing a high-fidelity MS image ${I}^{MS}$.}
    \label{fig:model}
    \vspace{-10pt}
\end{figure*}

\section{Multispectral Demosaicing}
\label{sec:method}

Figure \ref{fig:model} shows an overview of our MS demosaicing framework, specifically designed to handle both MS and RGB mosaic images captured in a dual-camera setup. Our framework takes MS and RGB mosaic images of the same scene captured with disparity and produces a demosaiced MS image while utilizing high-fidelity details of the RGB image.

We build our framework in two stages: demosaicing (\cref{ssec:demosaicing}) and fusion (\cref{ssec:fusion}).
In the first demosaicing stage, we employ two models to independently process the MS and RGB mosaic images.
In the fusion stage, a fusion module integrates the high-fidelity information from the demosaiced RGB image into the demosaiced MS image while addressing spatial disparities and spectral differences between the two images to produce the final demosaiced MS image with enhanced fidelity.

\subsection{MS and RGB Demosaicing}
\label{ssec:demosaicing}
In this stage, given the dual camera MS and RGB mosaic images, 
$I^{MS}_{4\times4}\in\mathbb{R}^{H\times W}$
and
$I^{RGB}_{2\times2}\in\mathbb{R}^{H\times W}$,
the goal is to reconstruct their demosaiced images ${I'}^{MS}\in\mathbb{R}^{H\times W \times16}$ and 
${I'}^{RGB}\in\mathbb{R}^{H\times W \times3}$ using the demosaicing networks $\mathcal{D}_{MS}$ and $\mathcal{D}_{RGB}$, respectively.
Here, 4$\times$4 and 2$\times$2 denote the mosaic pattern for MS and RGB mosaic images, respectively, and $H$ and $W$ represent the height and width of the images.
Formally, we have:
{\small
\begin{align}
\vspace{-2pt}
    {I'}^{MS} &= \mathcal{D}_{MS}(I^{MS}_{4\times4}), \label{eq:ms_demosaicing} \\
    {I'}^{RGB} &= \mathcal{D}_{RGB}(I^{RGB}_{2\times2}), \label{eq:rgb_demosaicing}
\vspace{-2pt}
\end{align}
}%
where we employ NAFNet~\cite{Chen2022NAFNet} as backbone networks for $\mathcal{D}_{MS}$ and $\mathcal{D}_{RGB}$, selected based on their reliable performance in MS and RGB demosaicing tasks~\cite{arad2022ntire_demosaicing, klap}.

Note that we perform separate demosaicing for the MS and RGB mosaic images to enable more precise and effective fusion in the subsequent stage, rather than attempting to fuse them directly. Aligning mosaic images with different patterns and spectral bands presents significant challenges, complicating the accurate fusion of MS and RGB images.


\subsection{Cross-Spectral Multi-Scale Fusion}
\label{ssec:fusion}
The RGB image ${I'}^{RGB}\in\mathbb{R}^{H\times W \times3}$, restored from 2$\times$2 mosaics, captures more high-frequency spatial details of a scene in contrast to the MS image ${I'}^{MS}\in\mathbb{R}^{H\times W \times16}$ reconstructed from 4$\times$4 mosaics. In this stage, our goal is to transfer the high-frequency spatial details from ${I'}^{RGB}$ to ${I'}^{MS}$, generating ${I}^{MS}$ with enhanced details.

Although ${I'}^{RGB}$ and ${I'}^{MS}$ capture the same scene, it is not straightforward to directly utilize ${I'}^{RGB}$ for enhancing ${I'}^{MS}$, as the image pair is misaligned, due to the disparity introduced within the dual-camera setup, and each image contains different spectral information. 
To address this, we compose this stage into two modules: first, \emph{Cross-spectral Disparity Estimation} that computes dense-correspondences between ${I'}^{MS}$ and ${I'}^{RGB}$; and second, \emph{Multi-scale Spectral Fusion} that integrates high-frequency spatial details of ${I'}^{RGB}$ with fewer spectral bands into ${I'}^{MS}$ with more spectral measurements to produce a final MS image ${I}^{MS}$, while compensating for the disparity between the images. In the following, we describe each module in more detail.

\vspace{-12pt}
\paragraph{Cross-Spectral Disparity Estimation}

Computing dense correspondence between images with different spectra remains as a challenging problem~\cite{Luo2016MatchingConventional, Chen2015MatchingConventional}. While some studies focus on cross-modal matching, they are limited to specific cases like RGB with Near InfraRed~\cite{Kim2016MatchingRGBNIR}, InfraRed~\cite{Chiu2011MatchingRGBIR}, varying illuminations~\cite{Shen2014MatchingRGBillum}, or MS images with limited spectral bands~\cite{Tosi22RGBMSMatching}.
A recent high-resolution (HS) image super-resolution approach~\cite{laiunaligned} uses stereo RGB images as guidance. For RGB image alignment, the method computes per-pixel flow between the HS and RGB images by first converting the HS image to RGB, leveraging the spectral response functions of both HS and RGB images. This process is relatively straightforward, as the HS image contains dense, contiguous spectral information that spans the RGB spectrum.

Inspired by this, we estimate cross-spectral disparity estimation between ${I'}^{MS}$ and ${I'}^{RGB}$
by employing a pre-calibrated color conversion matrix $C\in\mathbb{R}^{16\times3}$ to transform ${I'}^{MS}\in\mathbb{R}^{H\times W\times 16}$ into the proxy RGB ${I'}^{RGB_{MS}}\in\mathbb{R}^{H\times W\times 3}$, aligning it to the color space of the RGB image $I'^{RGB}$. Mathematically, we have:
{\small
\begin{equation}
\vspace{-2pt}
    {I'}^{RGB_{MS}}=r^{-1}\left(r\left({I'}^{MS}\right)\cdot C\right),
    \label{eq:color_conv}
\vspace{-1pt}
\end{equation}
}%
where {\small $r\left({I'}^{MS}\in\mathbb{R}^{H\times W\times 16}\right) \rightarrow {I'_{r}}^{MS}\in\mathbb{R}^{(H\times W)\times 16}$} is a reshaping operator and the color conversion matrix $C$ is pre-calibrated by computing a least squares transformation between the RGB and MS color chart image pairs (refer to the supplementary material for more details).

Then, we estimate the optical flow $w\in\mathbb{R}^{H\times W\times 2}$ between ${I'}^{RGB_{MS}}$ and ${I'}^{RGB}$ using the pre-trained flow estimation network $\mathcal{S}$~\cite{Teed2020RAFT}. Specifically, $w$ is obtained as:
{\small
\begin{equation}
\vspace{-2pt}
    w=\mathcal{S}({I'}^{RGB_{MS}},\,{I'}^{RGB}).
\label{eq:flow}
\end{equation}
}%
Note that the flow estimation network $\mathcal{S}$ first preprocesses ${I'}^{RGB}$ and ${I'}^{RGB_{MS}}$, mapping them into the sRGB color space using corresponding camera metadata (i.e., white balance and color correction matrices) to align the images with the color space used for the flow estimation task.
\vspace{-12pt}
\paragraph{Multi-Scale Spectral Fusion}
We now fuse the high-fidelity details of the RGB image ${I'}^{RGB}$ into the MS image ${I'}^{MS}$ to produce the final MS image ${I}^{MS}$, while compensating for the disparity between the two images using the estimated flow map $w$.
We propose a multi-scale spectral fusion network $\mathcal{F}$ to address this challenge.

The network $\mathcal{F}$ takes ${I'}^{MS}$ as its primary input to produce the final enhanced output ${I}^{MS}$. 
We introduce a \emph{spectral alignment layer} ($SAL$) to incorporate details from the RGB image. The layer takes $L$-level multi-scale RGB feature maps $\{{f'}^{RGB}_{l},\,\,l\in [1,2,...,L]\}$ extracted from the RGB demosaicing network and flow map $w$ to provide refined RGB feature map $f'^{{MS}_{RGB}}_{l}$ to each level of the fusion network $\mathcal{F}$.
We formally define the fusion process as:
{\small \begin{align}
\vspace{-2pt}
    f'^{{MS}_{RGB}}_{l} &= SAL({f'}^{RGB}_{l}, w),\label{eq:SAL}\\
    {I}^{MS} &= \mathcal{F}({I'}^{MS},\,\,{f'}^{{MS}_{RGB}}_{l\in{[1,2,...,L]}}), \label{eq:fusion}
\vspace{-2pt}
\end{align}
}%
where we adopt NAFNet~\cite{Chen2022NAFNet} as the backbone for the network $\mathcal{F}$. Here, the refined RGB feature maps $f'^{{MS}_{RGB}}_l$ are aligned with the MS image and adapted for fusion with the intermediate MS features within the network $\mathcal{F}$.

$SAL$ addresses geometric and spectral disparities simultaneously using the deformable convolution network (DCN)~\cite{Dai2027DCN}, which integrates seamlessly with the flow map $w$ and is particularly effective at capturing spatial features through adaptive sampling patterns.
$SAL$ computes the refined RGB feature map $f'^{{MS}_{RGB}}_{l}$ as the following:
{\small
\begin{equation}
\vspace{-2pt}
    f'^{{MS}_{RGB}}_{l}(p) = \sum_{i\in\Omega} k(i) {f'}^{RGB}_{l}(p+p_w+i+\Delta {i}),
\label{eq:dcn}
\vspace{-2pt}
\end{equation}
}%
where $p$ is the location on the output feature map $ f'^{{MS}_{RGB}}_{l}$, and 
$i$ enumerates the locations $\Omega$ in the deformable convolution kernel $k$.
The optical flow offset at location $p$, denoted as $p_w=w(p)/2^{l-1}$, represents the MS-to-RGB image displacement downscaled by $2^{l-1}$, enabling multi-scale geometric alignment of RGB to MS features. Here, $\Delta {i}$ represents the deformable kernel offsets learned by $SAL$, enabling the convolution kernel $k$ to adapt spatially and capture fine structural details in the multi-scale RGB features ${f'}^{RGB}_l$, such as edges and textures.

\begin{figure*}
    \centering
    \includegraphics[width=0.95\textwidth]{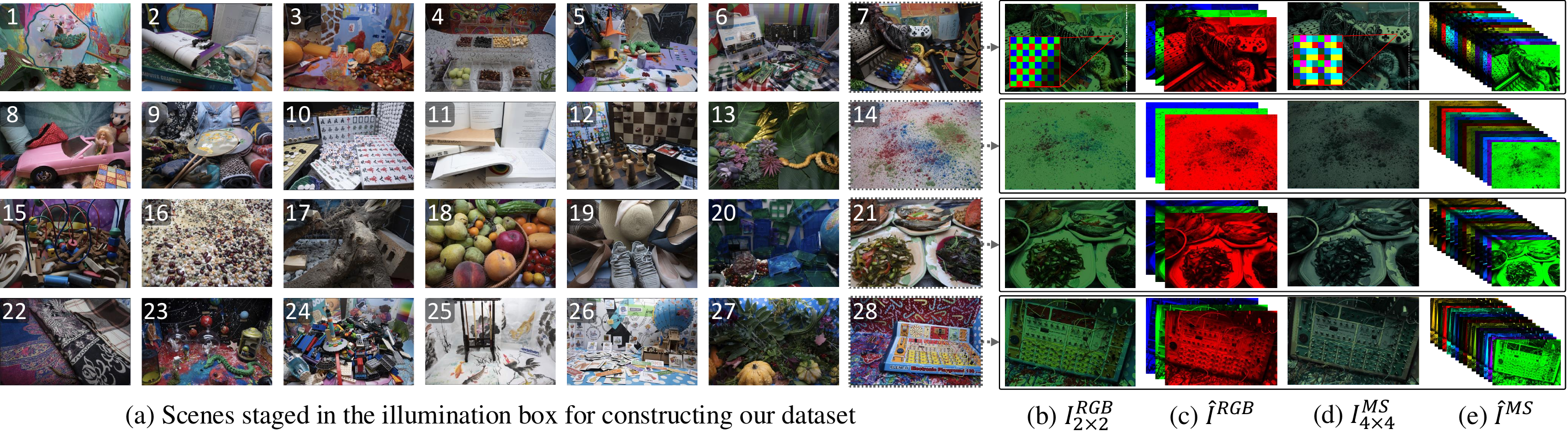}
    \vspace{-3mm}
    \caption{Representative examples from our dataset: (a) 28 scenes staged in the illumination box and (b–e) samples of quadruplets captured from scenes 7, 14, 21, and 28. Each quadruplet consists of (b) a 1-channel RGB mosaic image $I^{RGB}_{2\times2}$, (c) the corresponding 3-channel RGB demosaiced ground-truth $\hat{I}^{RGB}$, (d) a 1-channel MS mosaic image $I^{MS}_{4\times4}$, and (e) its corresponding 16-channel MS demosaiced ground-truth $\hat{I}^{MS}$. Note that the images within each quadruplet share the same spatial resolution. We also visualize the mosaic patterns (zoomed-in red cropped box) and the disparity between the RGB and MS mosaic images (white dashed line) in the first row.}
    \label{fig:scenes}
    \vspace{-12pt}
\end{figure*}

\subsection{Network Training}
\label{ssec:training}
The training process consists of two stages: demosaicing and fusion.
To train our network, we use our dual-camera RGB-MS dataset, consisting of quadruplets of mosaic MS and RGB images, each paired with ground-truth demosaic MS and RGB images, denoted as ${I}^{MS}_{4\times4}$, $I^{RGB}_{2\times2}$, ${\hat{I}}^{MS}$, and ${\hat{I}}^{RGB}$, respectively. \cref{sec:dataset} discusses more dataset details. 

\vspace{-12pt}
\paragraph{MS and RGB Demosaicing} We first train the MS and RGB demosaicing networks $\mathcal{D}_{MS}$ and $\mathcal{D}_{RGB}$ independently (\cref{ssec:demosaicing}). For $\mathcal{D}_{MS}$, we use the L2 loss between the predicted MS demosaiced image $I'^{MS}$, and the ground-truth MS image $\hat{I}^{MS}$.
Similarly, for the network $\mathcal{D}_{RGB}$, we apply the L2 loss between the predicted RGB image ${I'}^{RGB}$, and the ground-truth RGB image $\hat{I}^{RGB}$:
{\small
\begin{align}
\vspace{-2pt}
    \mathcal{L}_{MS}&=\|I'^{MS} - \hat{I}^{MS}\|_{2}, \label{eq:ms_dm_loss}\\
    \mathcal{L}_{RGB}&=\|{I'}^{RGB} - \hat{I}^{RGB}\|_{2}. \label{eq:rgb_dm_loss}
\vspace{-4pt}
\end{align}
}

\vspace{-12pt}
\paragraph{Cross-Spectral Fusion} In this stage, we train the multi-scale spectral fusion module (\cref{ssec:fusion}), which comprises the spectral alignment layer ($SAL$) and the fusion network $\mathcal{F}$ to incorporate high-fidelity details from $I'^{RGB}$ into the demosaiced MS image $I'^{MS}$, producing an enhanced MS image $I^{MS}$. The MS and RGB demosaicing networks and the optical flow estimation network, $\mathcal{S}$, remain fixed during this stage. We apply an L2 loss between the final MS image $I^{MS}$ and the ground-truth MS image $\hat{I}^{MS}$:
{\small
\begin{equation}
\vspace{-2pt}
    \mathcal{L}_{fusion}=\|I^{MS} - \hat{I}^{MS}\|_{2}.
\end{equation}
}


\section{Dual-Camera RGB-MS Dataset}
\label{sec:dataset}


\begin{figure}
    \centering
    \includegraphics[width=0.9\linewidth]{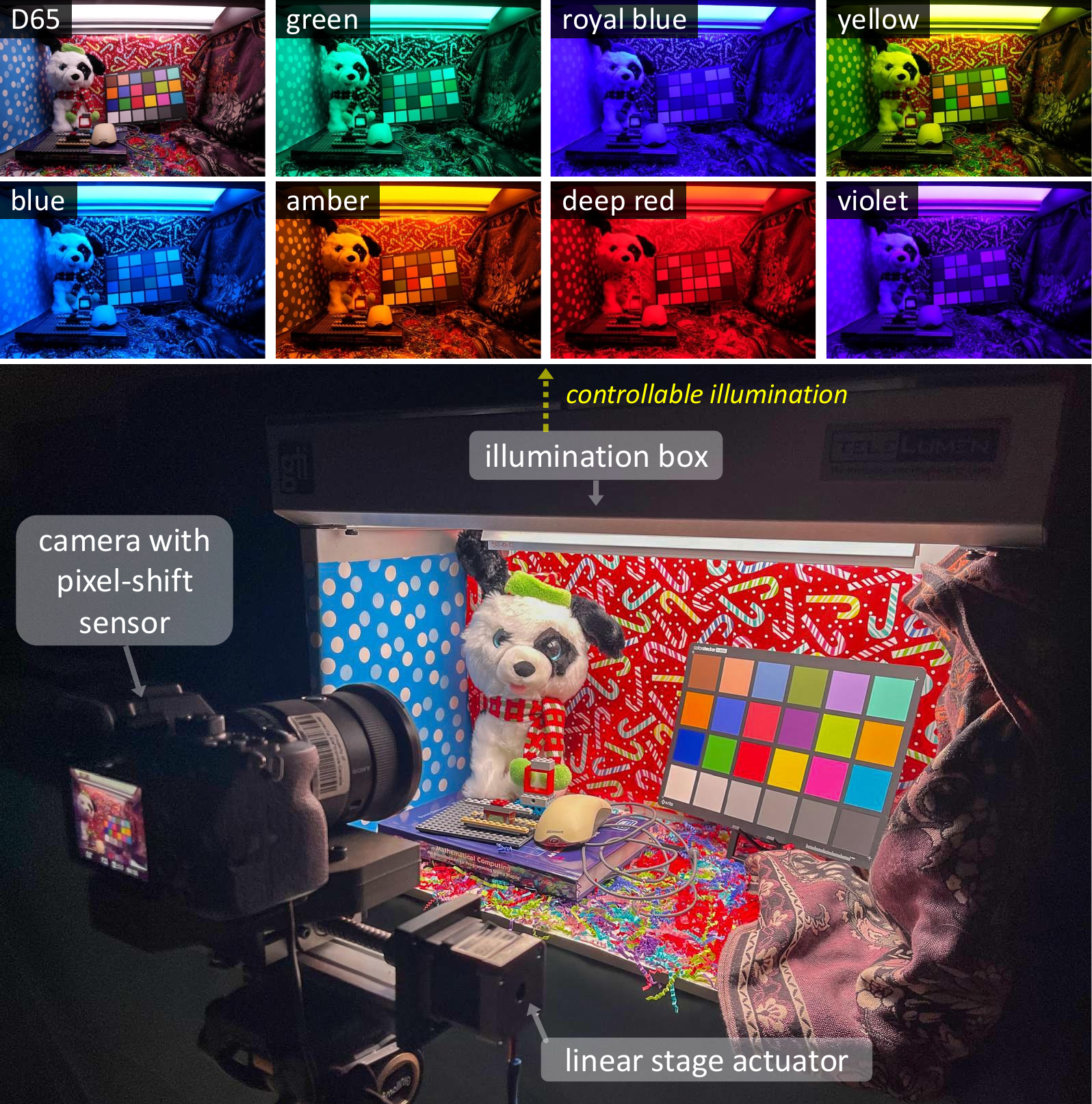}
    \vspace{-2mm}
    \caption{Dual-camera MS-RGB capturing system.}
    \label{fig:setup}
    \vspace{-18pt}
\end{figure}







To train and validate our network, we introduce a dataset containing quadruplets of mosaiced RGB and MS images, each paired with ground-truth demosaiced RGB and MS images (Figs.~\ref{fig:scenes}b-e). 
The ground-truth demosaiced images are high-quality captures (details of capture in Section~\ref{sec:gathering}) obtained using our imaging system that simulates an asymmetric dual-camera setup (\cref{fig:setup}). The system simulates an RGB camera capturing a scene in 3 RGB channels and an MS camera capturing the same scene in 16 multispectral channels, with a spatial disparity between them. We synthesize the mosaic images by converting the demosaiced images into 1-channel mosaics. The dataset comprises 502 quadruplets across 28 challenging scenes (Fig.~\ref{fig:scenes}a) with high textures and detailed features.
The dataset has training, validation, and test sets, containing 352, 47, and 103 image quadruplets captured from 20, 2, and 6 scenes, respectively,
where each image is in camera raw space at the resolution of 1440$\times$2160 pixels. Additional details about capture setup are given in the supplementary material. 

Existing MS datasets~\cite{glatt2024beyond, JDM-HDRNet_ECCV2024} are not well-suited for our task due to the absence of high-quality, ground-truth demosaiced images alongside the MS data. Additionally, these datasets often feature very low-resolution MS and RGB images in sRGB space already processed by camera pipeline~\cite{JDM-HDRNet_ECCV2024}, or paired RGB and MS images with minimal scene overlap due to extremely large disparities~\cite{glatt2024beyond}. 
Our method, designed for handheld devices like smartphones with minimal multi-camera disparity, led us to collect a dataset with realistic disparity and accurate ground-truth demosaic images. It includes both RGB and synthetic MS images in raw space, with no processing from the camera pipeline, making it ideal for training and evaluating our approach in the early stages of the onboard camera pipeline.

\subsection{Data Gathering Pipeline}
\label{sec:gathering}
To collect the dataset, we build an imaging system consisting of a camera mounted on a linear stage actuator and a controllable illumination box (\cref{fig:setup}),
that allows us to simulate an asymmetric dual-camera setup, where MS and RGB cameras have a constant relative baseline. 

The linear stage actuator moves the camera to two adjustable positions, one for MS and the other for RGB capture.
We utilize an Arduino/Genuino microcontroller to precisely control the camera movement between positions, ensuring a constant relative baseline between MS and RGB captures. In practice, we set the baseline to 1 cm.
The controllable illumination box (Telelumen Octa Light Player) simulates multispectral data capturing, featuring configurable light sources that distribute evenly throughout the scene within the box. Capturing occurs in a lab setting in a dark room to ensure that the box is the sole lighting source.

We use a Sony Alpha 1 camera as the capturing device.
For each capture, the system generates a pair of 16-channel MS and 3-channel RGB demosaic images using the pixel shift mode featured in the camera, which shifts the sensor during capture to enable sensor-level demosaicing.
Each image is initially captured at a resolution of 5760$\times$8640 pixels. To mitigate noise introduced by the small pixel size of the sensor, we downsample these images to a resolution of 1440$\times$2160 pixels to generate ground-truth demosaiced images.
We create our 1-channel MS and RGB input mosaic images by applying a 4$\times$4 MSFA pattern and 2$\times$2 Bayer CFA pattern to the MS and RGB ground-truth demosaiced images, respectively.
Additionally, we simulate noise on the mosaic images using a Poisson-Gaussian noise model~\cite{plotz}, calibrated at ISO 400. For this calibration, we use 90 images of a color chart (30 images taken at three different exposures) captured by the camera.


Like most consumer cameras, the camera in our system uses a CFA that captures the RGB spectrum, making RGB image acquisition straightforward. We obtain RGB images by configuring the light sources within the controllable illumination box to simulate the CIE D65 daylight illuminant. For MS image acquisition, we simulate an MSFA by capturing multiple RGB images of a scene under varying light sources. The following provides further background for simulating MS capture using the illumination box.


\vspace{-12pt}
\paragraph{Multispectral Image Acquisition}
To begin, we consider the image formation process in our setup under a uniform light source across the scene. 
Formally, the color information of a mosaic image $I$ at location $x$ can be described as:
\vspace{-2pt}
{\small
\begin{equation}
\label{eq:image}
    I(x) = \int_\gamma \underbrace{S(y) C_{rgb}(x, y)}_{\textit{CRF}_{\textit{RGB}}} \underbrace{L(y) R(x, y)}_{\textit{scene irradiance}}\, dy + z,
\end{equation}
}
\vspace{-2pt}

\noindent where $\textit{CRF}_{\textit{RGB}}$ represents the RGB camera response function (composed of the sensor's spectral sensitivity $S(y)$ and the RGB CFA response function $C_{rgb}(x, y)$), $L(y)$ is the spectral power distribution (SPD) of light, and $R(x, y)$ is the scene reflectance. 
$z$ denotes unwanted noise, typically characterized by signal-dependent and additive components \cite{abdelhamed2018high}. The integral over the visible range $\gamma$ at wavelength $y$ provides the color information in the mosaic raw image. 

Next, let us combine the camera response function with the light SPD emitted by the illumination box in narrow spectral bands, which mimic the spectral filters used in MS systems. We assume that the scene is lit by a uniform, broadband, and neutral ``virtual'' light source with an SPD denoted as $J\left(\cdot\right)$, which spans all wavelengths equally. \cref{eq:image} can then be rewritten as:
\begin{equation}
\vspace{-2pt}
    \label{eq:image}
    \resizebox{0.85\linewidth}{!}{$%
    I(x) = \int_\gamma \underbrace{S(y) C_{rgb}(x, y) L(y)}_{\textit{CRF}_{\textit{MS}}}  \underbrace{J(y) R(x, y)}_{\textit{simulated scene irradiance}}\, dy + z,
    $}
\vspace{-4pt}
\end{equation}

\noindent where $\textit{CRF}_{\textit{MS}}$ represents our virtual MS camera response function, and the scene irradiance is now simulated by the scene reflectance under the assumption of a constant, uniform, broadband, neutral virtual light source.

Since our capturing system uses the illumination box as the only physical light source in the scene, the box allows us to control the SPD of the light.  This capability enables us to simulate the MS mosaic image by capturing the scene with the RGB CFA, varying the SPD of the light within the illumination box, and performing multiple captures to simulate the response function of an MS camera.



The box provides seven primary wavelengths, ranging from 380 nm to 760 nm, which can be combined in various ways to create customizable light sources. For MS image acquisition, the system captures the same scene seven times, each under a different wavelength combination, resulting in a 21-channel MS image (7 wavelength combinations $\times$ 3 RGB channels). This is then reduced to a 16-channel MS image by discarding the 5 spectral channels with the least information. Detailed discussions and experiments are included in the supplementary material.

{
\setlength{\aboverulesep}{0mm}
\setlength{\belowrulesep}{0mm}
\renewcommand{\arraystretch}{1.0} 
\setlength{\tabcolsep}{0pt} 

\begin{table*}[t]
\centering
\scalebox{0.95}{
\small
\begin{tabular}{
>{\centering}p{0.07\textwidth} 
>{\centering}p{0.07\textwidth} 
>{\centering}p{0.14\textwidth} 
>{\centering}p{0.14\textwidth} 
>{\centering}p{0.14\textwidth} 
>{\centering}p{0.02\textwidth} 
>{\centering}p{0.08\textwidth} 
>{\centering}p{0.08\textwidth} 
>{\centering}p{0.08\textwidth} 
>{\centering}p{0.02\textwidth} 
>{\centering}p{0.08\textwidth} 
>{\centering\arraybackslash}p{0.08\textwidth} 
}

\toprule
\multirow{2}{*}[-0.5\dimexpr \aboverulesep + \belowrulesep + \cmidrulewidth]{\makecell{$\mathcal{D}_{MS}$\\(\cref{eq:ms_demosaicing})}} &
\multirow{2}{*}[-0.5\dimexpr \aboverulesep + \belowrulesep + \cmidrulewidth]{\makecell{$\mathcal{D}_{RGB}$\\(\cref{eq:rgb_demosaicing})}} &
\multicolumn{3}{c}{RGB guidance for fusion network $\mathcal{F}$ (\cref{eq:fusion})} &&

\multirow{2}{*}[-0.5\dimexpr \aboverulesep + \belowrulesep + \cmidrulewidth]{\makecell{PSNR$\uparrow$}}  & 
\multirow{2}{*}[-0.5\dimexpr \aboverulesep + \belowrulesep + \cmidrulewidth]{\makecell{SSIM$\uparrow$}} &
\multirow{2}{*}[-0.5\dimexpr \aboverulesep + \belowrulesep + \cmidrulewidth]{\makecell{SAM$\downarrow$}} && 
\multirow{2}{*}[-0.5\dimexpr \aboverulesep + \belowrulesep + \cmidrulewidth]{\makecell{Params\\(MB)}} &
\multirow{2}{*}[-0.5\dimexpr \aboverulesep + \belowrulesep + \cmidrulewidth]{\makecell{MACs\footnotemark[1]\\(T)}} \\
\cmidrule(lr){3-5}

&& \multirow{1}{*}[-0.5\dimexpr \aboverulesep + \belowrulesep + \cmidrulewidth]{\makecell{\scriptsize{$I'^{RGB}$}}}
& \multirow{1}{*}[-0.5\dimexpr \aboverulesep + \belowrulesep + \cmidrulewidth]{\makecell{\scriptsize{$I'^{RGB}(p + w(p))$}}}&
\multicolumn{1}{c}{$SAL$} && &&&&&\\

\midrule



\cm &     &      &&                    && 40.89 & 0.9766 & 2.604 && 111.24 & 0.78 \\
\arrayrulecolor{gray}
\cmidrule{1-12}
\rowcolor{lightlightgray}
\cm & \cm & \cm &  & && 40.90 & 0.9767 & 2.597 && 124.17 & 1.91 \\
\cm & \cm & & \cm &  && 41.75 & 0.9808 & 2.520 && 124.17 & 1.91 \\
\cmidrule{1-12}
\rowcolor{lightlightgray}
\cm & \cm & &&\cm && \textbf{41.92} & \textbf{0.9811} & \textbf{2.422} && 130.03 & 2.53 \\

\bottomrule
\end{tabular}}
\vspace{-6pt}
\caption{Ablation study evaluating the effect of RGB guidance strategies of the proposed MS demosaicing framework. We examine direct usage of the RGB demosaic $I'^{RGB}$, the warped image $I'^{RGB}(p + w(p))$ aligned to the MS image using the flow $w$ (\cref{eq:flow}), and feature-based guidance with the proposed spectral alignment layer ($SAL$, \cref{eq:SAL,eq:dcn}).}
\label{tab:ablation}
\vspace{-14pt}
\end{table*}
}
{
\setlength{\aboverulesep}{0mm}
\setlength{\belowrulesep}{0mm}
\renewcommand{\arraystretch}{1.00} 
\setlength{\tabcolsep}{0pt} 

\begin{table}[t]
\centering
\scalebox{0.9}{
\small
\begin{tabular}{
p{0.14\linewidth} 
p{0.25\linewidth} 
>{\centering}p{0.12\linewidth} 
>{\centering}p{0.12\linewidth} 
>{\centering}p{0.12\linewidth} 
>{\centering}p{0.01\linewidth} 
>{\centering}p{0.12\linewidth} 
>{\centering\arraybackslash}p{0.12\linewidth} 
}

\toprule
\multirow{2}{*}[-0.5\dimexpr \aboverulesep + \belowrulesep + \cmidrulewidth]{\makecell{Input}} & 
\multirow{2}{*}[-0.5\dimexpr \aboverulesep + \belowrulesep + \cmidrulewidth]{\makecell{Model}} & 
\multirow{2}{*}[-0.5\dimexpr \aboverulesep + \belowrulesep + \cmidrulewidth]{\makecell{PSNR$\uparrow$}} & 
\multirow{2}{*}[-0.5\dimexpr \aboverulesep + \belowrulesep + \cmidrulewidth]{\makecell{SSIM$\uparrow$}} &
\multirow{2}{*}[-0.5\dimexpr \aboverulesep + \belowrulesep + \cmidrulewidth]{\makecell{SAM$\downarrow$}} &&
Params (MB) &
MACs\footnotemark[1] (T) \\
\midrule


\multirow{7}{*}[-0.5\dimexpr \aboverulesep + \belowrulesep + \cmidrulewidth]{\makecell{\scriptsize{$I^{MS}_{4\times4}$}}} &SSMT~\cite{ssmt} & 33.97 & 0.932 & 6.723 && 9.76 & 26.01\\

&MCAN~\cite{feng2021MCan} & 39.02 & 0.966 & 3.947 && 5.24 & 0.91 \\
&MCAN-L & 40.36 & 0.974 & 2.996  && 53.69 & 7.80\\
\arrayrulecolor{lightgray}
\cmidrule{2-8}
&NAFNet~\cite{Chen2022NAFNet} & 40.89 & 0.977 & 2.604 && 111.25 & 0.78\\
&NAFNet-L & 40.68& 0.976 &2.654 && 158.58 &1.51\\
&Restormer~\cite{Zamir2021Restormer} & 40.61 & 0.977 & 2.643 && 99.70 & 6.72 \\
&Restormer-L & 40.58 & 0.976 &2.660 && 148.71 & 11.82 \\

\arrayrulecolor{gray}
\cmidrule{1-8}

\multirow{5}{*}[-0.5\dimexpr \aboverulesep + \belowrulesep + \cmidrulewidth]{\makecell{\scriptsize{$I^{MS}_{4\times4}$}\\\scriptsize{\&}\\\scriptsize{$I^{RGB}_{2\times2}$}}} 
&DCT~\cite{dct} & 38.44 & 0.962 & 4.748 && 31.91 & 15.22\\
&HSIFN~\cite{laiunaligned} & 36.50 & 0.963 & 3.198 && 90.21 & 18.79\\

\arrayrulecolor{lightgray}
\cmidrule{2-8}
&MCAN+Ours & \textcolor{RoyalBlue}{\textbf{41.85}} & \textcolor{RoyalBlue}{\textbf{0.981}}  & 2.572 && 24.03 & 2.66\\
&NAFNet+Ours & \textbf{41.92} & \textbf{0.981} & \textbf{2.422}&& 130.03 & 2.53\\
&Restormer+Ours & 41.48 & 0.981 & \textcolor{RoyalBlue}{\textbf{2.474}} && 118.49 & 8.47\\

\arrayrulecolor{black}
\bottomrule
\end{tabular}}
\vspace{-6pt}
\caption{Quantitative comparison for MS demosaicing.}
\label{tab:demosaicing}
\vspace{-16pt}
\end{table}
}

\section{Experiments}




\label{sec:results}

We train our model using the proposed dual-camera MS-RGB dataset (\cref{sec:dataset}). 
During training, we use the Adam optimizer~\cite{King2015Adam} with a learning rate of $1\!\times\!10^{\texttt{-}3}$. Following the two-stage training strategy (\cref{ssec:training}), we train the model for 200k iterations in both the demosaicing and fusion stages. For each iteration, we randomly sample batches of 8 quadruplets (mosaiced MS and RGB images, along with their corresponding demosaiced outputs) from the training set and cropped into 256$\times$256 patches. For the quantitative evaluation of MS restoration quality, we measure the Peak Signal-to-Noise Ratio (PSNR), Structural Similarity (SSIM)~\cite{wang2004ssim}, and Spectral Angle Mapper (SAM)~\cite{yuhas1992SAM}
.

\subsection{Ablation Study}
We conduct ablation studies focusing on the impact of the proposed cross-spectral multi-scale fusion module (\cref{ssec:fusion}) in incorporating RGB guidance for reconstructing the final MS demosaiced image.
We compare the baseline MS demosaicing network $\mathcal{D}_{MS}$ (\cref{eq:ms_demosaicing}), with NAFNet~\cite{Chen2022NAFNet} as the backbone, and three variants that utilize the RGB demosaicing network $\mathcal{D}_{RGB}$ (\cref{eq:rgb_demosaicing}) to provide RGB guidance to the multispectral fusion network $\mathcal{F}$.
The variants employ RGB guidance through image-based or feature-based methods. In the image-based guidance, the RGB demosaicked image $I'^{RGB}$ is used directly, with or without alignment to the MS image. Each of these is concatenated with the MS demosaicked image $I'^{MS}$ and fed into the fusion network $\mathcal{F}$. The feature-based guidance leverages configurations of the proposed spectral alignment layer ($SAL$).



Table~\ref{tab:ablation} presents quantitative results. The baseline model performs the worst due to the absence of RGB guidance (first row). Providing RGB guidance by concatenating $I'^{RGB}$ with the MS image $I'^{MS}$ yields slight improvements (second row), while flow-based alignment further enhances performance by addressing geometric misalignment (third row).
Incorporating multi-scale RGB features refined by the proposed $SAL$, which adapts deformable convolution~\cite{Dai2027DCN}, achieves the best performance (last row) and demonstrates its effectiveness in utilizing RGB features to facilitate cross-spectral fusion with multi-scale features (see supplementary for a more detailed analysis of $SAL$).

\subsection{Comparison on MS Demosaicing}
In this section, we evaluate the proposed method in two key asymmetric dual-camera scenarios. The first scenario focuses on asymmetry in the CFA patterns of the RGB and MS sensors. The second introduces an additional asymmetry where the MS sensor has a lower resolution than the RGB sensor, reflecting a practical consideration where a smaller MS sensor is used in the dual-camera setup.

We compare our RGB-guided MS demosaicing framework with previous image restoration methods, including hyperspectral image restoration and multispectral demosaicing methods: SSMT~\cite{ssmt}, MCAN~\cite{feng2021MCan}, DCT~\cite{dct}, HSIFN~\cite{laiunaligned}, NAFNet~\cite{Chen2022NAFNet, chu2022nafssr}, and Restormer~\cite{Zamir2021Restormer}. Among these, SSMT and MCAN are MS mosaic-to-MS reconstruction methods. NAFNet and Restormer are general-purpose image restoration models with state-of-the-art performance in tasks such as demosaicing~\cite{klap, xu2023demosaicformer} and super-resolution~\cite{chu2023scnafssr, zhu2023srrestormer}. DCT and HSIFN are hybrid MS-to-HS reconstruction methods that use RGB as a guidance. All baselines are retrained from scratch on our dataset using the same training configuration as our model and all are dtrained with the same number of iterations for fairness. 


For comparison, we categorize these methods into two types based on their input: a single MS mosaic (``{\scriptsize $I^{MS}_{4\times4}$}''), and a MS mosaic with an auxiliary RGB mosaic images (``{\scriptsize $I^{MS}_{4\times4}$}\,{\scriptsize $\&$}\,{\scriptsize $I^{RGB}_{2\times2}$}''). We adapt each model to handle the appropriate mosaiced inputs and generate demosaiced MS images.
Specifically, we adapt SSMT, MCAN, NAFNet, and Restormer to work with mosaiced MS images. DCT and HSIFN are modified to accept mosaiced MS images as input, while using mosaiced RGB as an auxiliary guidance. Additional details are provided in the supplementary. \footnotetext[1]{Measured on 1440$\times$2160 MS and RGB mosaic images.} \footnotetext[2]{Measured on 360$\times$540 MS and 1440$\times$2160 RGB mosaic images.}

In comparison, our framework is validated in a plug-and-play manner by integrating MCAN, NAFNet, and Restormer into a multispectral (MS) demosaicing network $D_{MS}$, on top of the RGB demosaicing network $D_{RGB}$ (\cref{eq:rgb_demosaicing}) and the cross-spectral fusion module (\cref{ssec:fusion}).
To ensure fairness, we account for the increased model complexity introduced by our approach by also evaluating baseline models with increased capacity, denoted as ``-L”. 

\begin{figure*}
    \centering
    \includegraphics[width=0.95\textwidth]{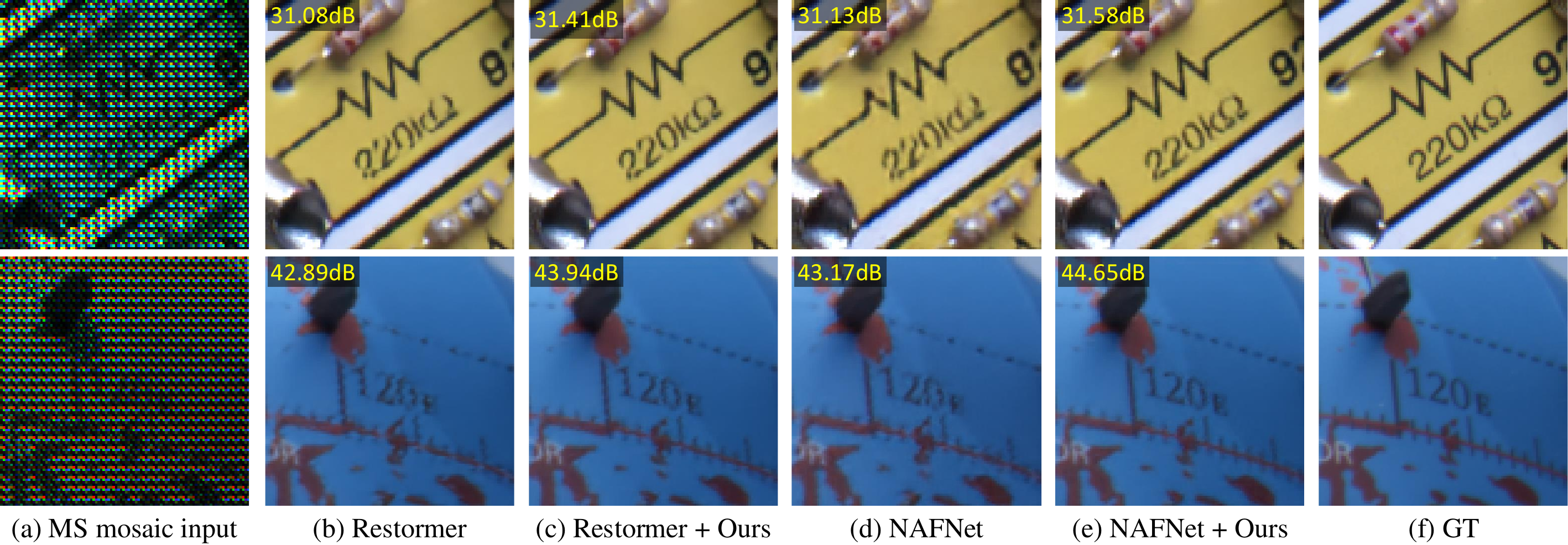}

    \vspace{-8pt}
    \caption{
    Qualitative comparison of MS demosaicing results for a dual-camera scenario where we consider MS and RGB sensors with the same spatial resolution but using asymmetric CFAs.
    Following \cite{Mihoubi2017MDPPI, feng2021MCan}, we visualize the predicted MS demosaics by converting them to the sRGB color space leveraging the color conversion matrix $C$ (\cref{eq:color_conv}) and camera metadata, using CIE D65 as the reference white point.
}
    \vspace{-12pt}
    \label{fig:exp_dm}
\end{figure*}

\vspace{-12pt}
\paragraph{Scenario 1: Asymmetric CFA Pattern}
In this scenario, we examine MS demosaicing in an asymmetric dual-camera configuration where the MS and RGB sensors share the same spatial resolution while using CFAs with different mosaic patterns: a 4$\times$4 MFSA for the MS sensor and a 2$\times$2 Bayer pattern for the RGB sensor.

{
\setlength{\aboverulesep}{0mm}
\setlength{\belowrulesep}{0mm}
\renewcommand{\arraystretch}{1.00} 
\setlength{\tabcolsep}{0pt} 

\begin{table}[t]
\centering
\scalebox{0.9}{
\small
\begin{tabular}{
p{0.14\linewidth} 
p{0.25\linewidth} 
>{\centering}p{0.12\linewidth} 
>{\centering}p{0.12\linewidth} 
>{\centering}p{0.12\linewidth} 
>{\centering}p{0.01\linewidth} 
>{\centering}p{0.12\linewidth} 
>{\centering\arraybackslash}p{0.12\linewidth} 
}

\toprule
\multirow{2}{*}[-0.5\dimexpr \aboverulesep + \belowrulesep + \cmidrulewidth]{\makecell{Input}} & 
\multirow{2}{*}[-0.5\dimexpr \aboverulesep + \belowrulesep + \cmidrulewidth]{\makecell{Model}} & 
\multirow{2}{*}[-0.5\dimexpr \aboverulesep + \belowrulesep + \cmidrulewidth]{\makecell{PSNR$\uparrow$}} & 
\multirow{2}{*}[-0.5\dimexpr \aboverulesep + \belowrulesep + \cmidrulewidth]{\makecell{SSIM$\uparrow$}} &
\multirow{2}{*}[-0.5\dimexpr \aboverulesep + \belowrulesep + \cmidrulewidth]{\makecell{SAM$\downarrow$}} &&
Params (MB) &
MACs\footnotemark[2] (T) \\
\midrule

\arrayrulecolor{gray}
\cmidrule{1-8}

\multirow{7}{*}[-0.5\dimexpr \aboverulesep + \belowrulesep + \cmidrulewidth]{\makecell{\scriptsize{$I^{MS\downarrow_{4}}_{4\times4}$}}}
&SSMT~\cite{ssmt} & 32.08 & 0.902 & 6.651 && 12.71 & 26.92\\
&MCAN~\cite{feng2021MCan} & 31.97 & 0.900  & 5.755 && 5.43 & 0.08\\
&MCAN-L & 32.87 & 0.916 & 4.884 && 53.88 & 0.51 \\
\arrayrulecolor{lightgray}
\cmidrule{2-8}
&NAFSR~\cite{chu2022nafssr} & 32.98 & 0.917 & 4.736 && 59.19 & 2.66\\
&NAFSR-L &32.95 & 0.919 & 4.439 && 100.93 &4.54\\
&Restormer~\cite{Zamir2021Restormer} & 32.34 & 0.908 & 5.166 && 99.99 & 0.93 \\
&Restormer-L & 32.45 & 0.911 & 4.882 && 149.00 & 1.26\\

\arrayrulecolor{gray}
\cmidrule{1-8}

\multirow{5}{*}[-0.5\dimexpr \aboverulesep + \belowrulesep + \cmidrulewidth]{\makecell{\scriptsize{$I^{MS\downarrow_{4}}_{4\times4}$}\\\scriptsize{\&}\\\scriptsize{$I^{RGB}_{2\times2}$}}}
&DCT~\cite{dct} & 31.12 & 0.893 & 6.613 && 42.21 & 15.78\\
&HSIFN~\cite{laiunaligned} & 29.92 & 0.866 & 6.982 && 90.35 & 6.61\\

\arrayrulecolor{lightgray}
\cmidrule{2-8}
&MCAN+Ours  &\textcolor{RoyalBlue}{\textbf{37.55}} & \textcolor{RoyalBlue}{\textbf{0.962}} & \textcolor{RoyalBlue}{\textbf{3.629}} && 24.22 & 1.82 \\
&NAFSR+ Ours & \textbf{37.67} & \textbf{0.964} & \textbf{3.567}&& 77.98& 4.41 \\
&Restormer+Ours & 36.70  & 0.960 &  3.722&& 118.77 & 2.68\\

\arrayrulecolor{black}
\bottomrule
\end{tabular}}
\vspace{-6pt}
\caption{Quantitative comparison for 4$\times$ MS demosaicing.}
\label{tab:exp_sr}
\vspace{-16pt}
\end{table}
}

For comparison, the baseline methods are trained using L2 loss between the predicted demosaiced MS image and ground-truth MS images (i.e., \cref{eq:ms_dm_loss}), while our models are trained using the complete pipeline described in \cref{ssec:training}.

Table~\ref{tab:demosaicing} provides quantitative results. NAFNet performs best among the baseline models, followed by Restormer, MCAN, and SSMT.
For capacity-increased baselines, while MCAN-L shows improvements, NAFNet-L and Restormer-L perform worse than their original versions, likely due to overfitting when using only MS mosaic input.
By incorporating our proposed method (rows marked ``+ Ours”), all models show significant improvements in MS restoration quality, demonstrating the adaptability of our approach across varying architectures.
The qualitative results in Fig.~\ref{fig:exp_dm} show how our method induces a baseline model to produce better details and structures. The supplementary materials provide additional examples to confirm the benefits of our approach in asymmetric CFA pattern scenarios.


\vspace{-14pt}
\paragraph{Scenario 2: Asymmetric Sensor Resolution}
This scenario extends Scenario 1 by introducing an additional asymmetry in sensor resolution, where the MS and RGB sensor captures low- and high-resolution mosaics, respectively.
In this scenario, we aim to reconstruct low-resolution MS mosaic images into high-resolution MS demosaiced images while preserving spectral fidelity.

For comparison, we adapt the baseline methods that takes MS mosaic image as an input to produce MS demosaic images at the desired spatial resolution.
Specifically, we modify SSMT~\cite{ssmt}, DCT~\cite{dct}, HSIFN~\cite{laiunaligned}, MCAN~\cite{feng2021MCan}, Restormer~\cite{Zamir2021Restormer} by appending upsampling layers consisting of multiple convolutional and pixel-shuffle layers~\cite{shi2016pixelshuffel}.
We employ NAFSR~\cite{chu2022nafssr}, a variant of NAFNet designed for the super-resolution task (refer supplement for details).
The training follows the same procedure as in Scenario 1, except that we synthesize low-resolution MS mosaic input images by downsampling the ground-truth MS demosaic images, followed by mosaicing to simulate data captured by a smaller MS sensor.
For downsampling, we use strided box filtering at a target scaling factor of 4.



\begin{figure}[t]
    \centering
    \includegraphics[width=1.00\linewidth]{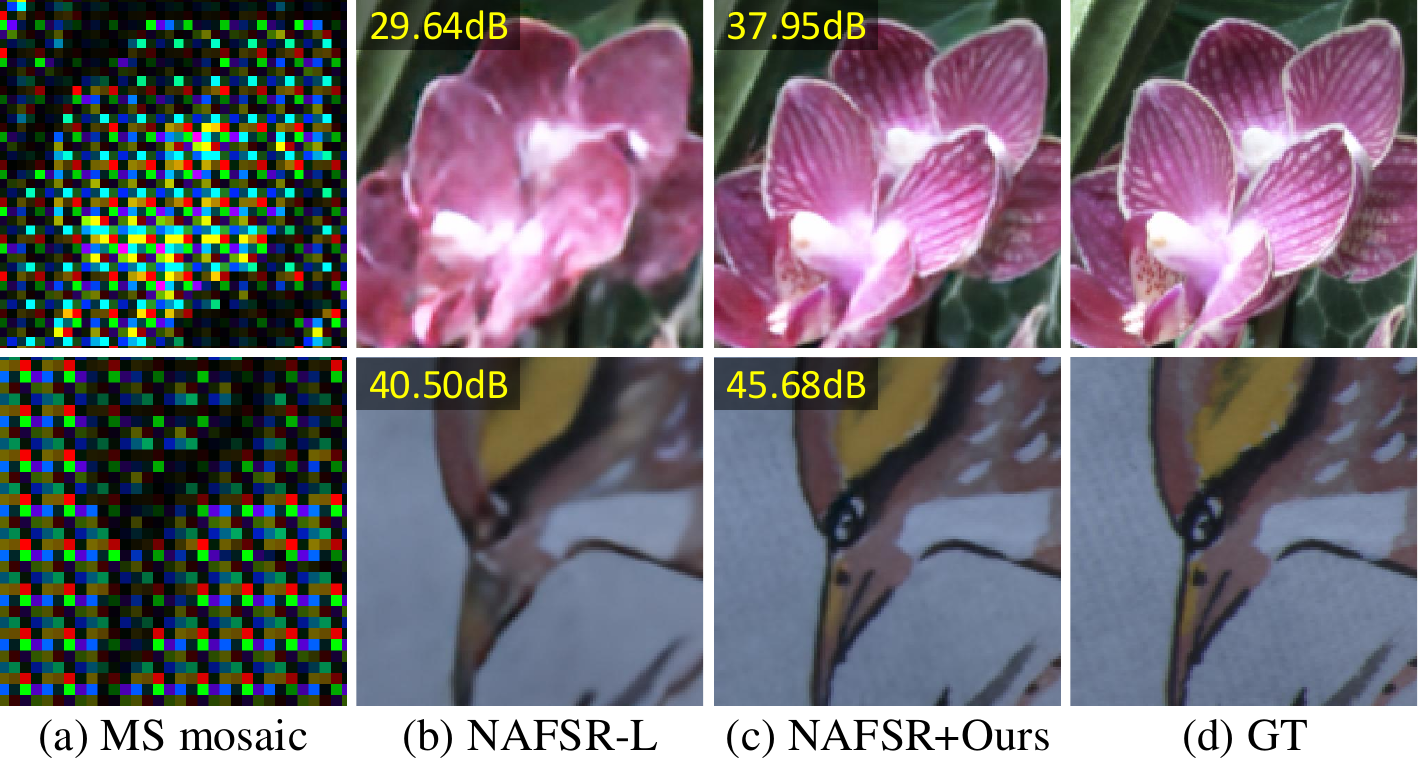}
    \vspace{-6mm}
    \caption{
    Qualitative comparison of 4$\times$ MS demosaicing results for a dual-camera configuration with asymmetric sensor resolution.
    The results are in the sRGB color space as described in \cref{fig:exp_dm}.}
    \label{fig:exp_sr}
    \vspace{-12pt}
\end{figure}

Table~\ref{tab:exp_sr} presents the quantitative results. Among the baselines, NAFSR-L achieves the best MS reconstruction performance, followed by Restormer-L and MCAN-L. Integrating our approach (rows marked ``+ Ours”) significantly enhances performance across all baselines, particularly in spectral accuracy, as SAM scores reflect. Unlike Scenario 1, where capacity increase led to overfitting for NAFNet and Restormer, ambiguity in dealing with low-resolution MS mosaic images allows these models to benefit from additional capacity.
Figure~\ref{fig:exp_sr} provides qualitative results.
Our approach significantly improves reconstructing high-fidelity details and accurate structures, demonstrating the advantages of the proposed dual-camera scenario in enhancing MS images by leveraging high-quality RGB mosaics.

\section{Conclusion}
\label{sec:conclusion}
This paper proposes an MS demosaicing framework leveraging high-fidelity RGB guidance from a dual-camera MS-RGB setup to enhance MS image quality. Our framework integrates a cross-spectral fusion module to address geometric misalignment and spectral disparities, effectively combining RGB and MS information. We also introduce a dual MS-RGB dataset with high-fidelity ground-truth MS images, enabling accurate evaluation of RGB-guided MS demosaicing. Experiments demonstrate the state-of-the-art MS restoration quality of our approach, highlighting the potential of dual-camera systems to advance MS imaging.

\vspace{-12pt}
\paragraph{Limitation}
Our method enhances MS image restoration using high-fidelity RGB guidance but relies on dual RGB-MS camera setups, which are not yet widely adopted in commercial devices. Its effectiveness also depends on RGB image quality and precise cross-spectral alignment, as misalignment or noise can impact MS reconstruction.
Future work could focus on reducing computational overhead to improve feasibility for resource-constrained devices.


\clearpage
\setcounter{page}{1}
\maketitlesupplementary
\section{Dual-camera MS-RGB Dataset}
The proposed dataset provides high-quality image quadruplets consisting of mosaiced RGB and MS images alongside their respective demosaiced ground truths. These images are collected using a custom-built imaging system designed to simulate an asymmetric dual-camera setup, as described in Sec.~4.1 of the main paper.\blfootnote{Code and data are available at \href{https://ms-demosaic.github.io/}{https://ms-demosaic.github.io/}}\blfootnote{$^*$These authors contributed equally to this work.}

\begin{figure*}
    \centering
    \includegraphics[width=1.0\textwidth]{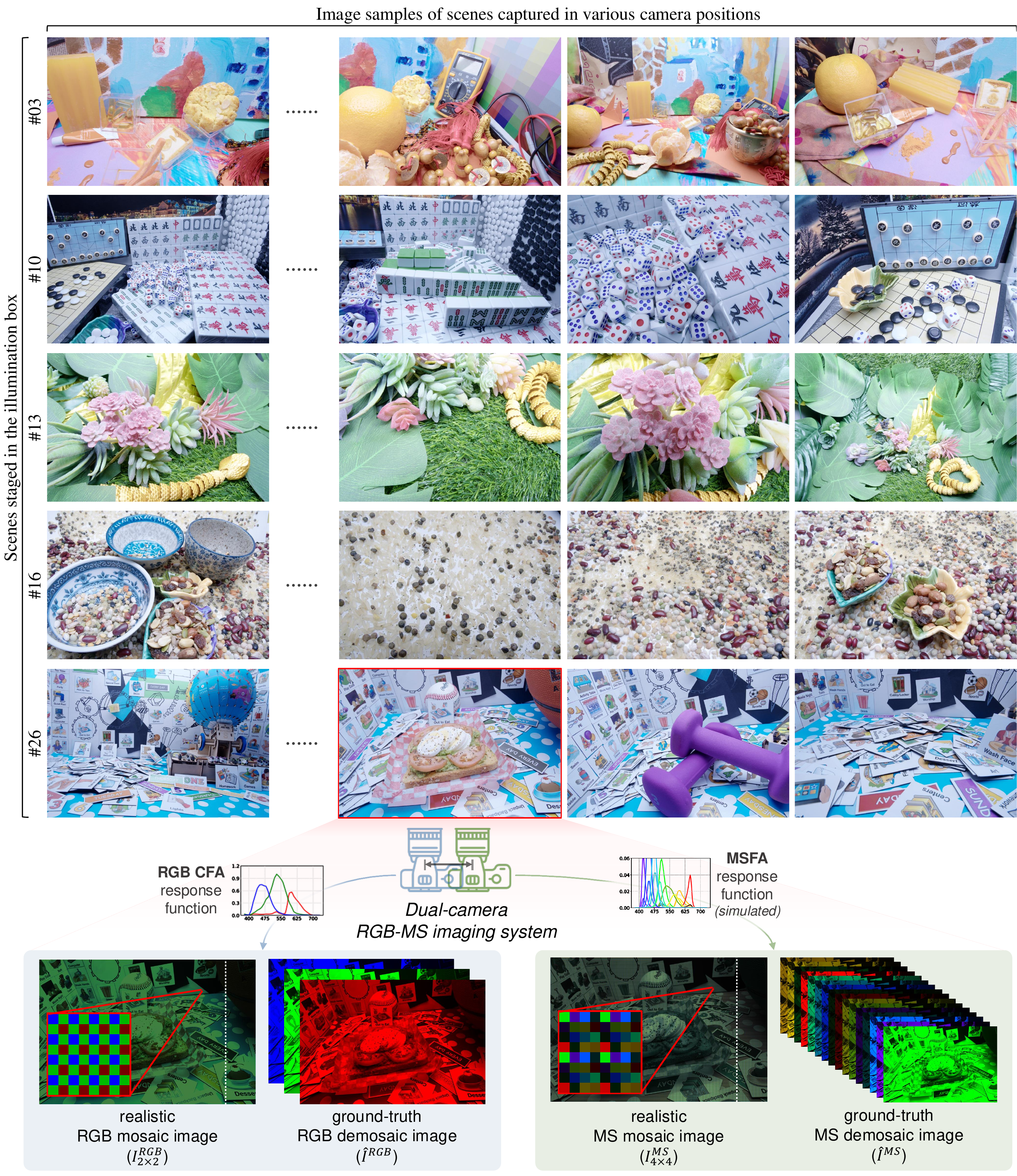}
    \vspace{-6mm}
    \caption{Overview of the data capturing pipeline: The system uses a Sony Alpha 1 camera with an RGB Bayer CFA sensor and pixel-shift mode to capture high-resolution ground-truth RGB and multispectral (MS) images via sub-pixel shifts. The camera, mounted on a linear stage actuator, captures staged scenes from different positions while maintaining a fixed baseline. RGB images are acquired in an illumination box simulating CIE D65 daylight, while MS images use a multispectral filter array (MSFA) under varied lighting. The dataset includes 502 quadruplets from 28 challenging scenes, featuring diverse staged setups.}
    \label{fig:data_overview}
\end{figure*}

Our system employs a Sony Alpha 1 camera with a RGB Bayer color filter array (CFA) sensor, featured with pixel-shift mode, to capture accurate ground-truth demosaiced RGB and MS images through sub-pixel shifts. The camera, mounted on a linear stage actuator, captures staged scenes in an illumination box from different positions while maintaining a fixed relative baseline between the RGB and MS captures (illustrated in \cref{fig:data_overview}). RGB acquisition is performed by configuring the illumination box to simulate the CIE D65 daylight illuminant, while MS acquisition is achieved by simulating a multispectral filter array (MSFA) through capturing the same scene under varying light sources within the box.
After capturing the demosaiced RGB and MS images, further processing generates the mosaiced images. This involves synthesizing noise to replicate realistic sensor conditions and applying mosaic patterns using a 2$\times$2 Bayer CFA for RGB images and a 4$\times$4 MSFA for MS images.

\subsection{16-Band MS Image Acquisition}
\label{ssec:2116}
For capturing MS image acquisition, our imaging system simulates a total of 21 multispectral response functions by combining the CFA response functions of the camera with the spectral power distributions (SPDs) of varying light sources provided by the illumination box. The illumination box provides seven primary wavelengths, ranging from 380 nm to 760 nm, which can be combined in various ways to create customizable light sources. By leveraging the configurable illumination feature provided by the box, the system captures the same scene seven times, each under a different wavelength combination, resulting in a 21-channel MS image (7 wavelength combinations $\times$ 3 RGB channels). This is then reduced to a 16-channel MS image by discarding 5 spectral channels with the least information.


\begin{figure*}
    \centering
    \includegraphics[width=1.0\textwidth]{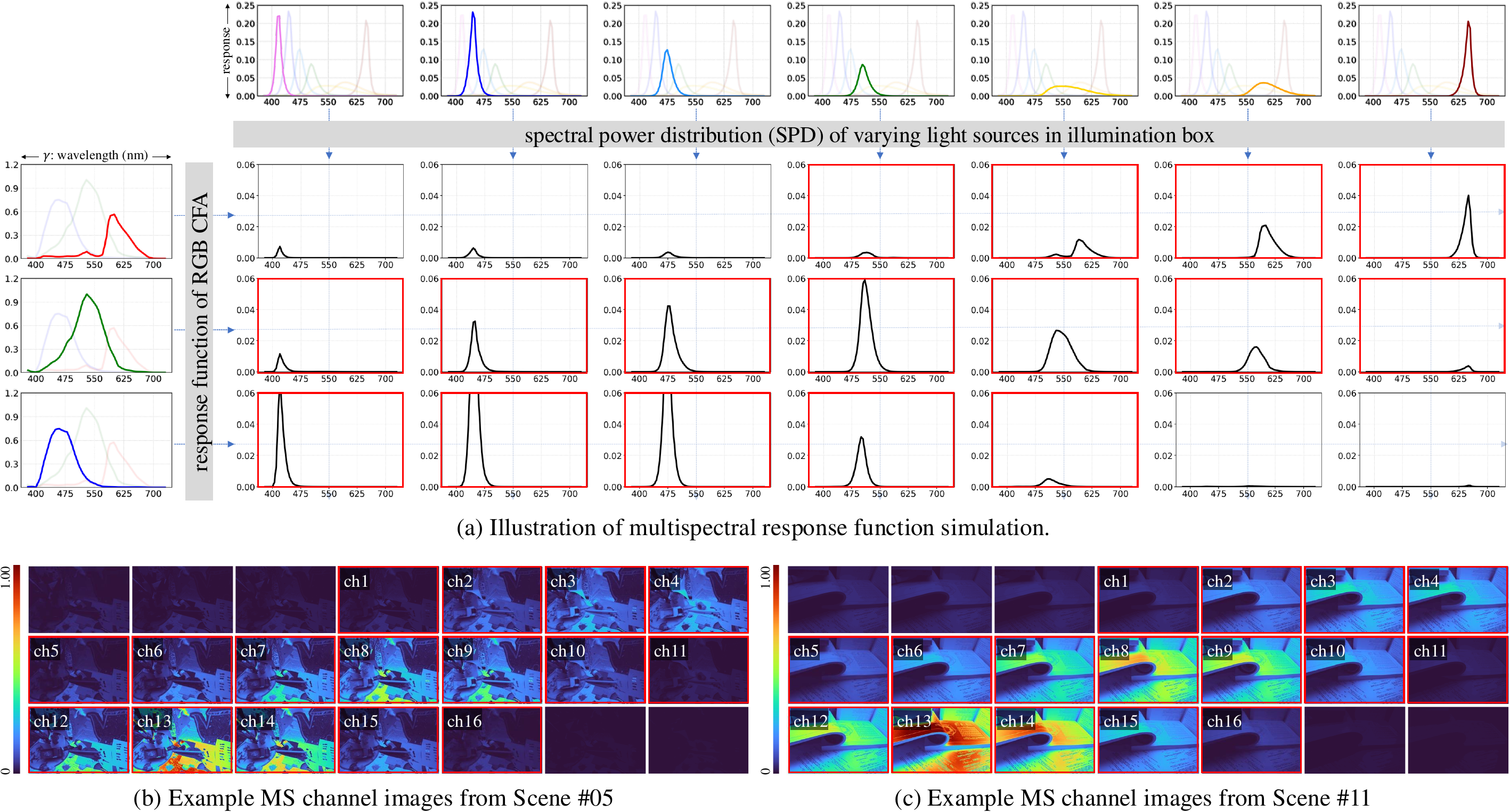}
    \vspace{-6mm}
    \caption{Illustration of 16-channel MS image acquisition in our imaging system. (a) shows the 4$\times$4 MSFA simulation using and RGB CFA camera (first column) and varying light sources provided by a configurable illumination box (top row). Given the RGB response function $C^i_{rgb}(x,\gamma)$ of the camera and 7 distinct SPDs $L^{j}(\gamma)$ provided by configurable illumination box, we simulate 21 multispectral response functions as $C_{ms}^{k}=C^i_{rgb}(x,\gamma)L^{j}(\gamma)$ for each $i\in[1,2,3]$ and $j\in[1,2,...,7]$. From these 21 response functions, 16 are selected to simulate a 4$\times$4 MSFA (highlighted by red boxes). (b-c) are examples of MS images of different scenes captured using the multispactral response functions. The selected 16-channel images (highlighted by red boxes) are used to construct the ground-truth MS demosaic image $\hat{I}^{MS}$ in the proposed dual-camera RGB-MS dataset.}
    \label{fig:21_16}
    \vspace{-12pt}
\end{figure*}

\cref{fig:21_16} illustrates the channel selection process for creating 16-channel MS images. Using an RGB camera and a configurable illumination box, we simulate 21 multispectral response functions (\cref{fig:21_16}a). 
The response functions of the RGB CFA camera, denoted as $C_{rgb}^{i}(x, \gamma)$, define the sensitivity of the red, green, and blue channels across the visible wavelength range $\gamma$. 
The illumination box provides spectral power distributions (SPDs), $L^j(\gamma)$, corresponding to seven distinct wavelength bands.
The RGB CFA response functions are calibrated using camSPECS V2, which captures images of several monochromatic light sources at different wavelengths and measures the output intensities for each channel. These measured values are compared against the known SPD of the light source to determine the spectral sensitivity of each CFA channel. For the SPDs of the illumination box, we use data provided by the manufacturer (Telelumen Octa Light Player).

\label{ssec:21_16}

The multispectral responses are derived by combining the RGB response functions with the SPDs of the box:
\begin{equation}
\vspace{-2pt}
C_{{ms}}^{k}(x, \gamma) = C_{rgb}^{i}(x, \gamma)L^{j}(\gamma),\, 
\scriptstyle{\forall i \in [1,2,3],\,j \in [1,2,\dots,7]}.
\label{eq:ms_synth}
\end{equation}
This results in a total of 21 response functions (\ie, $k \in [1,2,\dots,21]$), representing all combinations of RGB channels and illumination SPDs.

To reduce the 21 channels to 16, we consider the area of the response functions and select filter responses that are distributed across the visible spectrum. Specifically, we compute the integral of each response function $C_{ms}^{k}(x, \gamma)$ over the visible range $\gamma$, which quantifies the spectral contribution of each channel. Based on these integrals, we select the top 12 channels with the largest areas. For the remaining 8 channels, which have smaller spectral contributions, we heuristically choose 4 channels to ensure coverage across different wavelengths. The final 16 response functions, highlighted in red in \cref{fig:21_16}a, are mapped to the MSFA grid to create spatially multiplexed MS images (\cref{fig:21_16}b-c). These MS images are then combined to construct the high-quality MS demosaiced image, which serves as a key component of the proposed dual-camera RGB-MS dataset.

\subsection{Noise Calibration}
Given demosaiced RGB and MS images captured by our imaging system, we generate their corresponding mosaic images. To mitigate noise caused by the small pixel size of the sensor, we first downsample the pixelshift demosaiced images from 5640$\times$8760 to 1440$\times$2160. These downsampled clean images serve as our demosaiced ground-truth in our proposed dataset.
Next, we apply synthetic noise and then mosaic the images using a 2$\times$2 Bayer CFA for RGB and a 4$\times$4 MSFA for MS to obtain the final mosaic images.

To simulate realistic sensor noise, we use a Poisson-Gaussian noise model~\cite{plotz,Alessandro2008PGN,Qian2019PGN}. Given the clean image $I\!\in\!\mathbb{R}^{H\!\times\! W\!\times\! N}$, noisy image $Y\!\in\! \mathbb{R}^{H\!\times\! W\!\times\! N}$ is modeled as:
\begin{align}
\vspace{-2pt}
Y_n(x) = I_n(x) + \epsilon_n(I_n(x)),
\vspace{-2pt}
\end{align}
where $n$ denotes the channel index and $\epsilon_n(I_n(x))$ represents the noise at pixel location $x$. The noise distribution is calibrated using heteroscedastic modeling, which accounts for the per-pixel signal dependency of photon noise. Mathematically:
\begin{align}
\vspace{-2pt}
&\epsilon_n(I_n(x)) \sim \mathcal{N}(0, \sigma_n^2(I_n(x))), \text{where}\\
&\sigma_n^2(I_n(x)) = \beta_n^1 I_n(x) + \beta_n^2.
\vspace{-2pt}
\end{align}
Here, $\sigma_n^2(I_n(x))$ represents the intensity-dependent noise variance. The parameter $\beta_n^1$ models photon shot noise, proportional to the pixel intensity $I_n(x)$, while $\beta_n^2$ accounts for intensity-independent electronic read noise.

Following the procedure outlined in~\cite{plotz,Alessandro2008PGN}, we calibrate the noise parameters for each RGB CFA channel of the Sony Alpha 1 camera. To this end, we capture images of the X-Rite color chart under different exposures and ISO levels, and fit a linear model to the scatter plot of the calculated mean and variance pairs of pixel intensities at all homogeneous patches of the color chart images.
This linear fit describes the heteroscedastic noise variance as a function of pixel intensity, determined separately for each color channel at different ISO values. 
In practice, our pipeline synthesizes noise using $\beta_n^1$ and $\beta_n^2$ calibrated at ISO 400. Nonetheless, it is worth mentioning that noise synthesis can be easily extended to other ISO levels, as mosaic images can be regenerated using the high-quality ground-truth RGB and MS demosaiced images in our dataset.

Once $\beta_n^1$ and $\beta_n^2$ are calibrated for the camera, we apply the noise to the clean RGB and MS demosaiced images captured by our system, based on their spectral channel and intensity value. Note that the noise model calibrated for the RGB CFA is directly applicable to simulate MS images, which are simulated by combining RGB channel responses $C_{rgb}(x, \gamma)$ with varying SPDs $L^j(\gamma)$ from the illumination box (\cref{eq:ms_synth}). The noise variance depends only on the pixel intensity $I(x)$, which aggregates the contributions of SPDs through the image formation process (Eqs.~(11) and (12) in the main paper).
While SPDs can indirectly influence $I(x)$, the noise model itself is calibrated based on intensity and intrinsic sensor characteristics, making it agnostic to specific SPD.
As such, the calibrated noise model remains valid for simulated MS images, ensuring consistency in noise synthesis regardless of variations in the illuminant or spectral composition, as long as the pixel intensities are preserved.

\subsection{Color Conversion Matrix Calibration}
To enable cross-spectral alignment in our MS demosaicing framework (Sec.~3.2 of the main paper), our dataset includes a pre-calibrated MS-to-RGB color conversion matrix. 
This matrix converts the MS image into RGB color space, ensuring spectral compatibility between the MS and RGB images during geometric alignment. 
The matrix is calibrated using RGB and MS images of the X-Rite Digital-SG color chart, which contains 140 color patches.

\begin{figure}
    \centering
    \includegraphics[width=1.0\linewidth]{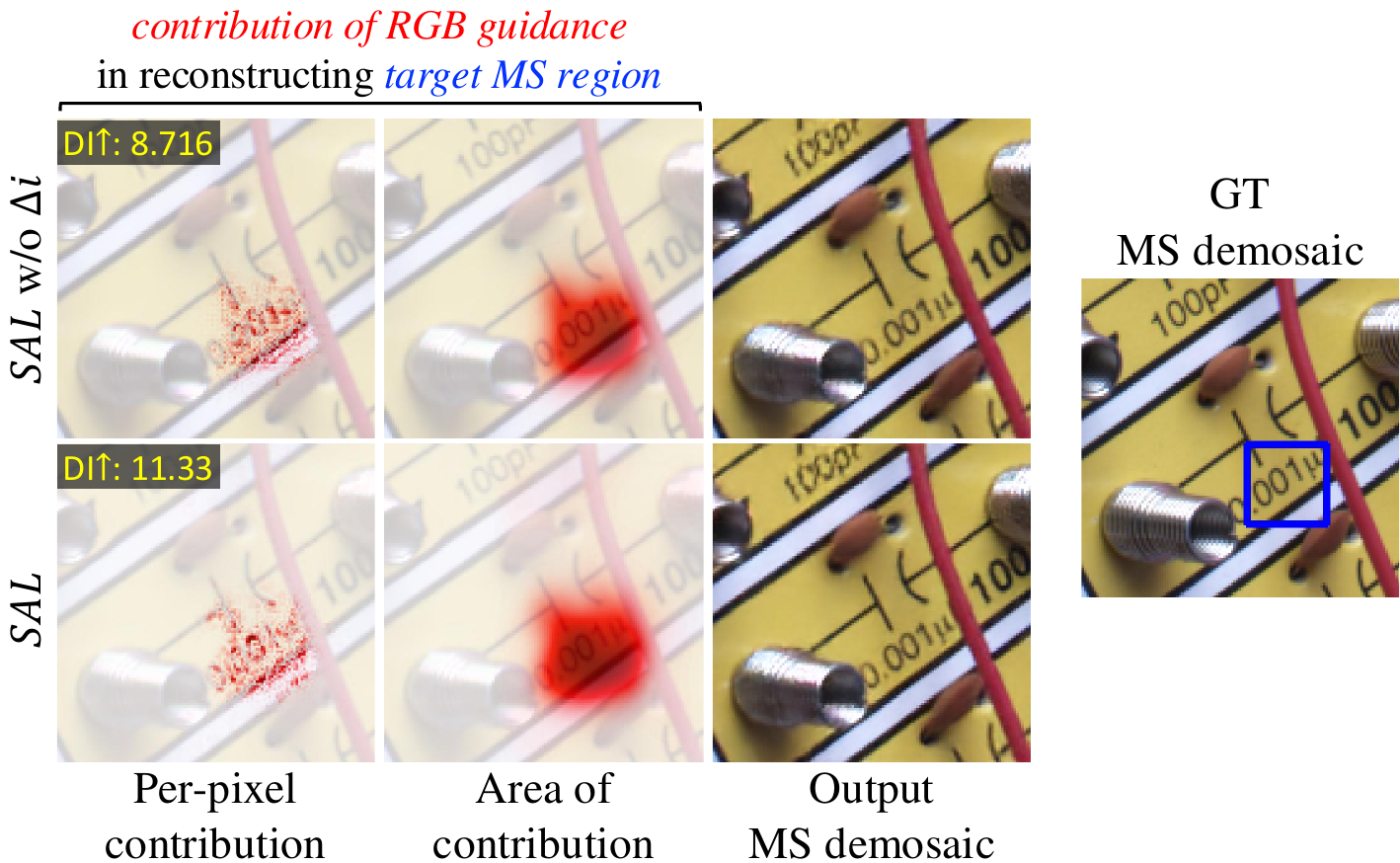}
    \vspace{-6mm}
    \caption{Qualitative analysis of the Spectral Alignment Layer ($SAL$). The figure compares results without spectral alignment offsets ($SAL$ w/o $\Delta i$, top row) and with the offsets ($SAL$, bottom row). Columns represent the per-pixel contribution of RGB guidance, the area of contribution, and the reconstructed MS demosaiced output. The results indicate that $SA$L enhances the effectiveness of RGB guidance in reconstructing accurate MS images. The ground-truth MS demosaiced image is shown on the far right.}
    \label{fig:qual_sal}
    \vspace{-12pt}
\end{figure}

The calibration process extracts average patch intensities, resulting in two matrices: $\mathbf{A}\in\mathbb{R}^{140\times16}$, representing the multispectral values, and $\mathbf{B}\in\mathbb{R}^{140\times3}$, representing the RGB values.
The color conversion matrix $C\in\mathbb{R}^{16\times3}$, which maps 16-channel MS to 3-channel RGB values, is computed using least-squares optimization:
\begin{equation}
\vspace{-2pt}
C = \underset{C}{\text{argmin}} \| \mathbf{A} \times C - \mathbf{B} \|^2.
\label{eq:cc}
\vspace{-4pt}
\end{equation}
The resulting conversion matrix $C$ is crucial in cross-spectral disparity estimation (Eq.~(3) in the main paper), where it transforms the MS image into the RGB color space before disparity estimation, enabling geometric alignment between MS and RGB images in the proposed framework.

\section{Analysis on MS Demosaicing Framework}
\subsection{Generalization}
\begin{figure}
    \centering
    \includegraphics[width=1.00\linewidth]{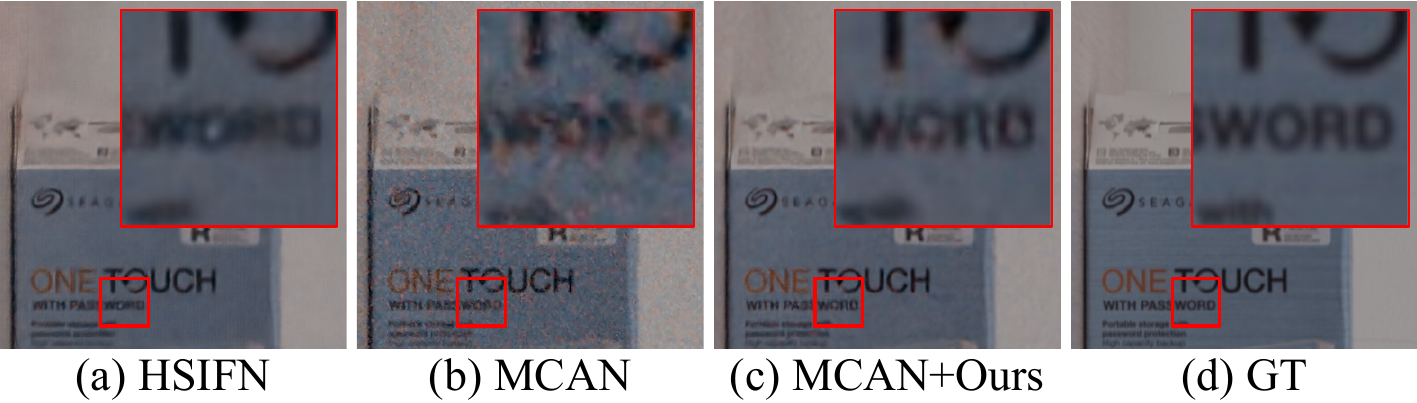}
    \vspace{-6mm}
    \caption{Qualitative comparison of RGB-guided MS demosaicing on the HS dataset~\cite{laiunaligned}. Trained solely on our dataset, MCAN + Ours recovers sharper spatial details and cleaner spectral content than MCAN and HSIFN, demonstrating the generalization capability of our method across varying MS imaging conditions.}
    \label{fig:exp_reb}
    \vspace{-12pt}
\end{figure}

While our method is primarily trained and evaluated on our large scale dataset---due to the availability of paired RGB-MS mosaics with high-fidelity GTs---we also conducted cross-dataset experiments on the HS dataset~\cite{laiunaligned}, which contains 60 HS-HS image pairs. Although it lacks raw mosaics, we simulate MS and RGB mosaics by projecting the demosaiced HS images onto our MSFA and CFA spectral response functions. This enables compatibility but introduces a domain gap due to differences in sensor characteristics. Simulated MS/RGB images can deviate from real sensor responses due to limitations in HS data precision, spectral response calibration, sensor nonlinearities, and noise characteristics~\cite{arad2022ntire}. Despite these challenges, our model generalizes well: MCAN + Ours achieves 42.48dB PSNR (24.03M, 2.66T MACs), outperforming MCAN (39.40dB, 5.24M, 0.91T) and the high-capacity HSIFN (40.27dB, 90.21M, 18.79T), demonstrating the effectiveness of our fusion strategy. \cref{fig:exp_reb} illustrates the qualitative advantages of our method.

\subsection{Effect of Spectral Alignment Layer (\texorpdfstring{$\boldsymbol{SAL}$}{SAL})}
We analyze the effectiveness of the Spectral Alignment Layer ($SAL$) in enhancing MS image demosaicing by leveraging RGB guidance. Figure~\ref{fig:qual_sal} illustrates the contribution of RGB guidance to reconstructing a target MS region, comparing results with and without $SAL$. The top row shows results without the spectral alignment offsets $\Delta i$ (Eq.~(7) in the main paper), while the bottom row includes the proposed $SAL$ (the last row in the table). The two models provides multi-scale RGB features $f'^{RGB}_l$ to the fusion network $\mathcal{F}$ (Eq.~(5) in the main paper). The per-pixel contribution and the area of contribution indicate that $SAL$ enables more effective integration of RGB guidance, leading to improved MS demosaicing quality, as reflected in the sharper and more accurate output.

Table~\ref{tab:per_ch_lam} presents quantitative results in terms of the Diffusion Index (DI)~\cite{Gu2021LAM}, which measures the range of contributed RGB pixels during MS demosaicing. Higher DI scores indicate better utilization of RGB guidance. Results show that incorporating $SAL$ consistently improves DI across all MS channels, with an average gain of 2.138, demonstrating the importance of spectral alignment in achieving higher fidelity MS reconstructions.

\subsection{Optical Flow Visualization}

\begin{figure*}
    \centering
    \includegraphics[width=1.0\linewidth]{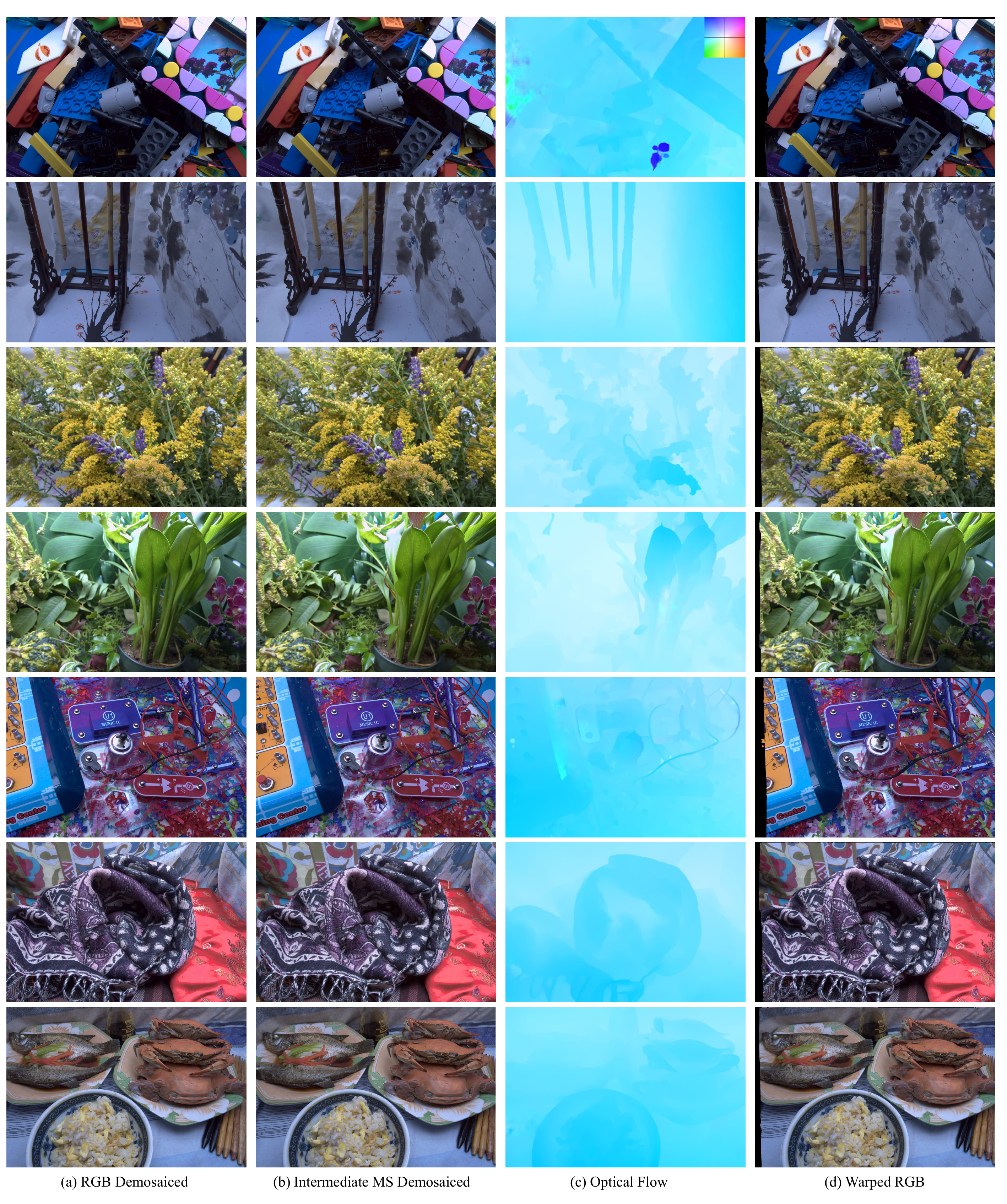}
    \vspace{-6mm}
    \caption{Visualization of optical flow and warping on various scenes in our test dataset. In the first two columns, we visualize the demosaiced RGB and the intermediate MS image (converted to the sRGB color space using the color conversion matrix $C$ (\cref{eq:cc}) and camera metadata, with CIE D65 as the reference white point), that are used to compute the flow. In the last two columns, we visualize the optical flow and the RGB image backwards warped into alignment with the MS image. Note that we visualize the warped RGB image, but in practice our $SAL$ warps \textit{features} across multiple scales.}
    \label{fig:flow}
    \vspace{-12pt}
\end{figure*}

The scenes in our dataset have objects at various depths and thus we utilize optical flow to perform alignment within our cross-spectral disparity estimation module. In Figure~\ref{fig:flow}, we visualize the optical flow field that warps the RGB features into alignment with the intermediate MS features. We visualize the intermediate MS image and demosaiced RGB image and the optical flow computed. The warped RGB visualization is a simplification of our $SAL$, which warps features across multiple scales from the RGB image. The optical flow between all images should be in the same direction since the cameras are separated by a fixed distance. However, the magnitude of the shift varies with scene depth--closer objects exhibit larger flow values, while farther objects have smaller flow values.

\subsection{RGB to MS Reconstruction}

In our experiments (Sec.~5 of the main paper), we demonstrate the effectiveness of using the RGB mosaic as guidance for MS restoration. However, a natural question arises: can we directly reconstruct the MS image from the RGB mosaic alone, bypassing the need for the MS mosaic?
To verify our approach, we assess our model against the earlier RGB-to-HS image reconstruction model, MSTPP$_{1\!\times}^{\text{rgb2ms}}$~\cite{mstpp}, which is adapted to process RGB mosaic images and produce demosaiced MS images for the 1$\times$ MS demosaicing task (\ie, Scenario 1 in Sec.~5.2 of the main paper).

For our method, we prepare three model variants. The first variant, NAFNet$_{1\!\times}^{\text{rgb2ms}}$, uses NAFNet~\cite{Chen2022NAFNet} as the backbone MS restoration network $\mathcal{D}_{MS}$. It takes only the RGB mosaic image $I^{RGB}_{2\times2}$ as input and directly produces the MS demosaiced image $I^{MS}$ as output. The second variant, NAFNet$_{1\!\times}$, also uses NAFNet as the backbone but instead takes the MS mosaic $I^{MS}_{4\times4}$ as input to reconstruct $I^{MS}$. The third variant, NAFNet$_{1\!\times}$ + Ours, integrates our proposed modules to leverage RGB guidance for MS demosaicing. These modules include the RGB demosaicing network $\mathcal{D}_{RGB}$ (Eq.~(2) in the main paper) and the cross-spectral fusion module (Sec.~3.2 of the main paper). 

For training,
MSTPP$_{1\!\times}^{\text{rgb2ms}}$ and our first model variant, NAFNet$_{1\!\times}^{\text{rgb2ms}}$, are trained using RGB mosaic images $I^{RGB}_{2\times2}$, whereas our second model variant, NAFNet$_{1\!\times}$, is trained to handle MS mosaic images $I^{MS}_{4\times4}$. The training process employs L2 loss between predicted MS demosaic images with the ground-truth $\hat{I}^{MS}$ (Eq.~(8) of the main paper).
Note that the $I^{RGB}_{2\times2}$ images are geometrically aligned with ground-truth MS images, as they are captured under CIE D65 daylight illumination from the same camera position as $\hat{I}^{MS}$. The third variant, NAFNet$_{1\!\times}$ + Ours, follows the full training pipeline outlined in Section 3.3 of the main paper, which incorporates RGB guidance through fusion to improve MS restoration.

{
\setlength{\aboverulesep}{0mm}
\setlength{\belowrulesep}{0mm}
\renewcommand{\arraystretch}{0.98} 
\setlength{\tabcolsep}{0pt} 

\begin{table}[t]
\centering
\small
\begin{tabular}{
>{\centering}p{0.22\linewidth} 
>{\centering}p{0.30\linewidth} 
>{\centering}p{0.33\linewidth} 
>{\centering\arraybackslash}p{0.15\linewidth} 
}

\toprule
\multirow{2}{*}[-0.5\dimexpr \aboverulesep + \belowrulesep + \cmidrulewidth]{\makecell{Target\\MS channel}} & 

\multicolumn{2}{c}{DI~\cite{Gu2021LAM} w.r.t. target MS channel ($\uparrow$)} &
\multirow{2}{*}[-0.5\dimexpr \aboverulesep + \belowrulesep + \cmidrulewidth]{Gain} \\
\cmidrule(lr){2-3}

&$SAL$ w/o $\Delta i$ & 
$SAL$ & \\
\midrule


1 &  9.035 & \textbf{11.22} & 2.185 \\
2 &  8.500 &  \textbf{9.976} & 1.476 \\
3 &  8.962 &  \textbf{8.964} & 0.002 \\
4 &  8.606 &  \textbf{9.753} & 1.146 \\
5 &  8.757 & \textbf{10.63} & 1.874 \\
6 &  7.388 &  \textbf{9.518} & 2.130 \\
7 &  7.296 &  \textbf{9.103} & 1.807 \\
8 &  10.04 & \textbf{10.12}  & 0.088 \\
9 &  8.203 &  \textbf{9.861} & 1.658 \\
10 &  8.728 &  \textbf{9.989} & 1.260 \\
11 &  9.215 & \textbf{12.31} & 3.095 \\
12 &  8.049 &  \textbf{9.033} & 0.984 \\
13 &  7.823 &  \textbf{8.094} & 0.270 \\
14 &  7.208 &  \textbf{8.310} & 1.102 \\
15 &  8.012 &  \textbf{9.837} & 1.825 \\
16 &  8.487 & \textbf{10.77} & 2.283 \\


\midrule
\midrule

\multirow{2}{*}[-0.5\dimexpr \aboverulesep + \belowrulesep + \cmidrulewidth]{\makecell{DI w.r.t all\\MS channels}} &
\multirow{2}{*}[-0.5\dimexpr \aboverulesep + \belowrulesep + \cmidrulewidth]{\makecell{7.373}} &
\multirow{2}{*}[-0.5\dimexpr \aboverulesep + \belowrulesep + \cmidrulewidth]{\makecell{\textbf{9.511}}} &
\multirow{2}{*}[-0.5\dimexpr \aboverulesep + \belowrulesep + \cmidrulewidth]{\makecell{2.138}} \\
&&&\\


\arrayrulecolor{black}
\bottomrule
\end{tabular}
\vspace{-6pt}
\caption{Quantitative results for 1$\times$ MS demosaicing, showing the effect of $SAL$ in terms of the Diffusion Index (DI)~\cite{Gu2021LAM}. The DI measures the range of involved RGB pixels during MS restoration. The reported values represent the average Diffusion Index (DI), calculated by selecting two random target regions from each of the 103 test images in the proposed RGB-MS dataset, resulting in a total of 206 target regions.}
\label{tab:per_ch_lam}
\vspace{-12pt}
\end{table}
}

Table~\ref{tab:rgb_to_ms} summarizes the results. The table highlights the clear advantage of using the MS mosaic input for MS reconstruction (first and second vs. third rows of the table). Moreover, providing RGB guidance with our proposed modules achieves the best MS restoration performance (fourth row), as the high-fidelity details from the RGB mosaic are effectively fused during MS restoration.

We also compare MSTPP$_{1\!\times}^{\text{rgb2ms}}$ and NAFNet$_{1\!\times}^{\text{rgb2ms}}$ with NAFSR$_{4\!\times}$~\cite{chu2022nafssr}, a variant of NAFNet designed for super-resolution and trained for the 4$\times$ MS demosaicing task, which reconstructs MS images from 4$\times$ downsampled MS mosaics (Scenario 2 in Sec.~5.2 of the main paper). As expected, MSTPP$_{1\!\times}^{\text{rgb2ms}}$ and NAFNet$_{1\!\times}^{\text{rgb2ms}}$ outperform NAFSR$_{4\!\times}$ across all metrics, as the latter relies on lower-resolution inputs. However, when NAFSR$_{4\!\times}$ is combined with our proposed modules (last row of Table~\ref{tab:rgb_to_ms}), it achieves competitive PSNR and SSIM scores compared to NAFNet$_{1\!\times}^{\text{rgb2ms}}$, while significantly improving the SAM score, which quantifies spectral fidelity. This demonstrates the efficacy of the proposed RGB-guided MS restoration scheme, even in challenging super-resolution settings.

{
\setlength{\aboverulesep}{0mm}
\setlength{\belowrulesep}{0mm}
\renewcommand{\arraystretch}{1.0} 
\setlength{\tabcolsep}{0pt} 

\begin{table}[t]
\centering
\small
\begin{tabular}{
p{0.35\linewidth} 
>{\centering}p{0.13\linewidth} 
>{\centering}p{0.13\linewidth} 
>{\centering}p{0.13\linewidth} 
>{\centering}p{0.00\linewidth} 
>{\centering}p{0.13\linewidth} 
>{\centering\arraybackslash}p{0.13\linewidth} 
}

\toprule
\multirow{2}{*}[-0.5\dimexpr \aboverulesep + \belowrulesep + \cmidrulewidth]{\makecell{Model}} & 
\multirow{2}{*}[-0.5\dimexpr \aboverulesep + \belowrulesep + \cmidrulewidth]{\makecell{PSNR$\uparrow$}} & 
\multirow{2}{*}[-0.5\dimexpr \aboverulesep + \belowrulesep + \cmidrulewidth]{\makecell{SSIM$\uparrow$}} &
\multirow{2}{*}[-0.5\dimexpr \aboverulesep + \belowrulesep + \cmidrulewidth]{\makecell{SAM$\downarrow$}} &&
Params (MB) &
MACs\footnotemark[1] (T) \\

\midrule
MSTPP$^{\text{rgb2ms}}_{1\times}$~\cite{mstpp} & 37.96 & 0.9746 & 4.430 && 85.81 & 12.36\\

NAFNet$_{1\!\times}^{\text{rgb2ms}}$ & 37.94 & 0.9734 & 4.370 && 111.25 & 0.78\\
\midrule
\midrule
NAFNet$_{1\!\times}$~\cite{Chen2022NAFNet} & 40.89 & 0.9766 &2.604 && 111.25 & 0.78\\
\rowcolor{lightlightgray}
NAFNet$_{1\!\times}$ + Ours & \textbf{41.92} & \textbf{0.9811} & \textbf{2.422}&& 130.03 & 2.53\\
\midrule
\midrule
NAFSR$_{4\!\times}$~\cite{chu2022nafssr} & 32.98 &  0.9173 & 4.736 && 59.19 & 2.66 \\
\rowcolor{lightlightgray}
NAFSR$_{4\!\times}$ + Ours & 37.67 & 0.9641 & 3.576 && 77.98 & 4.41\\

\arrayrulecolor{black}
\bottomrule
\end{tabular}
\vspace{-6pt}
\caption{Quantitative comparison for MS demosaicing.}
\label{tab:rgb_to_ms}
\vspace{-12pt}
\end{table}
}

\subsection{RGB Demosaicing using MS Reference}
While the primary focus of this work is on leveraging RGB guidance for MS restoration tasks in dual-camera setups with RGB and MS sensors, the complementary nature of the MS sensor motivates exploring the inverse scenario: leveraging MS guidance to enhance RGB demosaicing. Despite the lower spatial fidelity of MS mosaics due to their inherent low-resolution nature, they can potentially capture spectral details that cannot be captured by the RGB CFA sensor~\cite{shen2015multispectral}. Furthermore, MS sensor provide richer spectral diversity, which can be utilized during RGB demosaicing tasks for reconstructing accurate colors~\cite{niu2023nir, JDM-HDRNet_ECCV2024}.
Our proposed framework and dual-camera RGB-MS dataset are well-suited for extending to this task, demonstrating their flexibility in addressing different restoration scenarios.

To validate this idea, we adapt our proposed MS demosaicing framework by prioritizing RGB restoration in the fusion stage (Sec.~3.2 of the main paper).
Specifically, the demosaiced RGB image $I'^{RGB}$ is used as the primary input to the fusion network $\mathcal{F}$, while the intermediate multi-scale MS feature map $f'^{MS}_l$ is refined by the spectral alignment layer ($SAL$) and provided as auxiliary MS guidance to $\mathcal{F}$. This adjustment allows the fusion network to generate enhanced RGB demosaiced image $I^{RGB}$ by leveraging the spectral diversity of the MS features.

We evaluate the effectiveness of MS-guided RGB demosaicing by comparing three model variants. The baseline model, NAFNet~\cite{Chen2022NAFNet}, processes only RGB mosaic images $I^{RGB}_{2\times2}$ without MS guidance. To ensure a fair comparison, we also evaluate a capacity-increased version, NAFNet-L. Finally, the proposed method, NAFNet+Ours, incorporates MS guidance during the fusion stage, integrating $f'^{MS}_l$ to enhance RGB reconstruction.
All models are trained using the RGB demosaicing loss (Eq.~(9) in the main paper) on paired RGB mosaic images $I^{RGB}_{2\times2}$ and their corresponding ground-truth RGB demosaiced images $\hat{I}^{RGB}$.

\begin{figure}
    \centering
    \includegraphics[width=\linewidth]{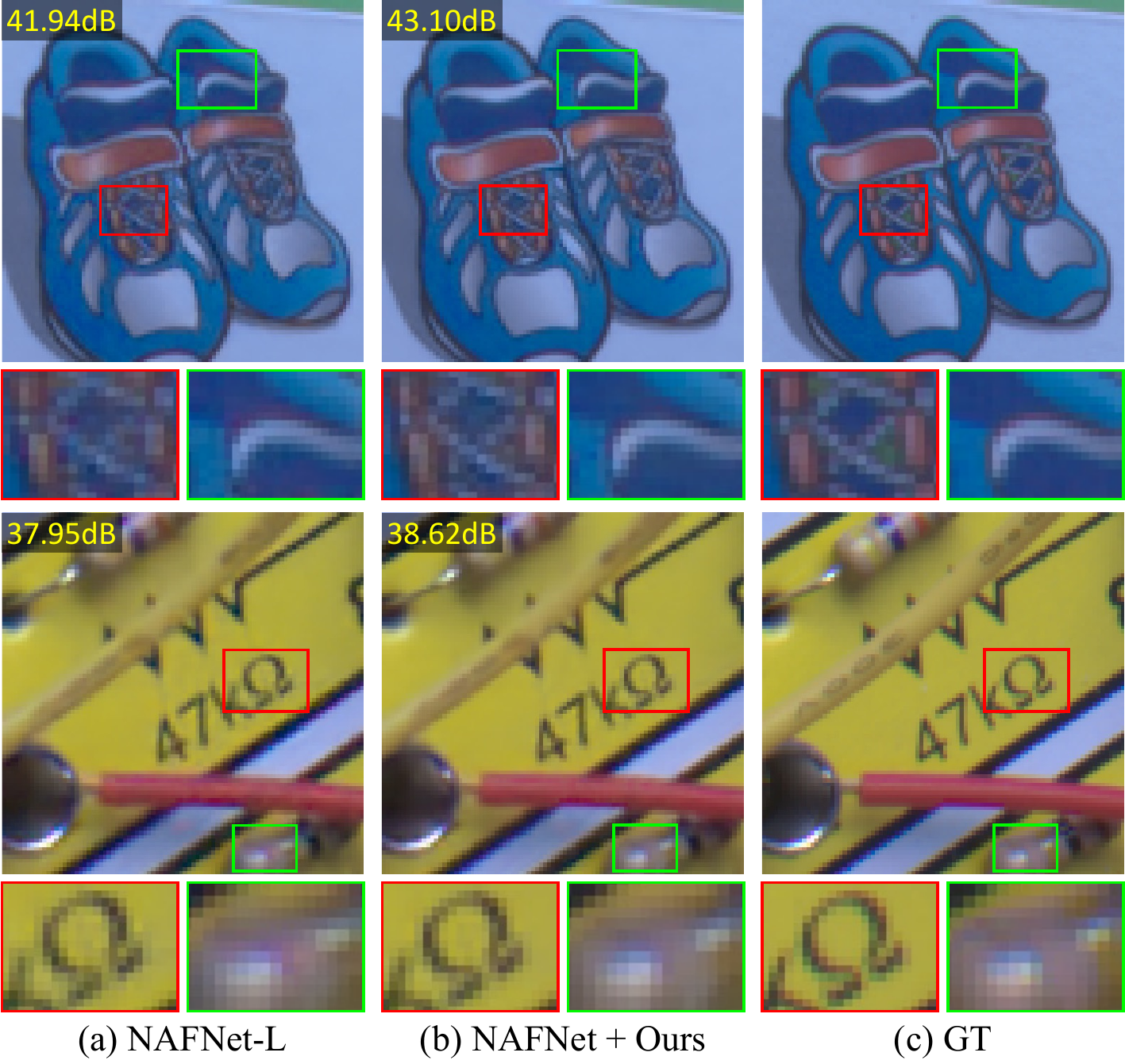}
    \vspace{-6mm}
    \caption{Qualitative comparison of RGB demosaicing: NAFNet-L processes only the RGB mosaic image as input, whereas NAFNet$+$Ours incorporates MS features as guidance during demosaicing. The zoomed-in cropped patches in the red and green boxes demonstrate the advantages of using MS guidance: enhanced detail (red box) and improved color accuracy (green box).}
    \label{fig:ms_help_rgb}
    \vspace{-12pt}
\end{figure}

Table~\ref{tab:ms_help_rgb} presents the quantitative results. Compared to NAFNet, NAFNet integrated with our modules achieves consistent improvement across all metrics (first vs. third rows of the table), demonstrating the benefit of incorporating MS guidance in RGB restoration. Compared to NAFNet-L, which benefits from increased model capacity, our method shows better performance, illustrating effectiveness of our method in utilizing MS guidance for enhancing RGB reconstruction quality.
Figure~\ref{fig:ms_help_rgb} further illustrates the benefits of the proposed method. The zoomed-in regions show how MS guidance contributes to improved detail recovery and color accuracy. Specifically, compared to NAFNet-L, our method recovers finer spatial details (red boxes in first vs. second columns) and addresses color inaccuracies caused by limited spectral diversity of RGB CFA sensor (green boxes), validating the effectiveness of MS-guided RGB demosaicing. 
Furthermore, the results highlight the flexibility of our dual-camera RGB-MS dataset and framework, indicating their potential to support both MS restoration and RGB reconstruction tasks.

\section{More Details and Discussion on Experiments}
In this section, we provide additional details comparing our model with the previous approaches (Sec.~5.2 of the main paper).

\subsection{Architectural Modification}
{
\setlength{\aboverulesep}{0mm}
\setlength{\belowrulesep}{0mm}
\renewcommand{\arraystretch}{1.0} 
\setlength{\tabcolsep}{0pt} 

\begin{table}[t]
\centering
\small
\begin{tabular}{
p{0.35\linewidth} 
>{\centering}p{0.13\linewidth} 
>{\centering}p{0.13\linewidth} 
>{\centering}p{0.13\linewidth} 
>{\centering}p{0.00\linewidth} 
>{\centering}p{0.13\linewidth} 
>{\centering\arraybackslash}p{0.13\linewidth} 
}

\toprule
\multirow{2}{*}[-0.5\dimexpr \aboverulesep + \belowrulesep + \cmidrulewidth]{\makecell{Model}} & 
\multirow{2}{*}[-0.5\dimexpr \aboverulesep + \belowrulesep + \cmidrulewidth]{\makecell{PSNR$\uparrow$}} & 
\multirow{2}{*}[-0.5\dimexpr \aboverulesep + \belowrulesep + \cmidrulewidth]{\makecell{SSIM$\uparrow$}} &
\multirow{2}{*}[-0.5\dimexpr \aboverulesep + \belowrulesep + \cmidrulewidth]{\makecell{SAM$\downarrow$}} &&
Params (MB) &
MACs\footnotemark[1] (T) \\

\midrule

NAFNet~\cite{Chen2022NAFNet} & 44.97 & 0.9844 & 1.566 && 111.23 & 0.77\\
\arrayrulecolor{gray}
NAFNet-L & 45.06 & 0.9847 &1.546 && 158.57 & 1.49\\
\rowcolor{lightlightgray}
NAFNet + Ours & \textbf{45.82} & \textbf{0.9874} & \textbf{1.450}&& 130.03 & 2.51\\

\arrayrulecolor{black}
\bottomrule
\end{tabular}
\vspace{-6pt}
\caption{Quantitative comparison for RGB demosaicing}
\label{tab:ms_help_rgb}
\end{table}
}

\paragraph{Adapting Alignment}
For both DCT~\cite{dct} and HSIFN~\cite{laiunaligned}, we perform alignment to ensure a fair comparison. DCT requires alignment, so we use our pre-alignment module to provide an aligned RGB image alongside the MS mosaic. HSIFN includes its own alignment network, which requires two RGB images; we use this network but supply RGB inputs from our pre-alignment module.

\vspace{-12pt}
\paragraph{Upsampling Module}  
For Scenario 2 of the Sec.~5.2 of the main paper, which aims for the 4$\times$ MS demosaicing task, the baseline MS demosaicing networks $\mathcal{D}_{MS}$ are modified to reconstruct high-resolution MS images from low-resolution MS mosaics.
For NAFSR~\cite{Chen2022NAFNet}, we adapt NAFSSR~\cite{chu2022nafssr}, originally designed for stereo super-resolution, by removing its stereo-specific cross-attention modules to accommodate the single mosaic input in our task. For MCAN and Restormer, we replace the final convolution layer, which produces a demosaiced image, with a feature extraction layer, followed by an upsampling module that generates high-resolution MS demosaiced images. The upsampling module comprises convolutional and pixel-shuffle~\cite{shi2016pixelshuffel} layers, with its architecture detailed in Table~\ref{tab:upsampling_module}.

\vspace{-14pt}
\paragraph{Discussion on HSIFN} While both our method and HSIFN~\cite{laiunaligned} use color mapping and optical flow for alignment, they differ in fusion strategy and backbone design. HSIFN applies spatial attention between warped RGB and RGB-mapped HS features prior to decoding, whereas our method adopts channel attention within the decoder~\cite{Chen2022NAFNet}, which is well suited for restoration. For alignment, we use deformable convolutions, which outperform the direct warping approach used in HSIFN (Table 1). Additionally, our model is more efficient in terms of computational cost (2.53T vs.~18.79T; Table 2).




\section{Additional Qualitative Results}

\begin{figure*}
    \centering
    \includegraphics[width=1.0\textwidth]{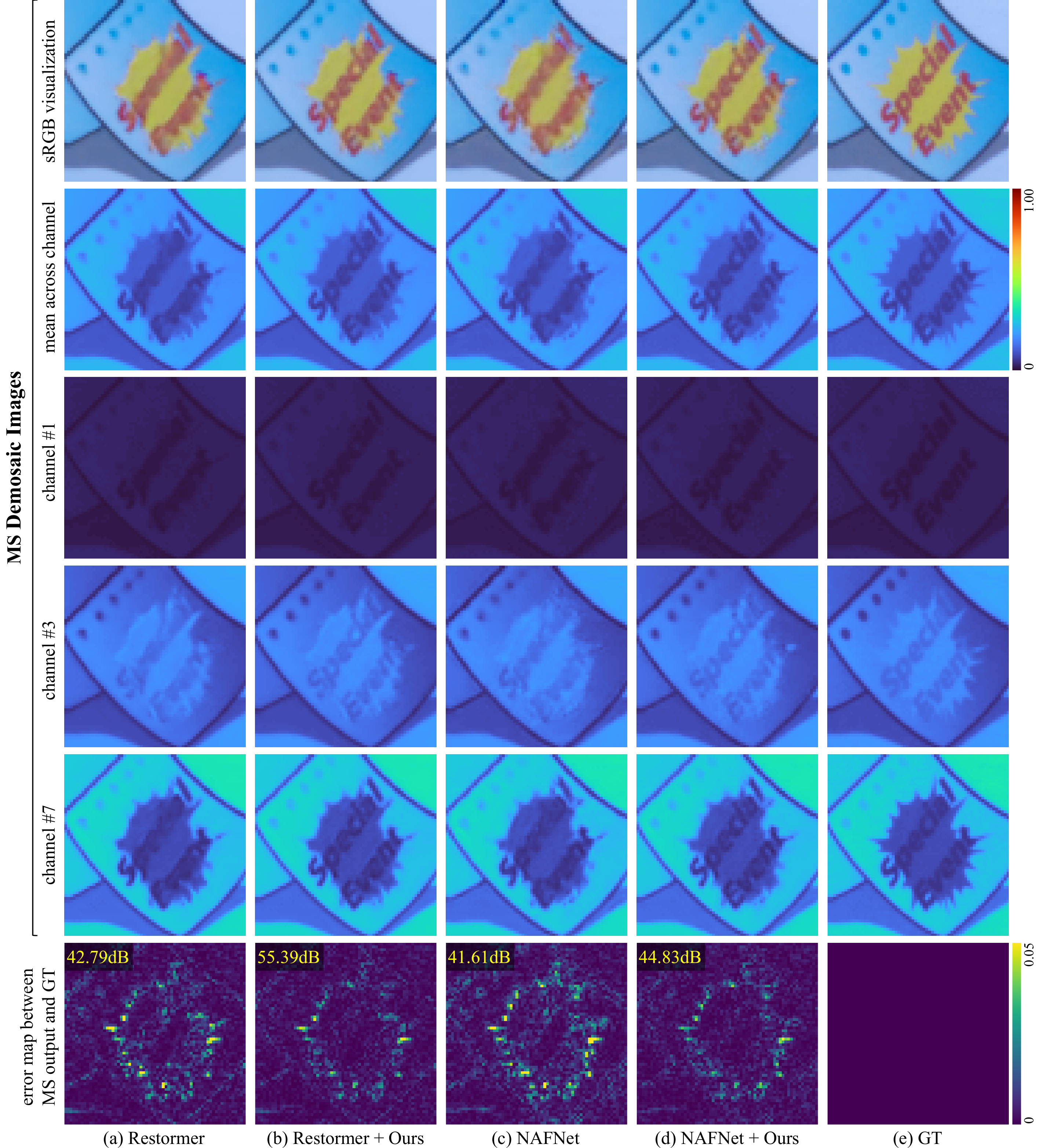}
    \vspace{-6mm}
    \caption{\textbf{Qualitative comparison of 1$\times$ MS demosaicing results} for a dual-camera scenario featuring MS and RGB sensors with the same spatial resolution but employing asymmetric CFAs. The top row shows the predicted MS demosaics converted to the sRGB color space using the color conversion matrix $C$ (\cref{eq:cc}) and camera metadata, with CIE D65 as the reference white point. The second row presents the MS demosaic output averaged across the channel dimension, while the third to fifth rows display per-channel MS demosaic outputs for the 1st, 3rd, and 7th channel indices, respectively. The final row visualizes the error maps between the restored and ground-truth MS demosaiced images.}
    \label{fig:1xresults10}
    \vspace{-4mm}
\end{figure*}
\begin{figure*}
    \centering
    \includegraphics[width=1.0\textwidth]{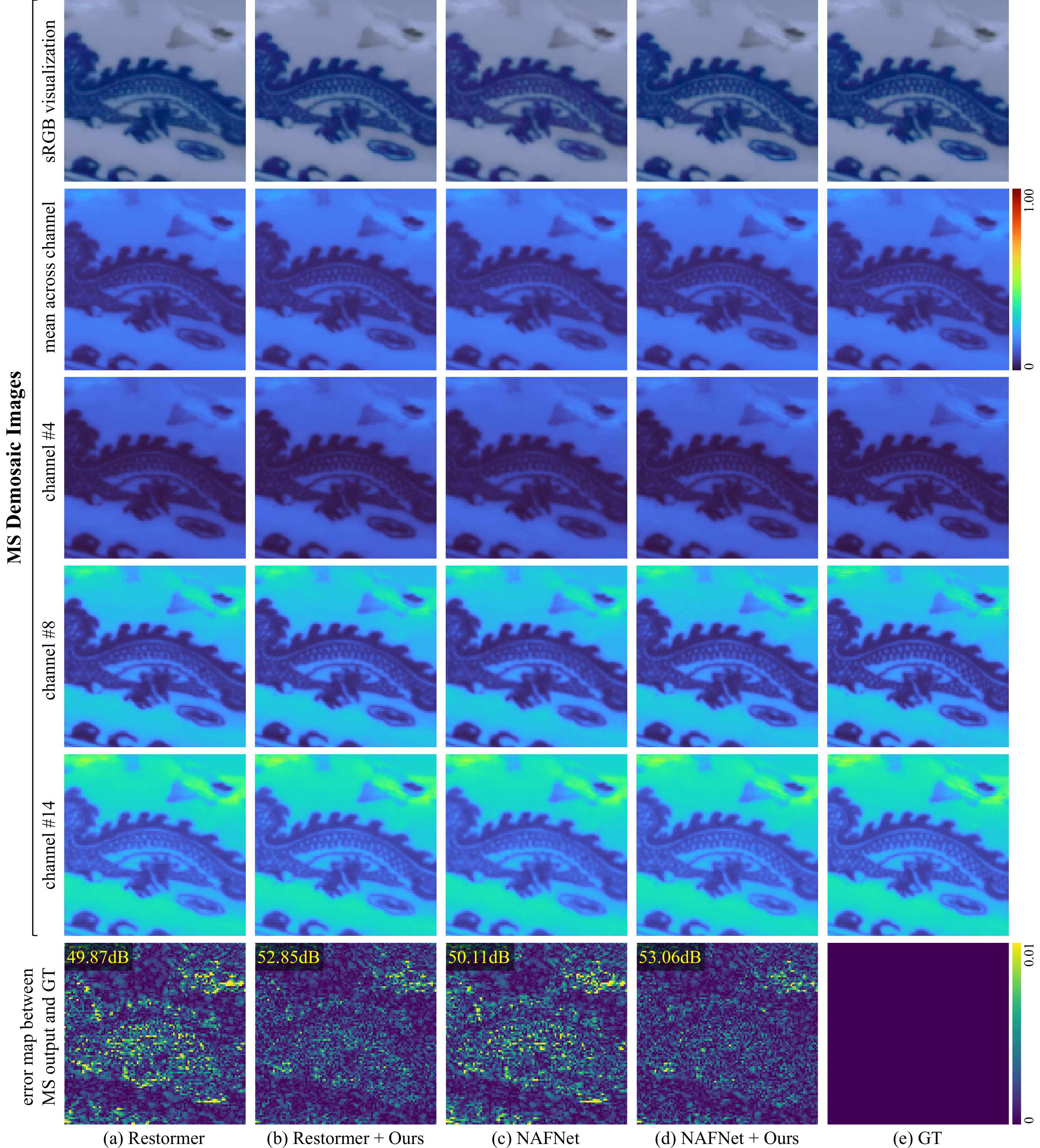}
    \vspace{-6mm}
    \caption{\textbf{Qualitative comparison of 1$\times$ MS demosaicing results} for a dual-camera scenario featuring MS and RGB sensors with the same spatial resolution but employing asymmetric CFAs. The top row shows the predicted MS demosaics converted to the sRGB color space using the color conversion matrix $C$ (\cref{eq:cc}) and camera metadata, with CIE D65 as the reference white point. The second row presents the MS demosaic output averaged across the channel dimension, while the third to fifth rows display per-channel MS demosaic outputs for the 4th, 8th, and 14th channel indices, respectively. The final row visualizes the error maps between the restored and ground-truth MS demosaiced images.}
    \label{fig:1xresults1}
    \vspace{-4mm}
\end{figure*}
\begin{figure*}
    \centering
    \includegraphics[width=1.0\textwidth]{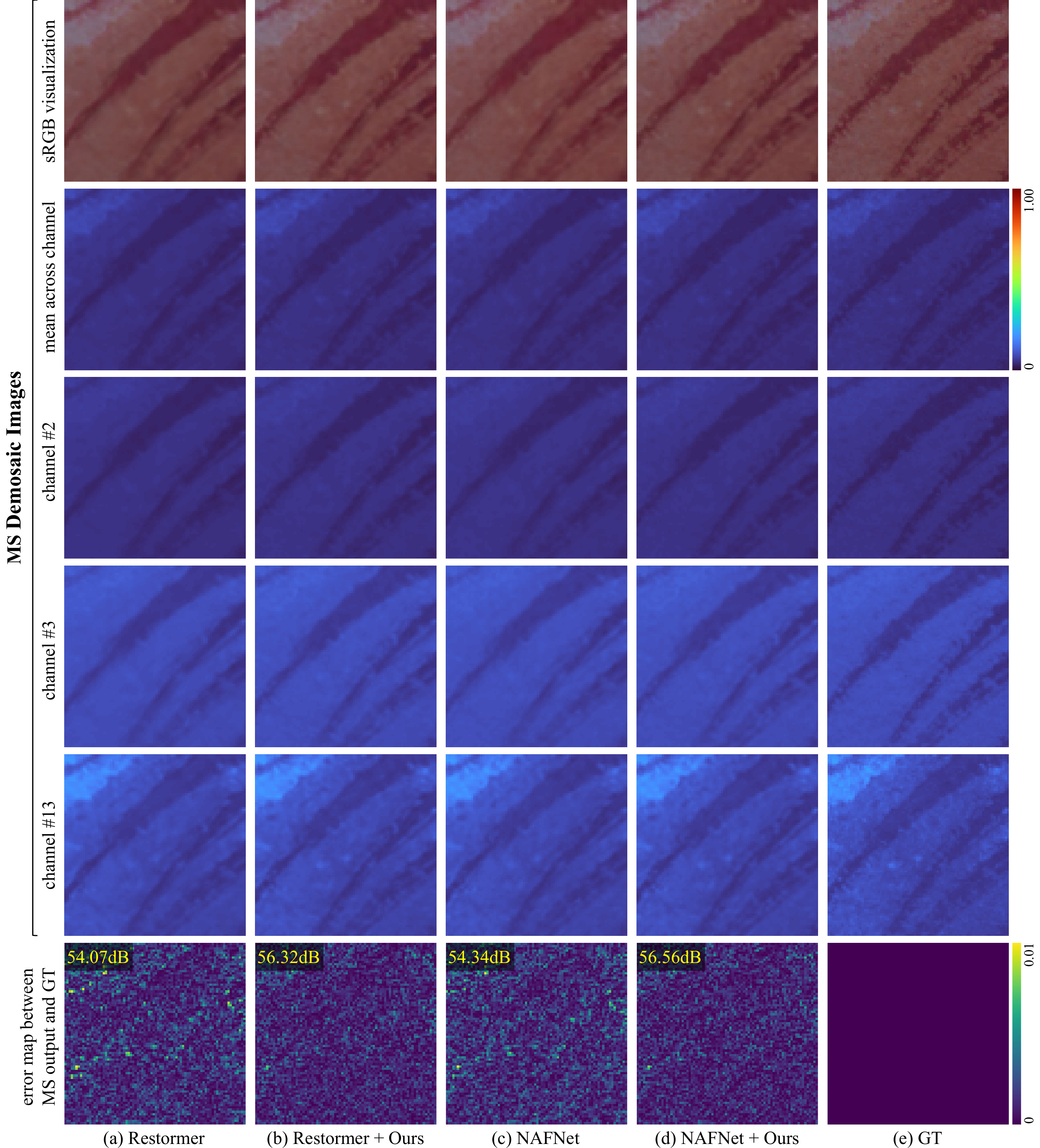}
    \vspace{-6mm}
    \caption{\textbf{Qualitative comparison of 1$\times$ MS demosaicing results} for a dual-camera scenario featuring MS and RGB sensors with the same spatial resolution but employing asymmetric CFAs. The top row shows the predicted MS demosaics converted to the sRGB color space using the color conversion matrix $C$ (\cref{eq:cc}) and camera metadata, with CIE D65 as the reference white point. The second row presents the MS demosaic output averaged across the channel dimension, while the third to fifth rows display per-channel MS demosaic outputs for the 2nd, 3rd, and 13th channel indices, respectively. The final row visualizes the error maps between the restored and ground-truth MS demosaiced images.}
    \label{fig:1xresults5}
    \vspace{-4mm}
\end{figure*}
\begin{figure*}
    \centering
    \includegraphics[width=1.0\textwidth]{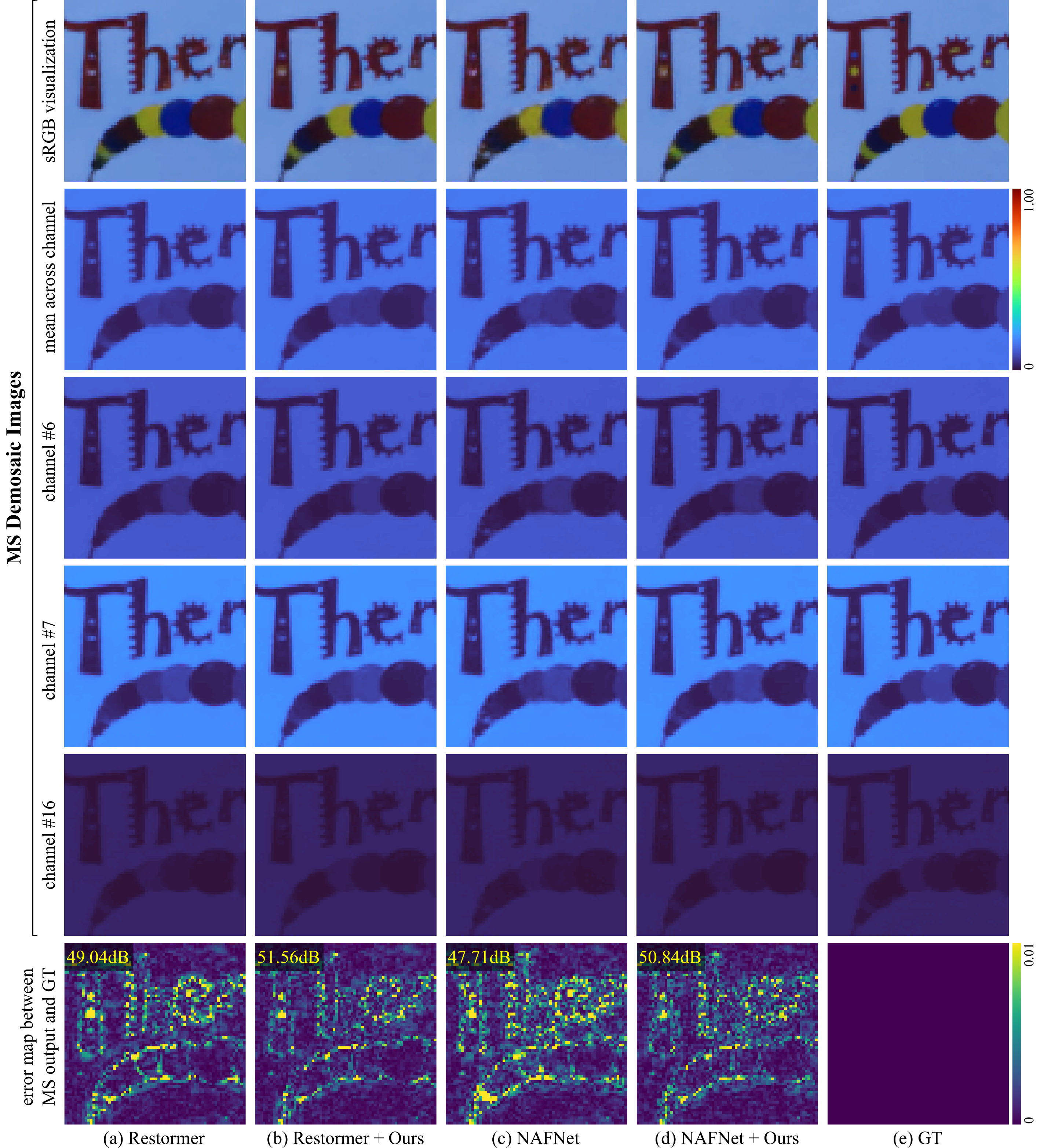}
    \vspace{-6mm}
    \caption{\textbf{Qualitative comparison of 1$\times$ MS demosaicing results} for a dual-camera scenario featuring MS and RGB sensors with the same spatial resolution but employing asymmetric CFAs. The top row shows the predicted MS demosaics converted to the sRGB color space using the color conversion matrix $C$ (\cref{eq:cc}) and camera metadata, with CIE D65 as the reference white point. The second row presents the MS demosaic output averaged across the channel dimension, while the third to fifth rows display per-channel MS demosaic outputs for the 6th, 7th, and 16th channel indices, respectively. The final row visualizes the error maps between the restored and ground-truth MS demosaiced images.}
    \label{fig:1xresults2}
    \vspace{-4mm}
\end{figure*}
\begin{figure*}
    \centering
    \includegraphics[width=1.0\textwidth]{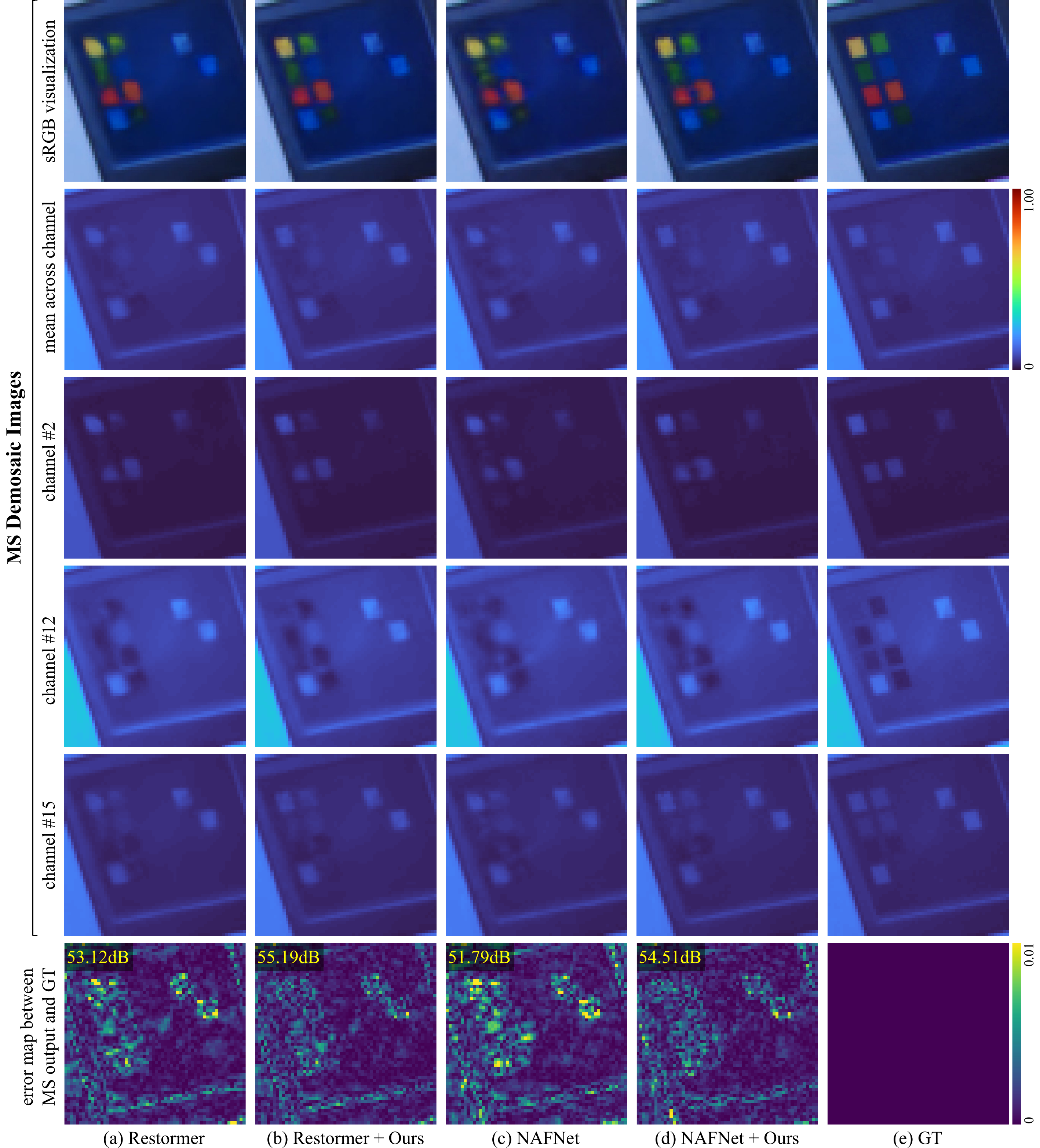}
    \vspace{-6mm}
    \caption{\textbf{Qualitative comparison of 1$\times$ MS demosaicing results} for a dual-camera scenario featuring MS and RGB sensors with the same spatial resolution but employing asymmetric CFAs. The top row shows the predicted MS demosaics converted to the sRGB color space using the color conversion matrix $C$ (\cref{eq:cc}) and camera metadata, with CIE D65 as the reference white point. The second row presents the MS demosaic output averaged across the channel dimension, while the third to fifth rows display per-channel MS demosaic outputs for the 2nd, 12th, and 15th channel indices, respectively. The final row visualizes the error maps between the restored and ground-truth MS demosaiced images.}
    \label{fig:1xresults3}
    \vspace{-4mm}
\end{figure*}
\begin{figure*}
    \centering
    \includegraphics[width=1.0\textwidth]{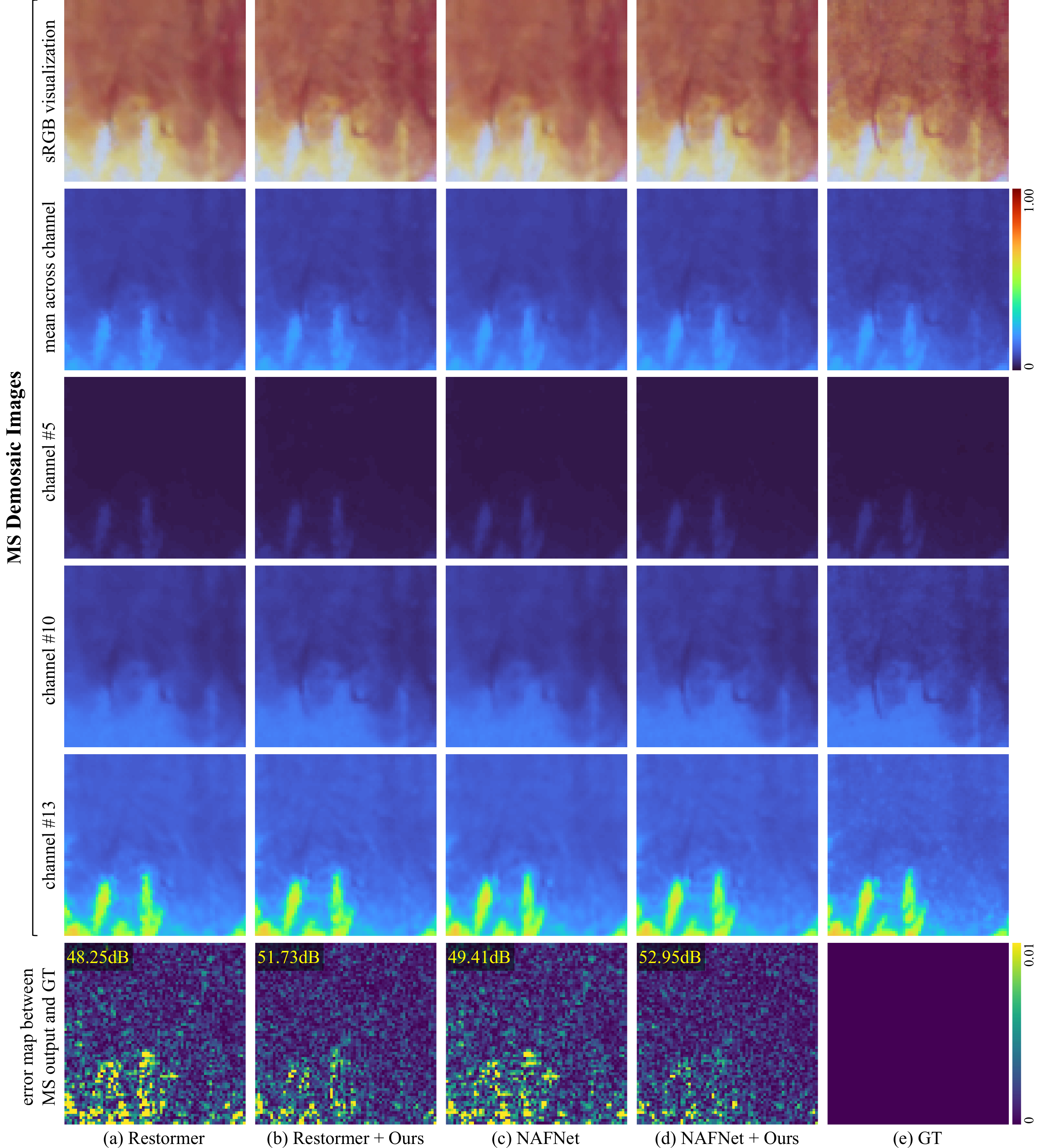}
    \vspace{-6mm}
    \caption{\textbf{Qualitative comparison of 1$\times$ MS demosaicing results} for a dual-camera scenario featuring MS and RGB sensors with the same spatial resolution but employing asymmetric CFAs. The top row shows the predicted MS demosaics converted to the sRGB color space using the color conversion matrix $C$ (\cref{eq:cc}) and camera metadata, with CIE D65 as the reference white point. The second row presents the MS demosaic output averaged across the channel dimension, while the third to fifth rows display per-channel MS demosaic outputs for the 5th, 10th, and 13th channel indices, respectively. The final row visualizes the error maps between the restored and ground-truth MS demosaiced images.}
    \label{fig:1xresults4}
    \vspace{-4mm}
\end{figure*}
\begin{figure*}
    \centering
    \includegraphics[width=1.0\textwidth]{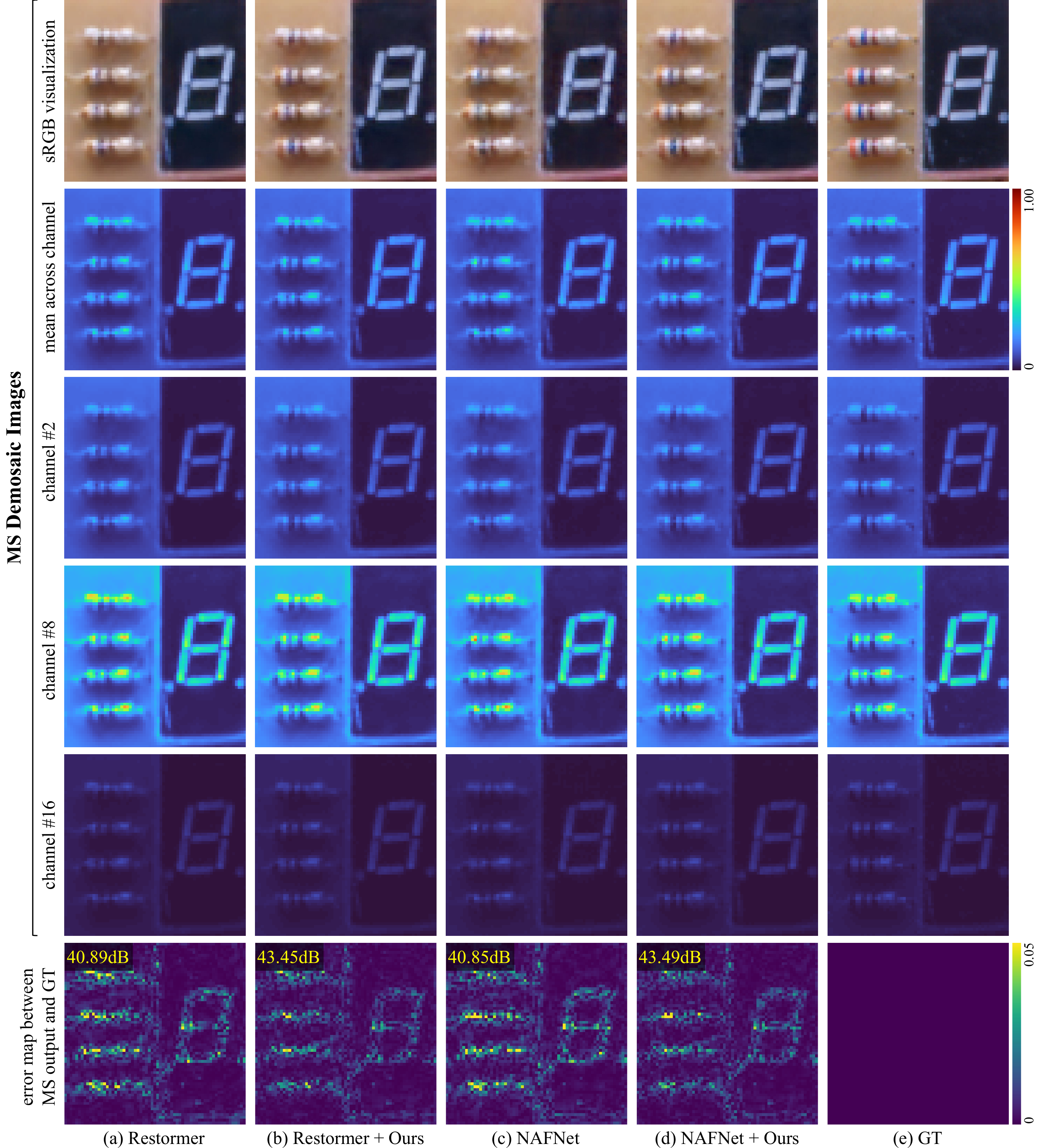}
    \vspace{-6mm}
    \caption{\textbf{Qualitative comparison of 1$\times$ MS demosaicing results} for a dual-camera scenario featuring MS and RGB sensors with the same spatial resolution but employing asymmetric CFAs. The top row shows the predicted MS demosaics converted to the sRGB color space using the color conversion matrix $C$ (\cref{eq:cc}) and camera metadata, with CIE D65 as the reference white point. The second row presents the MS demosaic output averaged across the channel dimension, while the third to fifth rows display per-channel MS demosaic outputs for the 2nd, 8th, and 16th channel indices, respectively. The final row visualizes the error maps between the restored and ground-truth MS demosaiced images.}
    \label{fig:1xresults7}
    \vspace{-4mm}
\end{figure*}
\begin{figure*}
    \centering
    \includegraphics[width=1.0\textwidth]{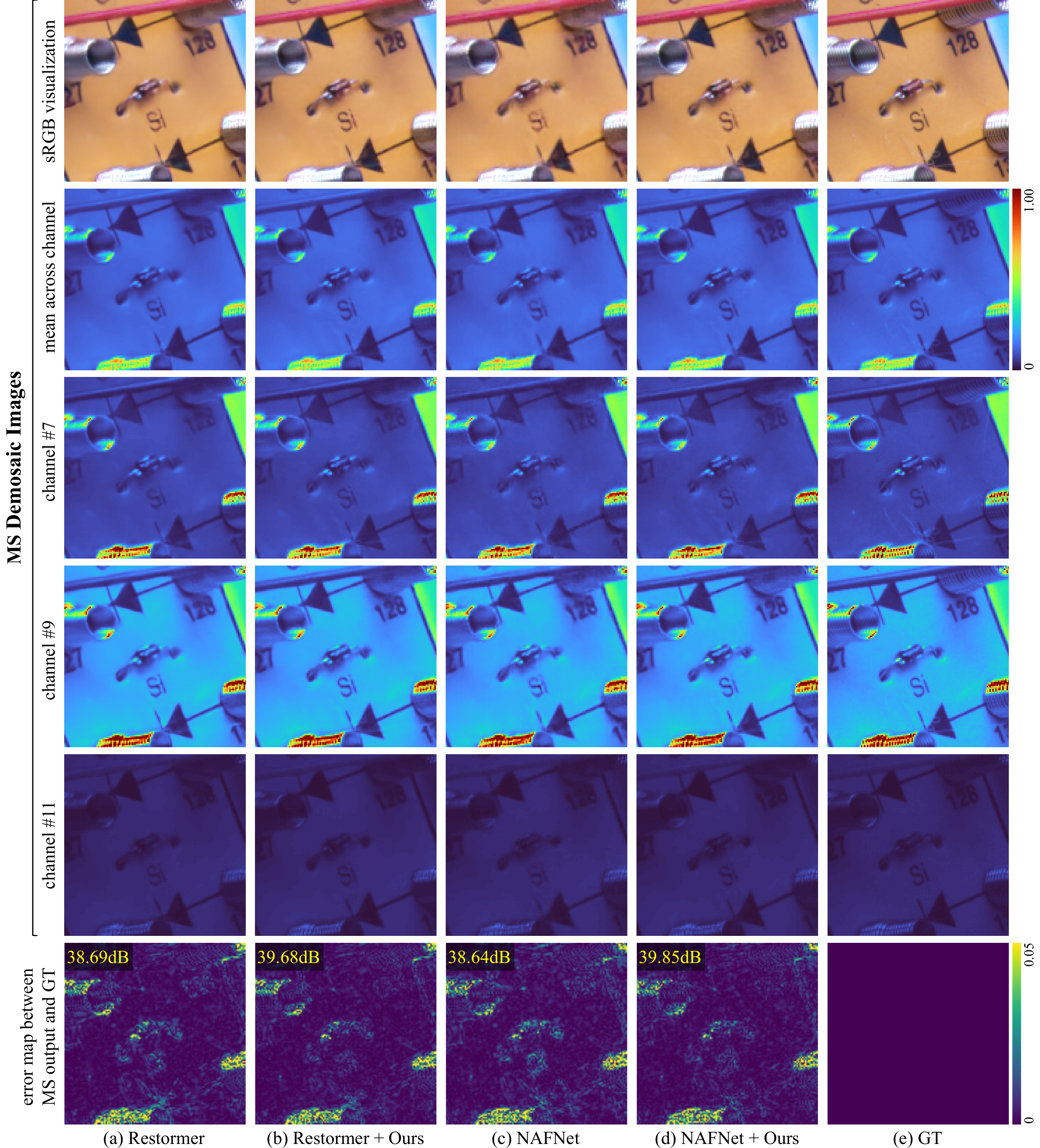}
    \vspace{-6mm}
    \caption{\textbf{Qualitative comparison of 1$\times$ MS demosaicing results} for a dual-camera scenario featuring MS and RGB sensors with the same spatial resolution but employing asymmetric CFAs. The top row shows the predicted MS demosaics converted to the sRGB color space using the color conversion matrix $C$ (\cref{eq:cc}) and camera metadata, with CIE D65 as the reference white point. The second row presents the MS demosaic output averaged across the channel dimension, while the third to fifth rows display per-channel MS demosaic outputs for the 7th, 9th, and 11th channel indices, respectively. The final row visualizes the error maps between the restored and ground-truth MS demosaiced images.}
    \label{fig:1xresults8}
    \vspace{-4mm}
\end{figure*}
\begin{figure*}
    \centering
    \includegraphics[width=1.0\textwidth]{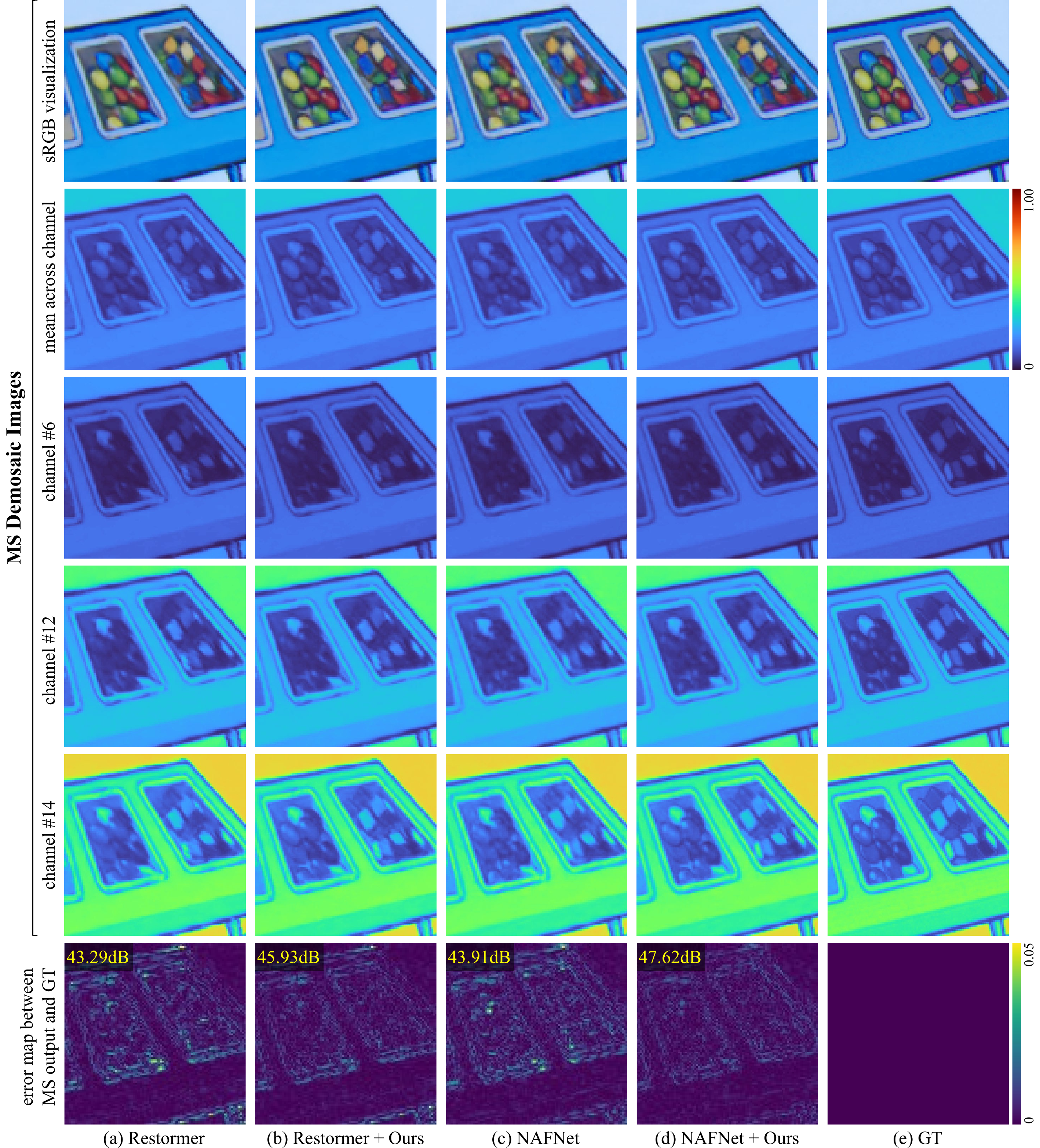}
    \vspace{-6mm}
    \caption{\textbf{Qualitative comparison of 1$\times$ MS demosaicing results} for a dual-camera scenario featuring MS and RGB sensors with the same spatial resolution but employing asymmetric CFAs. The top row shows the predicted MS demosaics converted to the sRGB color space using the color conversion matrix $C$ (\cref{eq:cc}) and camera metadata, with CIE D65 as the reference white point. The second row presents the MS demosaic output averaged across the channel dimension, while the third to fifth rows display per-channel MS demosaic outputs for the 6th, 12th, and 14th channel indices, respectively. The final row visualizes the error maps between the restored and ground-truth MS demosaiced images.}
    \label{fig:1xresults9}
    \vspace{-4mm}
\end{figure*}
\begin{figure*}
    \centering
    \includegraphics[width=1.0\textwidth]{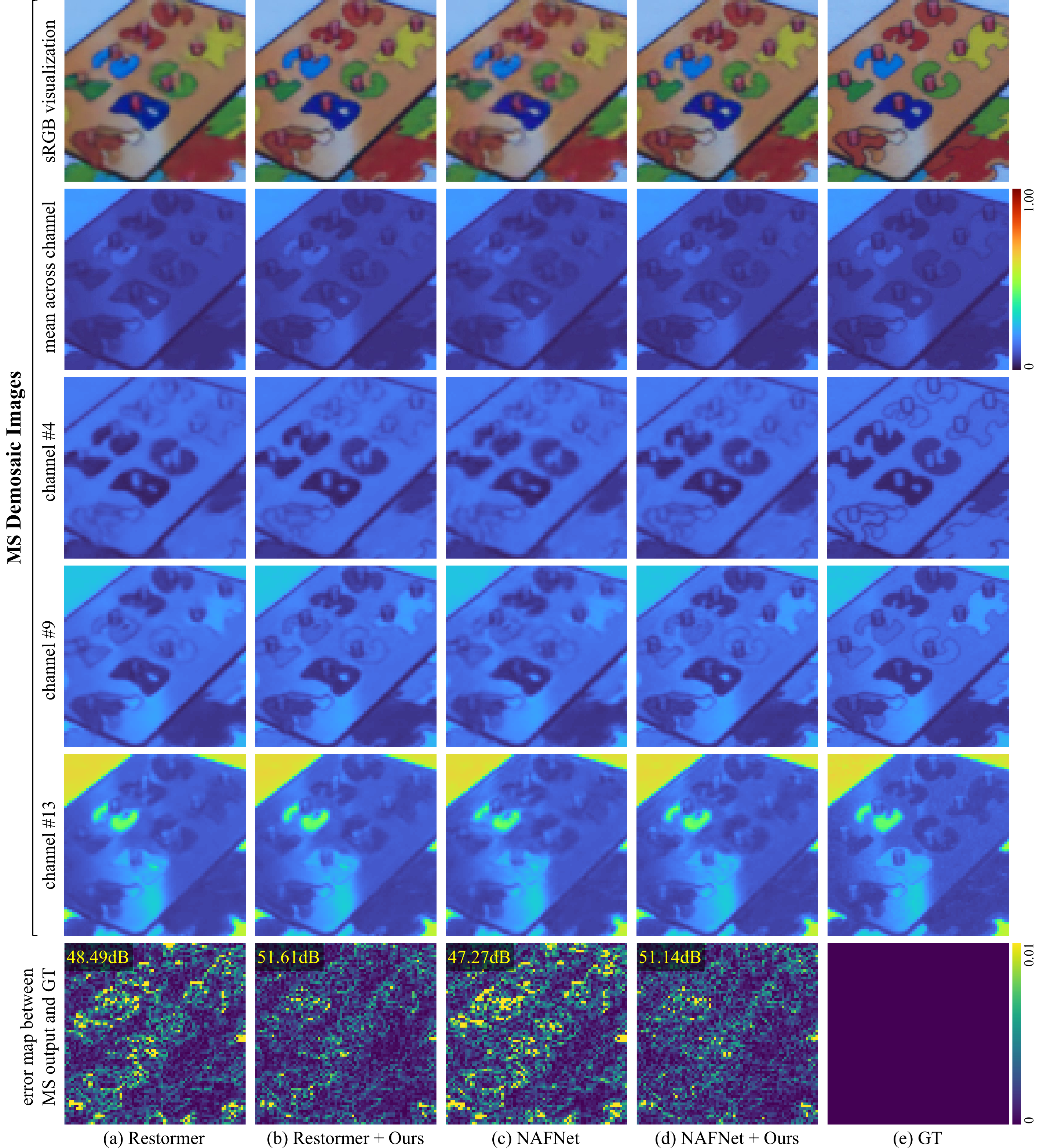}
    \vspace{-6mm}
    \caption{\textbf{Qualitative comparison of 1$\times$ MS demosaicing results} for a dual-camera scenario featuring MS and RGB sensors with the same spatial resolution but employing asymmetric CFAs. The top row shows the predicted MS demosaics converted to the sRGB color space using the color conversion matrix $C$ (\cref{eq:cc}) and camera metadata, with CIE D65 as the reference white point. The second row presents the MS demosaic output averaged across the channel dimension, while the third to fifth rows display per-channel MS demosaic outputs for the 4th, 9th, and 13th channel indices, respectively. The final row visualizes the error maps between the restored and ground-truth MS demosaiced images.}
    \label{fig:1xresults6}
    \vspace{-4mm}
\end{figure*}
\begin{figure*}
    \centering
    \includegraphics[width=1.0\textwidth]{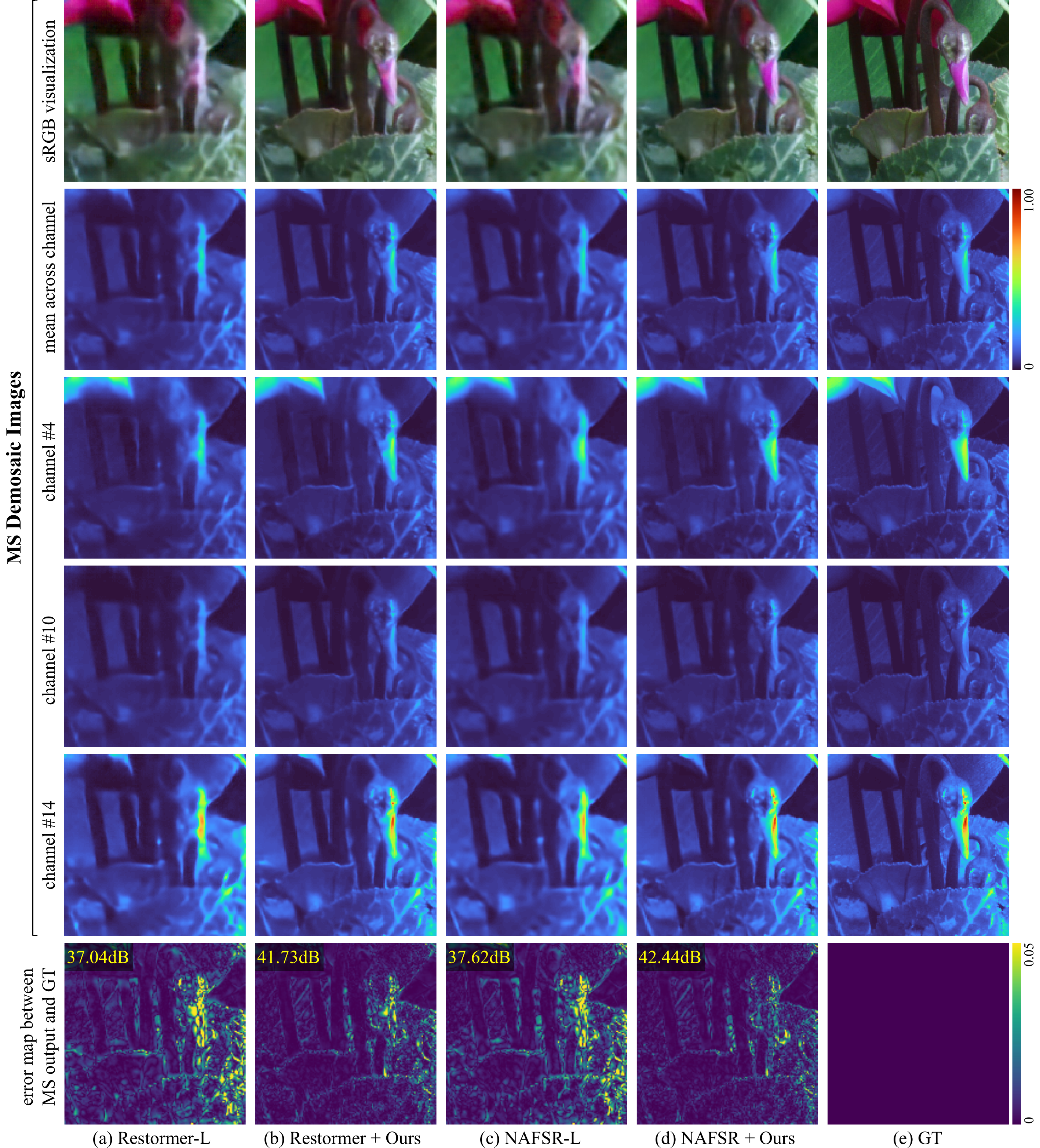}
    \vspace{-6mm}
    \caption{\textbf{Qualitative comparison of 4$\times$ MS demosaicing results} for a dual-camera scenario featuring MS and RGB sensors with the same spatial resolution but employing asymmetric CFAs. The top row shows the predicted MS demosaics converted to the sRGB color space using the color conversion matrix $C$ (\cref{eq:cc}) and camera metadata, with CIE D65 as the reference white point. The second row presents the MS demosaic output averaged across the channel dimension, while the third to fifth rows display per-channel MS demosaic outputs for the 4th, 10th, and 14th channel indices, respectively. The final row visualizes the error maps between the restored and ground-truth MS demosaiced images.}
    \label{fig:4xresults1}
    \vspace{-4mm}
\end{figure*}
\begin{figure*}
    \centering
    \includegraphics[width=1.0\textwidth]{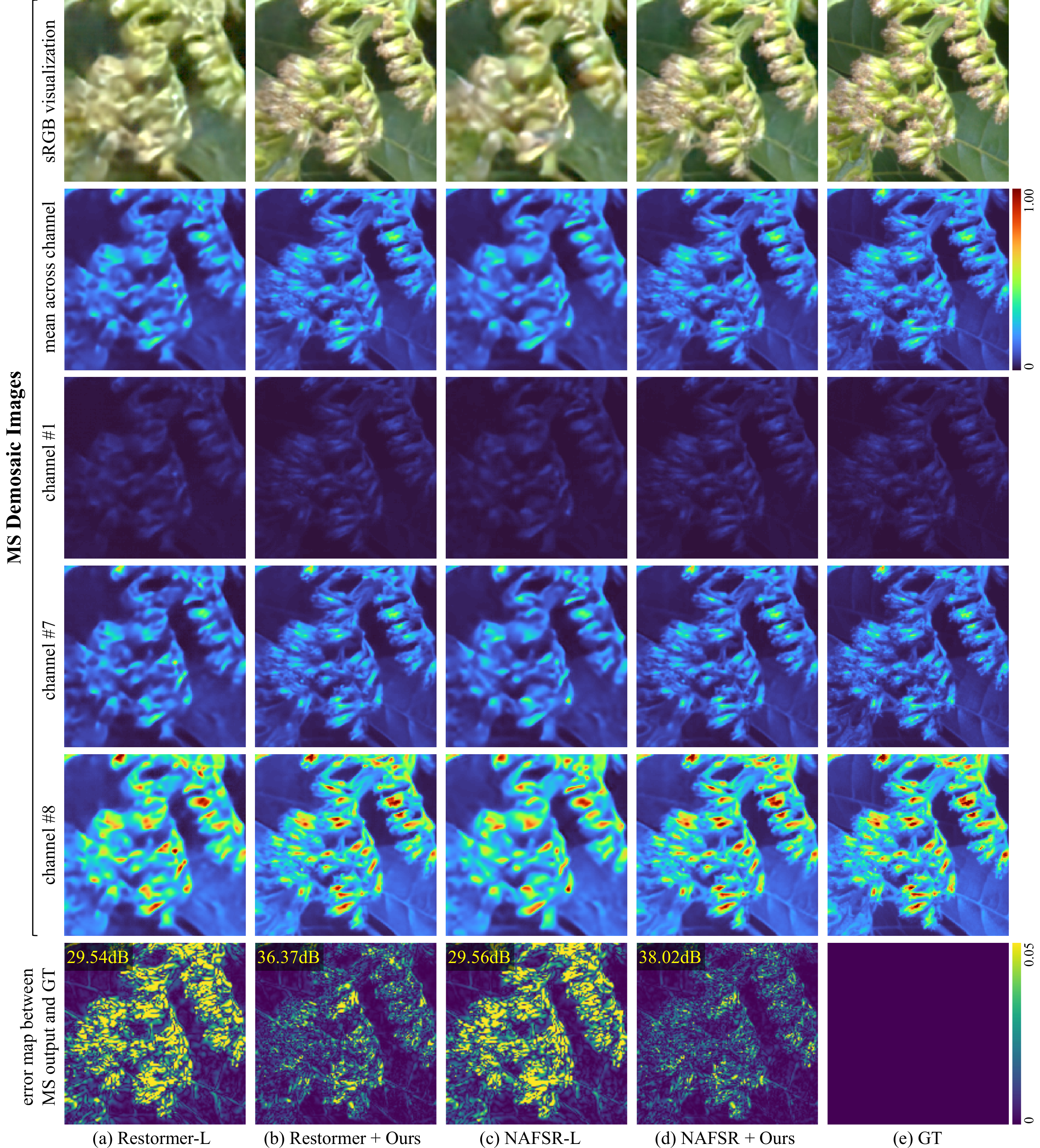}
    \vspace{-6mm}
    \caption{\textbf{Qualitative comparison of 4$\times$ MS demosaicing results} for a dual-camera scenario featuring MS and RGB sensors with the same spatial resolution but employing asymmetric CFAs. The top row shows the predicted MS demosaics converted to the sRGB color space using the color conversion matrix $C$ (\cref{eq:cc}) and camera metadata, with CIE D65 as the reference white point. The second row presents the MS demosaic output averaged across the channel dimension, while the third to fifth rows display per-channel MS demosaic outputs for the 1st, 7th, and 8th channel indices, respectively. The final row visualizes the error maps between the restored and ground-truth MS demosaiced images.}
    \label{fig:4xresults2}
    \vspace{-4mm}
\end{figure*}
\begin{figure*}
    \centering
    \includegraphics[width=1.0\textwidth]{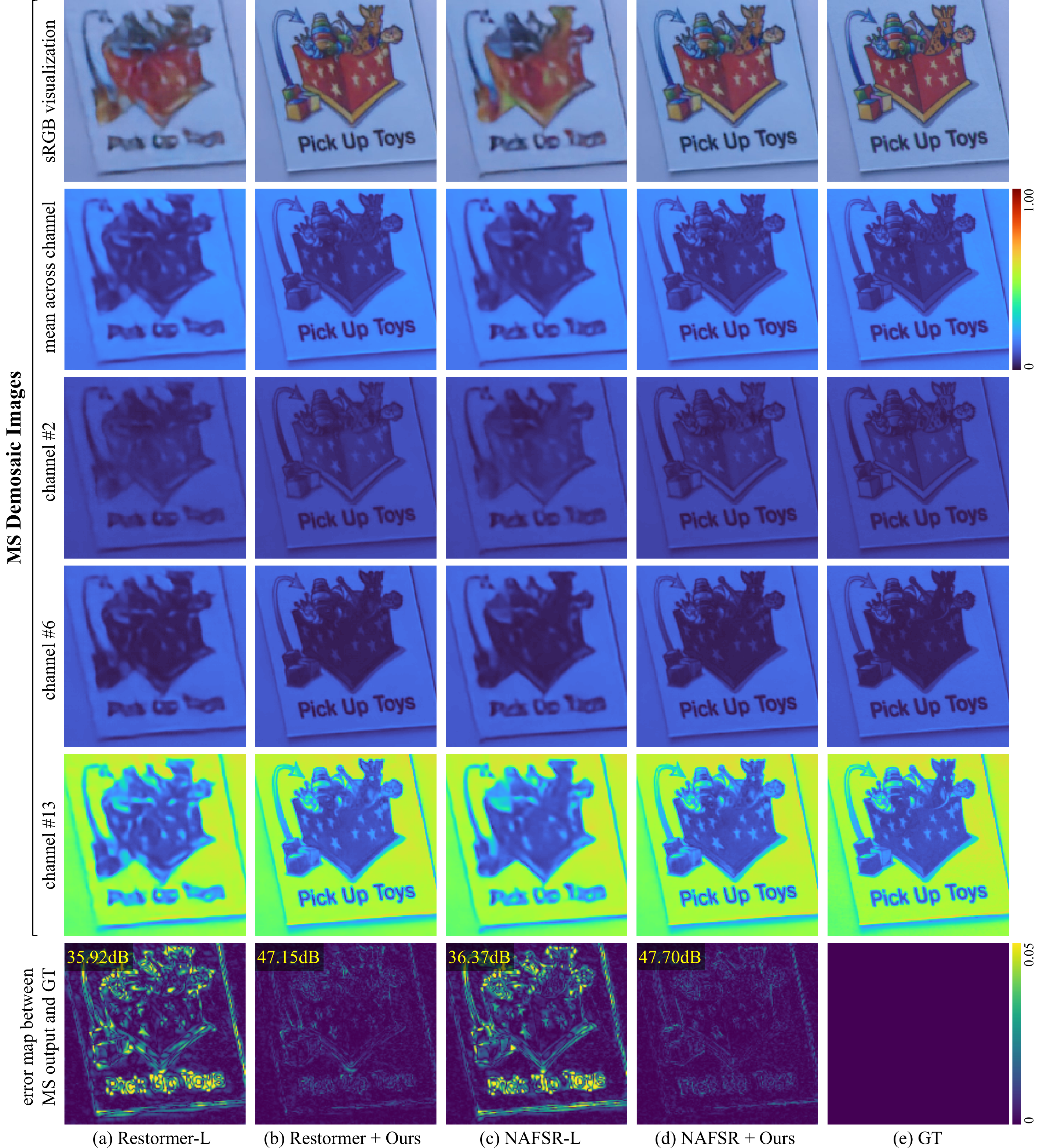}
    \vspace{-6mm}
    \caption{\textbf{Qualitative comparison of 4$\times$ MS demosaicing results} for a dual-camera scenario featuring MS and RGB sensors with the same spatial resolution but employing asymmetric CFAs. The top row shows the predicted MS demosaics converted to the sRGB color space using the color conversion matrix $C$ (\cref{eq:cc}) and camera metadata, with CIE D65 as the reference white point. The second row presents the MS demosaic output averaged across the channel dimension, while the third to fifth rows display per-channel MS demosaic outputs for the 2nd, 6th, and 13th channel indices, respectively. The final row visualizes the error maps between the restored and ground-truth MS demosaiced images.}
    \label{fig:4xresults3}
    \vspace{-4mm}
\end{figure*}
\begin{figure*}
    \centering
    \includegraphics[width=1.0\textwidth]{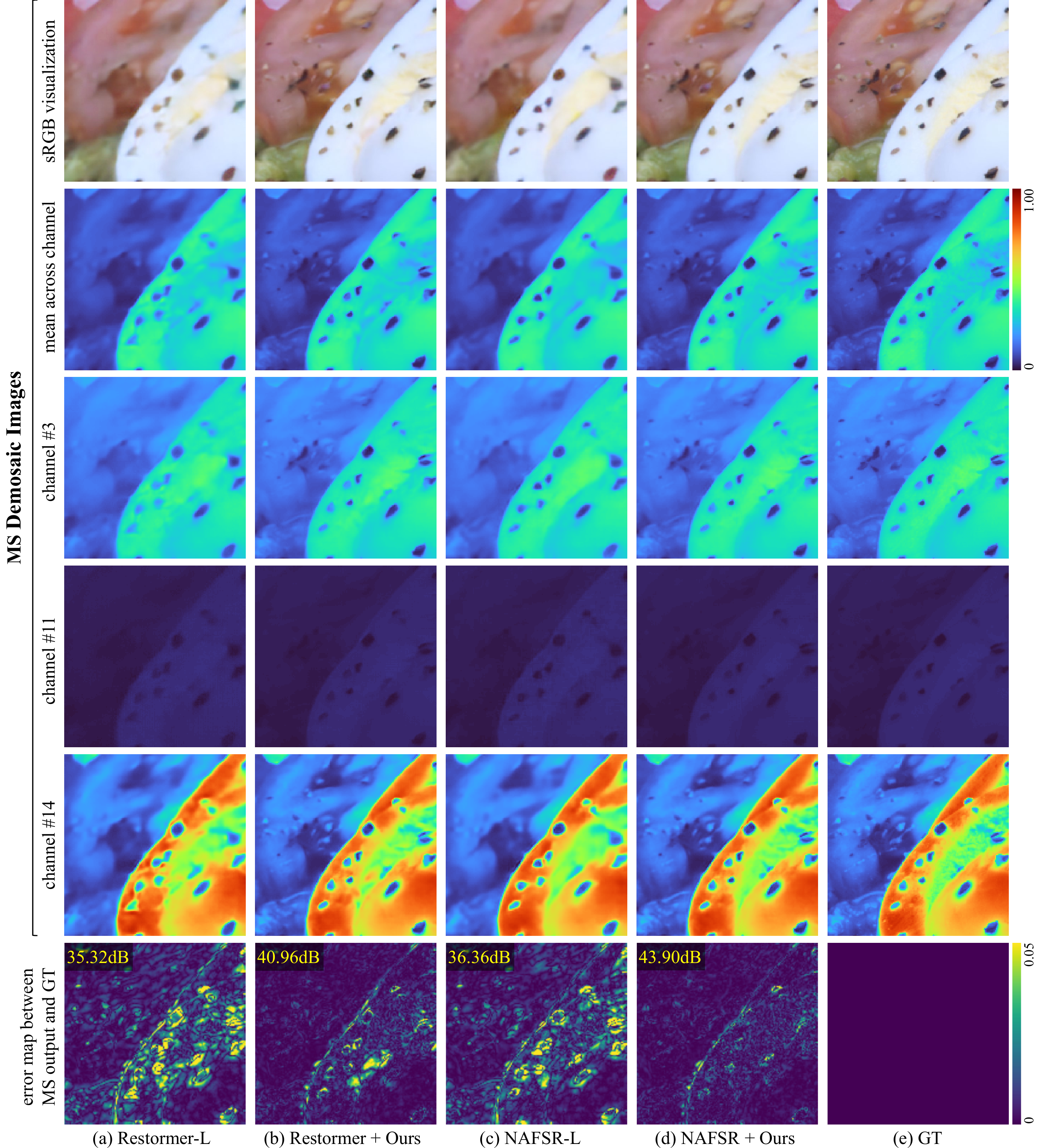}
    \vspace{-6mm}
    \caption{\textbf{Qualitative comparison of 4$\times$ MS demosaicing results} for a dual-camera scenario featuring MS and RGB sensors with the same spatial resolution but employing asymmetric CFAs. The top row shows the predicted MS demosaics converted to the sRGB color space using the color conversion matrix $C$ (\cref{eq:cc}) and camera metadata, with CIE D65 as the reference white point. The second row presents the MS demosaic output averaged across the channel dimension, while the third to fifth rows display per-channel MS demosaic outputs for the 3rd, 11th, and 14th channel indices, respectively. The final row visualizes the error maps between the restored and ground-truth MS demosaiced images.}
    \label{fig:4xresults4}
    \vspace{-4mm}
\end{figure*}
\begin{figure*}
    \centering
    \includegraphics[width=1.0\textwidth]{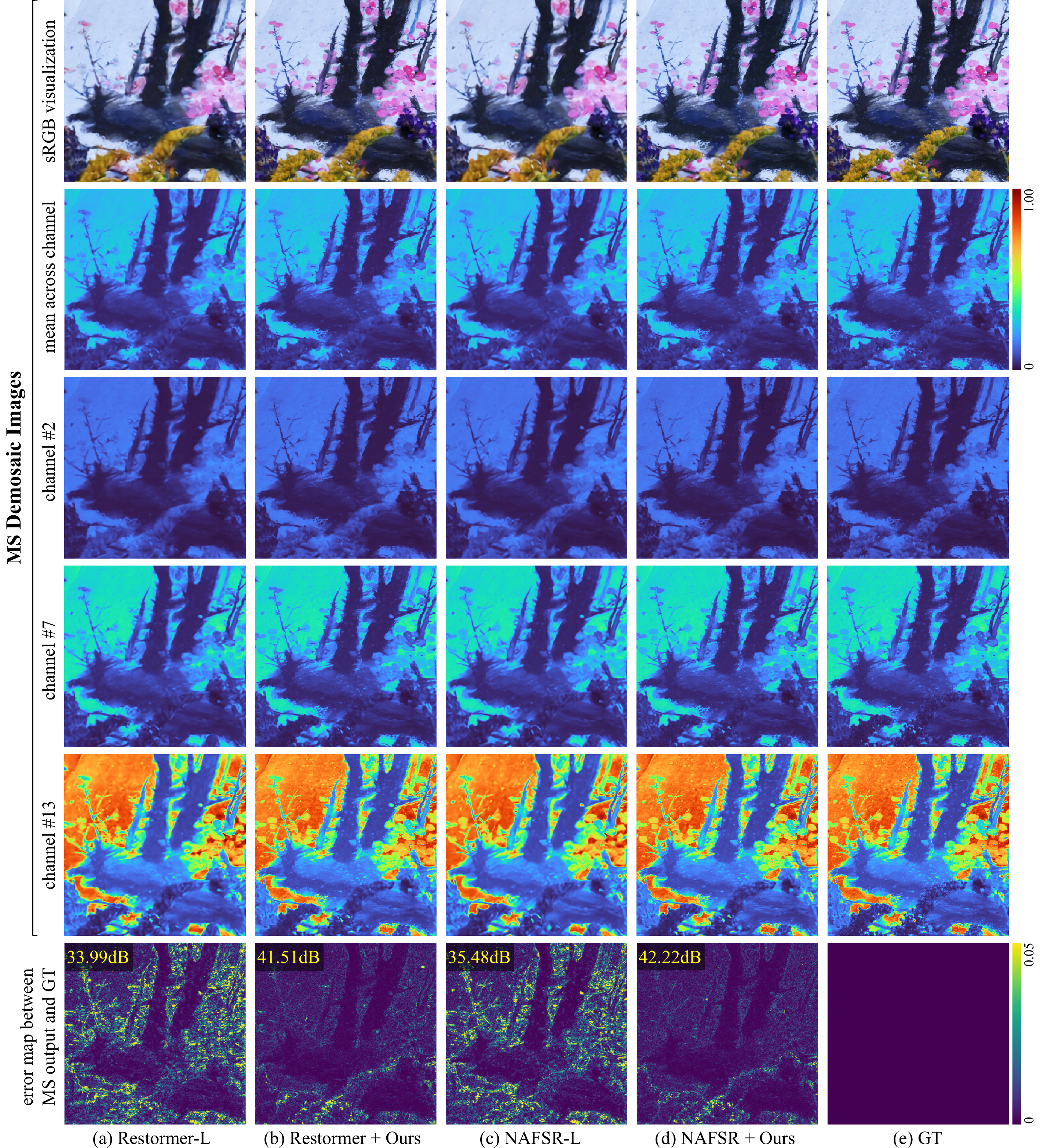}
    \vspace{-6mm}
    \caption{\textbf{Qualitative comparison of 4$\times$ MS demosaicing results} for a dual-camera scenario featuring MS and RGB sensors with the same spatial resolution but employing asymmetric CFAs. The top row shows the predicted MS demosaics converted to the sRGB color space using the color conversion matrix $C$ (\cref{eq:cc}) and camera metadata, with CIE D65 as the reference white point. The second row presents the MS demosaic output averaged across the channel dimension, while the third to fifth rows display per-channel MS demosaic outputs for the 2nd, 7th, and 13th channel indices, respectively. The final row visualizes the error maps between the restored and ground-truth MS demosaiced images.}
    \label{fig:4xresults5}
    \vspace{-4mm}
\end{figure*}
\begin{figure*}
    \centering
    \includegraphics[width=1.0\textwidth]{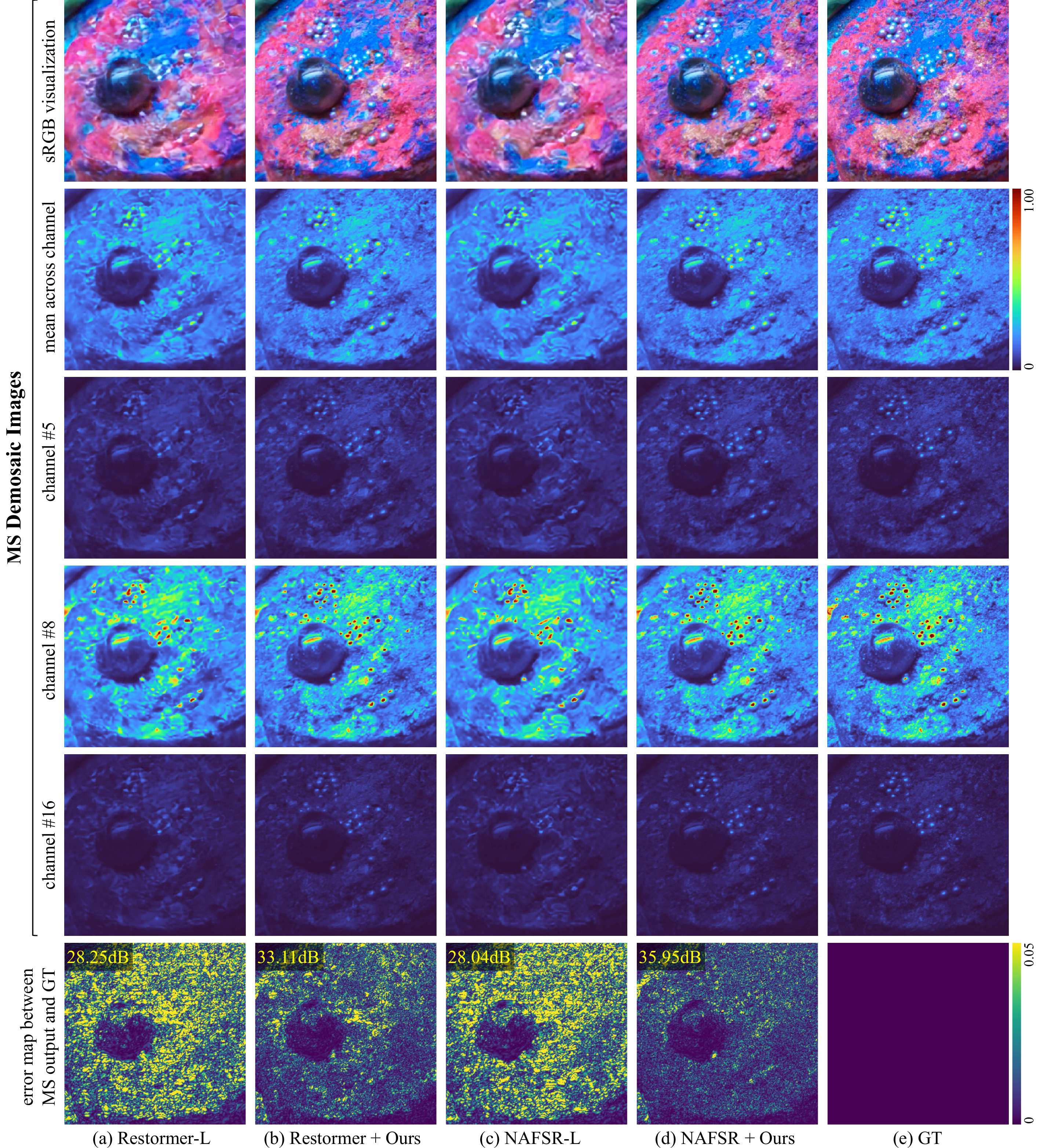}
    \vspace{-6mm}
    \caption{\textbf{Qualitative comparison of 4$\times$ MS demosaicing results} for a dual-camera scenario featuring MS and RGB sensors with the same spatial resolution but employing asymmetric CFAs. The top row shows the predicted MS demosaics converted to the sRGB color space using the color conversion matrix $C$ (\cref{eq:cc}) and camera metadata, with CIE D65 as the reference white point. The second row presents the MS demosaic output averaged across the channel dimension, while the third to fifth rows display per-channel MS demosaic outputs for the 5th, 8th, and 16th channel indices, respectively. The final row visualizes the error maps between the restored and ground-truth MS demosaiced images.}
    \label{fig:4xresults6}
    \vspace{-4mm}
\end{figure*}
\begin{figure*}
    \centering
    \includegraphics[width=1.0\textwidth]{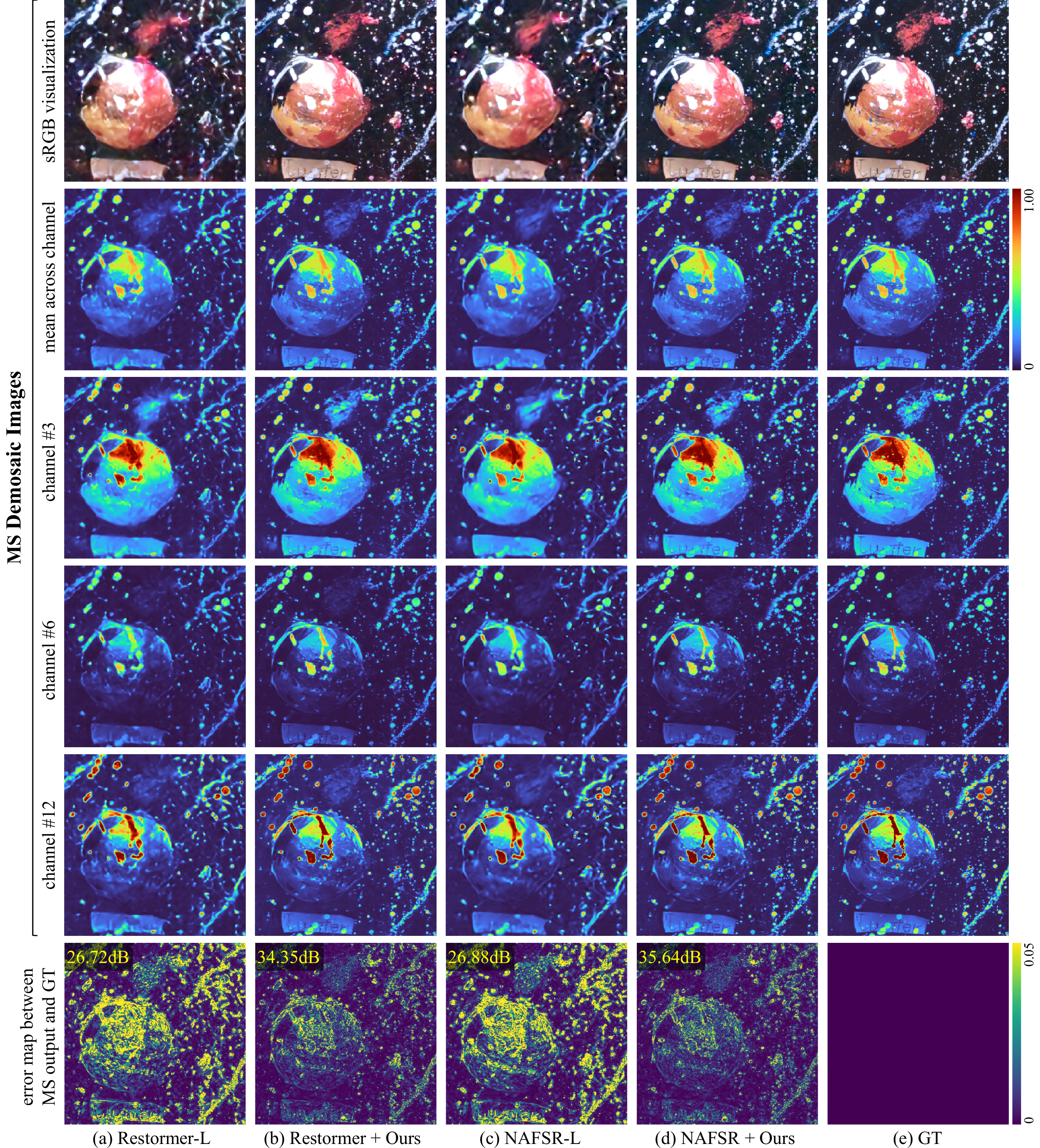}
    \vspace{-6mm}
    \caption{\textbf{Qualitative comparison of 4$\times$ MS demosaicing results} for a dual-camera scenario featuring MS and RGB sensors with the same spatial resolution but employing asymmetric CFAs. The top row shows the predicted MS demosaics converted to the sRGB color space using the color conversion matrix $C$ (\cref{eq:cc}) and camera metadata, with CIE D65 as the reference white point. The second row presents the MS demosaic output averaged across the channel dimension, while the third to fifth rows display per-channel MS demosaic outputs for the 3rd, 6th, and 12th channel indices, respectively. The final row visualizes the error maps between the restored and ground-truth MS demosaiced images.}
    \label{fig:4xresults7}
    \vspace{-4mm}
\end{figure*}
\begin{figure*}
    \centering
    \includegraphics[width=1.0\textwidth]{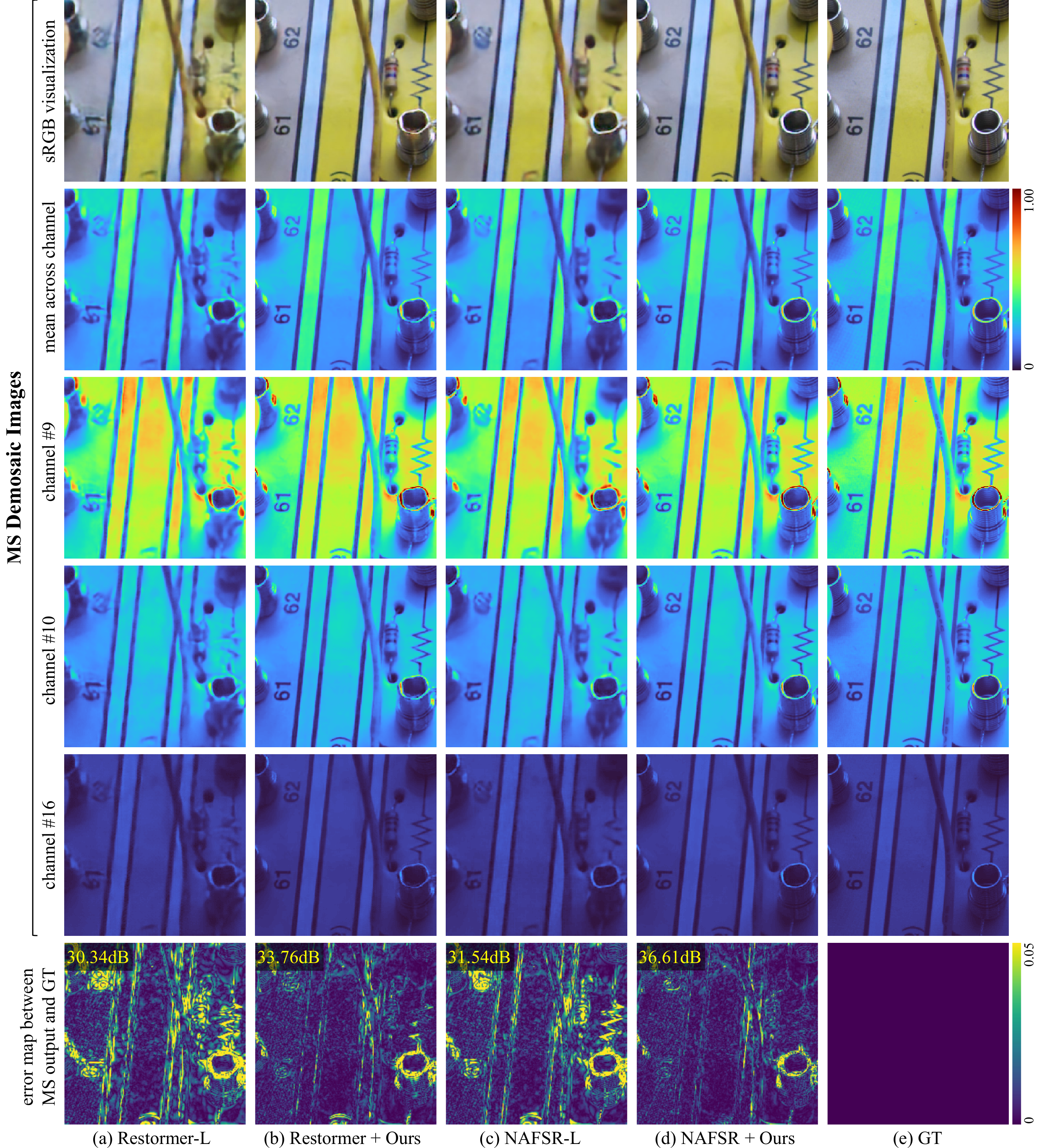}
    \vspace{-6mm}
    \caption{\textbf{Qualitative comparison of 4$\times$ MS demosaicing results} for a dual-camera scenario featuring MS and RGB sensors with the same spatial resolution but employing asymmetric CFAs. The top row shows the predicted MS demosaics converted to the sRGB color space using the color conversion matrix $C$ (\cref{eq:cc}) and camera metadata, with CIE D65 as the reference white point. The second row presents the MS demosaic output averaged across the channel dimension, while the third to fifth rows display per-channel MS demosaic outputs for the 9th, 10th, and 16th channel indices, respectively. The final row visualizes the error maps between the restored and ground-truth MS demosaiced images.}
    \label{fig:4xresults8}
    \vspace{-4mm}
\end{figure*}
\begin{figure*}
    \centering
    \includegraphics[width=1.0\textwidth]{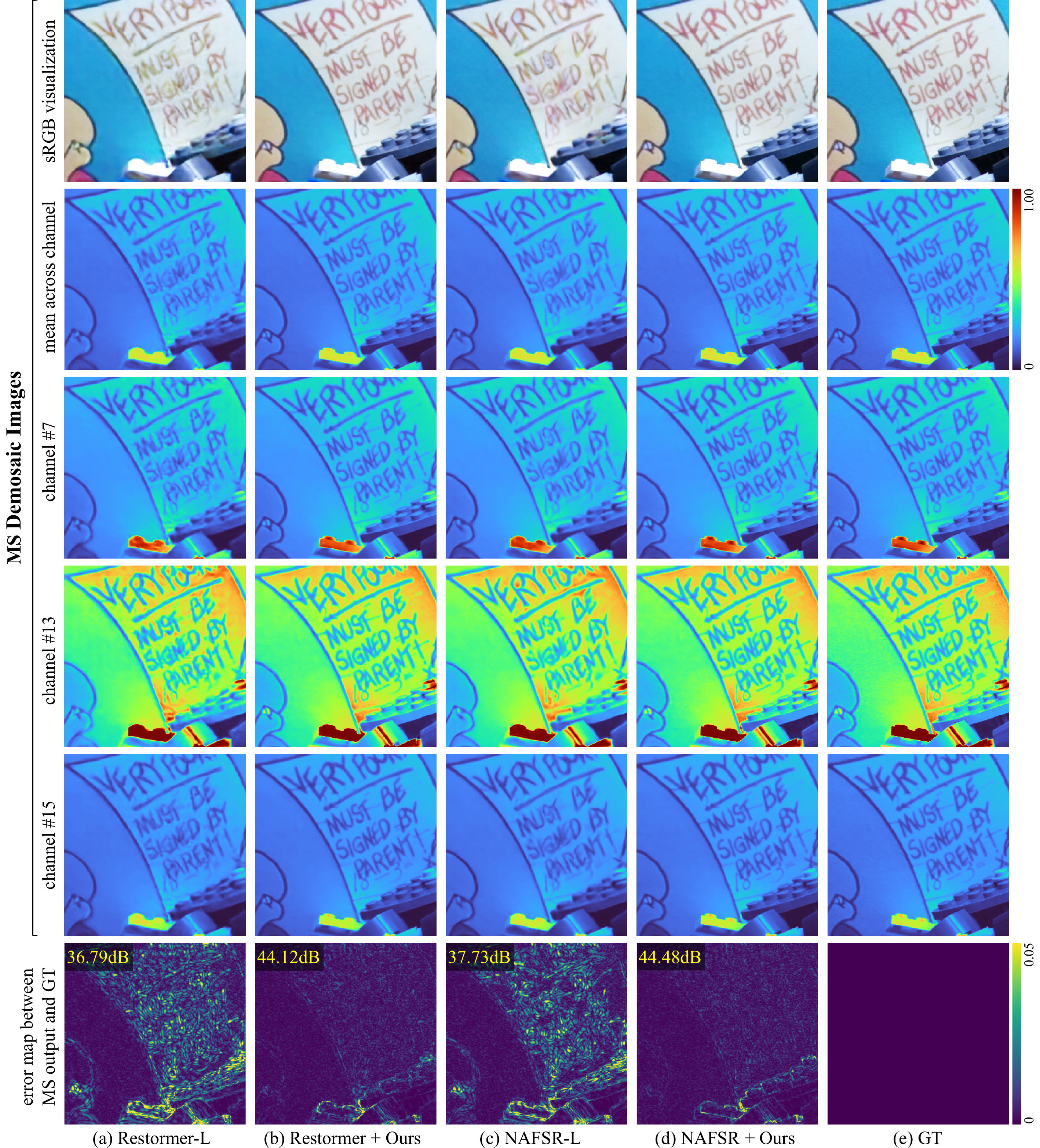}
    \vspace{-6mm}
    \caption{\textbf{Qualitative comparison of 4$\times$ MS demosaicing results} for a dual-camera scenario featuring MS and RGB sensors with the same spatial resolution but employing asymmetric CFAs. The top row shows the predicted MS demosaics converted to the sRGB color space using the color conversion matrix $C$ (\cref{eq:cc}) and camera metadata, with CIE D65 as the reference white point. The second row presents the MS demosaic output averaged across the channel dimension, while the third to fifth rows display per-channel MS demosaic outputs for the 7th, 13th, and 15th channel indices, respectively. The final row visualizes the error maps between the restored and ground-truth MS demosaiced images.}
    \label{fig:4xresults9}
    \vspace{-4mm}
\end{figure*}
\begin{figure*}
    \centering
    \includegraphics[width=1.0\textwidth]{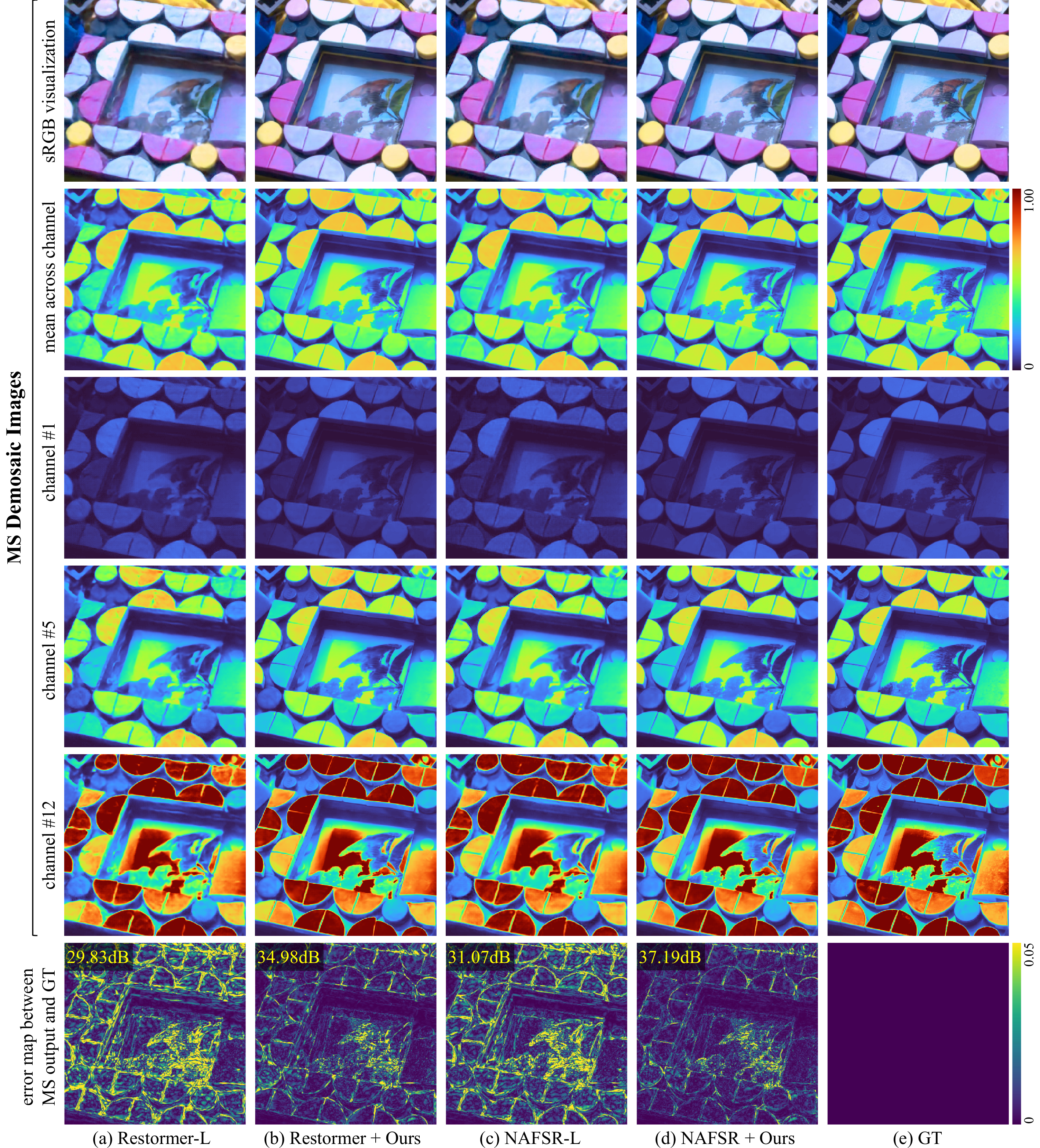}
    \vspace{-6mm}
    \caption{\textbf{Qualitative comparison of 4$\times$ MS demosaicing results} for a dual-camera scenario featuring MS and RGB sensors with the same spatial resolution but employing asymmetric CFAs. The top row shows the predicted MS demosaics converted to the sRGB color space using the color conversion matrix $C$ (\cref{eq:cc}) and camera metadata, with CIE D65 as the reference white point. The second row presents the MS demosaic output averaged across the channel dimension, while the third to fifth rows display per-channel MS demosaic outputs for the 1st, 5th, and 12th channel indices, respectively. The final row visualizes the error maps between the restored and ground-truth MS demosaiced images.}
    \label{fig:4xresults10}
    \vspace{-4mm}
\end{figure*}

We present qualitative results on the proposed dual-camera RGB-MS test set for the 1$\times$ MS demosaicing task (Scenario 1 in Sec.~5.2 of the main paper) in \cref{fig:1xresults1,fig:1xresults2,fig:1xresults3,fig:1xresults4,fig:1xresults5,fig:1xresults6,fig:1xresults7,fig:1xresults8,fig:1xresults9,fig:1xresults10}, and for the 4$\times$ MS demosaicing task (Scenario 2 in Sec.~5.2 of the main paper) in \cref{fig:4xresults1,fig:4xresults2,fig:4xresults3,fig:4xresults4,fig:4xresults5,fig:4xresults6,fig:4xresults7,fig:4xresults8,fig:4xresults9,fig:4xresults10}. The visualized results include MS demosaics converted to the sRGB color space, MS demosaic averaged across the channel dimension, per-channel MS demosaics, and error maps computed between the restored and ground-truth MS demosaiced images.

{
\setlength{\aboverulesep}{0mm}
\setlength{\belowrulesep}{0mm}
\renewcommand{\arraystretch}{1.0} 
\setlength{\tabcolsep}{0pt} 
\begin{table}[t]
    \centering
    \small
    \setlength\tabcolsep{3pt} 
    
    \begin{tabularx}{\columnwidth}{c X c c c c c}
    \toprule
    
    \texttt{Type} & \texttt{Input} & \texttt{Act} & \texttt{K} & \texttt{Ch} & \texttt{S} & \texttt{Output} \\
    
    \midrule
    
    \texttt{conv} & \scriptsize{$f'^{MS}_{l=1}(I^{MS}_{4\times 4})$} & \texttt{LReLU} & 3$\times$3 & $c \times 4$ & 1 & Conv$_1$ \\
    
    \texttt{PixShfl} & Conv$_1$ & - & - & $c$ & - & Up$_1$ \\
    
    \texttt{conv} & Up$_1$ & \texttt{LReLU} & 3$\times$3 & $c \times 4$ & 1 & Conv$_2$ \\
    
    \texttt{PixShfl} & Conv$_2$ & - & - & $c$ & - & Up$_2$ \\
    
    \texttt{conv} & Up$_2$ & \texttt{LReLU} & 3$\times$3 & $c$ & 1 & Final \\
    
    \bottomrule
    
    \end{tabularx}
    
    \vspace{-6pt}
    \caption{
    Architecture of the upsampling module for 4$\times$ MS demosaicing. Abbreviations: \texttt{Act} = Activation, \texttt{K} = Kernel size, \texttt{Ch} = Channels, \texttt{S} = Stride, \texttt{PixShfl} = PixelShuffle, \texttt{Up} = Upsampled. The input $f'^{MS}_{l=1}(I^{MS}_{4\times 4})$ represents the final feature map from the MS demosaicing network $\mathcal{D}_{MS}$ at scale level $l=1$.
    }
    \vspace{-12pt}
    \label{tab:upsampling_module}
\end{table}
}






\clearpage

{
    \small
    \bibliographystyle{compile/ieeenat_fullname}
    \bibliography{compile/main}
}
\end{document}